\documentclass{article}

\usepackage{arxiv}

\usepackage[utf8]{inputenc} % allow utf-8 input
\usepackage[T1]{fontenc}    % use 8-bit T1 fonts
\usepackage{hyperref}       % hyperlinks
\usepackage{url}            % simple URL typesetting
\usepackage{booktabs}       % professional-quality tables
\usepackage{amsfonts}       % blackboard math symbols
\usepackage{nicefrac}       % compact symbols for 1/2, etc.
\usepackage{microtype}      % microtypography
\usepackage{lipsum}		% Can be removed after putting your text content

\usepackage{graphicx}
\usepackage{floatrow}
\usepackage[numbers]{natbib}
\usepackage{doi}

\usepackage{tikz}
\usetikzlibrary{decorations.fractals,spy}
\usepackage{pifont}
\usepackage{xcolor}
\usepackage{colortbl}
\usepackage{multirow}

\usepackage{wrapfig}
\usepackage{amsmath}
\usepackage{algorithmicx}
\usepackage{algpseudocode}
\usepackage{algcompatible}
\usepackage[ruled]{algorithm}

\newcommand{\cmark}{\ding{51}}%
\newcommand{\xmark}{\ding{55}}%

\usepackage[table,xcdraw]{xcolor}

\title{Racing in Volume with Flow Ensembles}
\renewcommand{\undertitle}{}
\renewcommand{\headeright}{Mallick et al.}

\date{} 					% Or removing it

\author{%
  \begin{tabular}[t]{c}
    Saswat Subhajyoti Mallick \quad
    Riu Cherdchusakulchai \quad
    Marc Ruiz Olle \\
    Albert Mosella-Montoro \quad
    Jose Ribeiro-Gomes \quad 
    Francisco Vicente Carrasco \quad 
    Fernando De La Torre \\
    \end{tabular}%
  \quad
  \and
  \begin{tabular}[t]{c}
    Carnegie Mellon University \\
  \end{tabular}%
  \\
  \href{https://humansensinglab.github.io/monaco4d/}{Project Page}
}

\begin{document}
\maketitle

\begingroup
\renewcommand\thefootnote{}
\footnotetext{Published at ECCV 2026. The final authenticated version is available online at: \url{https://doi.org/10.1007/978-3-032-37041-9_15}}
\endgroup

\begin{figure}[h]
    \centering
    \includegraphics[width=0.8\linewidth]{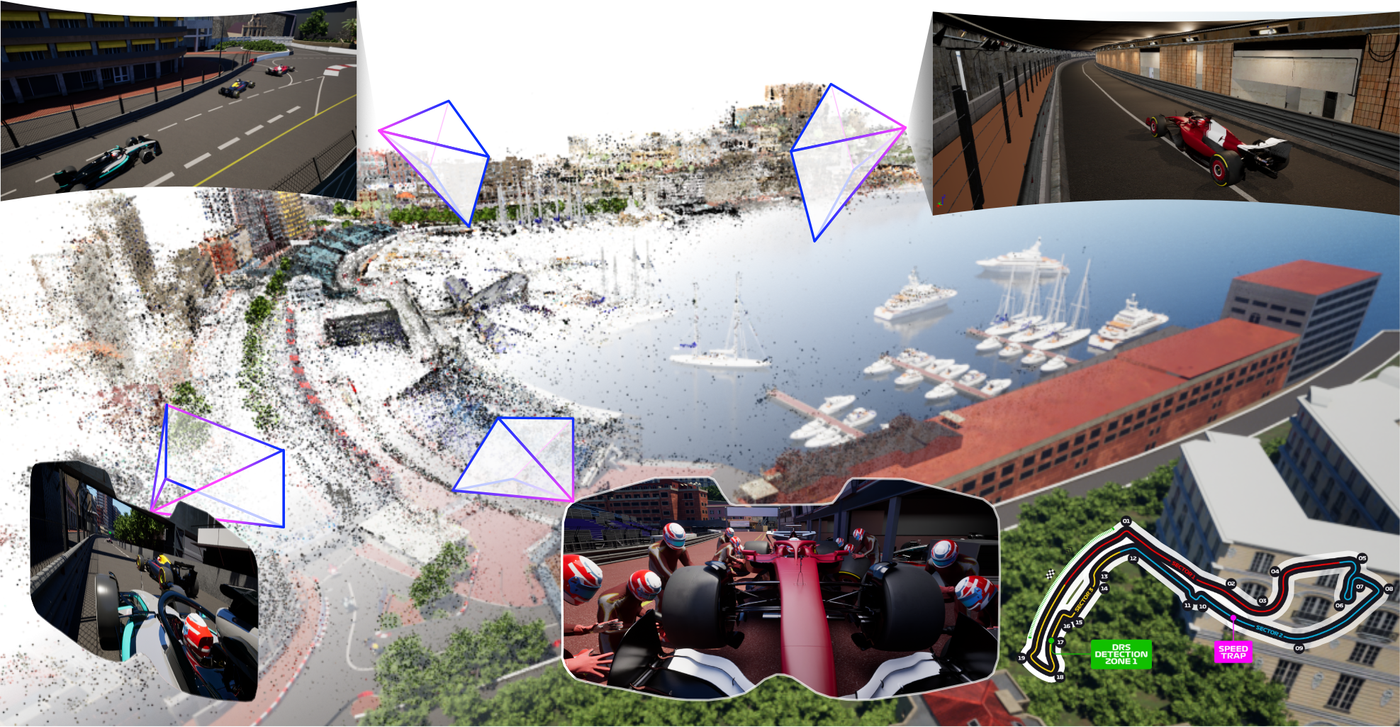}
    \caption{
        Given multi-view video streams, \textbf{FastFlowGS} reconstructs geometry for high fidelity, photorealistic novel-view rendering. We introduce \textbf{Monaco4D}, the first synthetic outdoor dataset, with Formula 1 racing scenarios to benchmark extreme motion.
    }

\label{fig:teaser}
\end{figure}

\begin{abstract}
Streaming 4D reconstruction has been demonstrated only indoors, on dense camera rigs surrounding subjects that move at human pace. Outdoor 4D reconstruction exists but relies either on cameras mounted on the moving vehicle itself, or on limited-coverage arrays observing quasi-static subjects offline. The case that actually matters for spectators is a fast-moving subject, watched from a sparse ring of allocentric cameras, streaming. No method targets this, and no benchmark exists to evaluate one. To this end, we introduce \textbf{FastFlowGS}, a streaming 4D Gaussian Splatting method for reconstructing fast-moving subjects from a small set of fixed external cameras, and \textbf{Monaco4D}, a photorealistic Unreal Engine 5 benchmark for high-speed outdoor reconstruction. FastFlowGS fuses sparse matches, semi-dense tracks, and dense optical flow by lifting each signal to 3D with geometric uncertainty and combining them through a Kalman-style temporal update. \textbf{Monaco4D} provides Formula 1 sequences under varied illumination from trackside, onboard, and drone viewpoints with dense ground truth. On CMU-Panoptic (a public dataset), FastFlowGS exceeds the strongest baseline by 12.6\% VMAF at 35\% greater efficiency. On Monaco4D, where existing streaming methods degrade severely, it improves dynamic-region PSNR by up to 18.6\% with 28.3\% lower per-frame optimization time. Dataset and additional details can be found at \href{https://humansensinglab.github.io/monaco4d/}{humansensinglab.github.io/monaco4d}.


\end{abstract}

\section{Introduction}
\label{sec:introduction}

\emph{``Auto racing began five minutes after the second car was built.''} \hfill $\sim$ Henry Ford

The instinct to race is older than the sport itself, and no series pushes it further than Formula 1. Over 820 million fans follow F1, yet nearly all of them watch through a single director-chosen frame that compresses a spatial spectacle into a confined, flat image. 
VR and AR offer a way out, placing the viewer anywhere in the scene, but they demand real-time 4D reconstruction of dynamic outdoor environments, a challenge that no sport pushes harder than Formula 1.

Cars pass trackside cameras at over 300\,km/h, their carbon-fibre bodywork specularly reflecting the environment, while lighting shifts abruptly inside tunnels. The core difficulty is not speed itself but the pixel displacements it produces. Filmed laterally at 30\,Hz, a Formula 1 car moves 200 to 400 pixels between consecutive frames, an order of magnitude beyond what optical flow~\cite{teed2020raftrecurrentallpairsfield,moing2024denseopticaltrackingconnecting} and point trackers~\cite{karaev2024cotracker3,doersch2024bootstap} handle reliably. 

Existing methods take one of two approaches, and neither holds at this scale.
Deformable representations~\cite{pumarola2020d,cao2023hexplanefastrepresentationdynamic,Wu_2024_CVPR,yang2023deformable3dgs} assume slow or camera-dominated motion and degrade when displacements grow large~\cite{wang2023flowsupervisiondeformablenerf}.
Streaming methods~\cite{sun20243dgstreamontheflytraining3d,gao2024hicomhierarchicalcoherentmotion,trackersplat,igs} each commit to a single pixel-space correspondence mechanism, so when that mechanism fails, the error propagates directly into the reconstruction.
Available benchmarks further reinforce the problem.
Driving datasets~\cite{nuscenes2019,sun2020scalabilityperceptionautonomousdriving,liao2022kitti360noveldatasetbenchmarks} are captured from ego-vehicles and lack the lateral displacements of fixed trackside cameras, while indoor sequences~\cite{Joo_2017_TPAMI,li2022neural3dvideosynthesis} have slow dynamics and controlled lighting.
Together, they leave the combination of large pixel displacements, diverse exocentric viewpoints, and uncontrolled outdoor illumination largely unexplored.

\textbf{Monaco4D} is the first multi-view benchmark built around this gap.
Rendered in Unreal Engine~5~\cite{unrealengine5} on a to-scale replica of the Monaco circuit with vehicle dynamics approximating real lap profiles, it provides seven sequences, five illumination conditions, three camera modalities (trackside, onboard, drone), and dense ground-truth annotations.
We pair it with \textbf{FastFlowGS}, a streaming Gaussian Splatting method built on a simple insight that \emph{no single correspondence mechanism is reliable under large displacements, but their disagreement reveals where each one fails.}
FastFlowGS fuses sparse feature matches, semi-dense point tracks, and dense optical flow, using cross-level agreement scoring to identify which signal is trustworthy at each location.
The reliable correspondences are lifted into 3D through uncertainty-aware triangulation and merged with a Kalman temporal prior, producing per-Gaussian positions that converge at a fraction of the cost of standard streaming approaches.
\\

%\noindent We evaluate on CMU Panoptic~\cite{Joo_2017_TPAMI} and Monaco4D.
Our contributions are:
\begin{enumerate}
    \item \textbf{FastFlowGS}, a streaming 4D Gaussian reconstruction method that fuses correspondence signals across scales to handle large pixel displacements.
    \item \textbf{Monaco4D}, the first multi-view outdoor dataset, featuring F1 sequences with extreme displacements, varied illumination, and camera modalities.
    \item SOTA on CMU-Panoptic~\cite{Joo_2017_TPAMI} (12.6\% higher VMAF, 35\% faster) and the first streaming reconstruction on Monaco4D, where prior methods fail entirely.
\end{enumerate}

\section{Related Work}
\label{sec:literature}

Dynamic scene reconstruction has advanced rapidly in controlled indoor and egocentric driving settings, yet fast-moving outdoor subjects observed from spectator viewpoints remain largely unexplored.

\textbf{Deformable Representations and Gaussian Playback.} A common paradigm models dynamic scenes via a canonical representation with per-frame deformations. D-NeRF~\cite{pumarola2020d} pioneered deformable radiance fields, K-Planes~\cite{fridovichkeil2023kplanesexplicitradiancefields} and HexPlane~\cite{cao2023hexplanefastrepresentationdynamic} introduced efficient explicit factorizations, Fourier PlenOctrees~\cite{wang2022fourierplenoctreesdynamicradiance} and VideoRF~\cite{wang2023videorfrenderingdynamicradiance} enable real-time rendering. All assume slow deformation or dominant camera motion and degrade under rapid object displacement~\cite{wang2023flowsupervisiondeformablenerf}. Gaussian extensions~\cite{Wu_2024_CVPR,li2024spacetimegaussianfeaturesplatting,yang2023deformable3dgs,oh2025hybrid3d4dgaussiansplatting,zhu2024motiongs,liang2025gaufregaussiandeformationfields} achieve large speedups for offline playback but target studio or indoor sequences where per-frame displacements rarely exceed tens of pixels.

\textbf{Streaming and Feed-Forward 4D Reconstruction.} Streaming methods accelerate per-frame convergence through diverse strategies: Dynamic-3DGS~\cite{dynamic3dgs} enforces persistent attributes with rigidity priors. 3DGStream~\cite{sun20243dgstreamontheflytraining3d} uses a Neural Transformation Cache; HiCoM~\cite{gao2024hicomhierarchicalcoherentmotion} and ReCon-GS~\cite{fu2025recongscontinuumpreservedgaussianstreaming} use hierarchical grids with dynamic reconfiguration; QUEEN~\cite{girish2024queen}, GIFStream~\cite{li2025gifstream4dgaussianbasedimmersive}, and Instant Gaussian Stream~\cite{igs} employ residual quantization, feature streams, and flow-based anchor control, respectively. TrackerSplat~\cite{trackersplat}, the closest predecessor to FastFlowGS, initializes Gaussians from point trackers with per-frame finetuning; we extend it with multi-resolution fusion, uncertainty-aware triangulation, and a variational position prior. Feed-forward approaches~\cite{zhang2026d4rt,karhade2025any4d,xu20254dgt,wang2024dust3rgeometric3dvision,leroy2024groundingimagematching3d,zhang2024monst3r,han2025d,wang2025continuous,chen2025long3rlongsequencestreaming,khafizov2025gcut3rguided3dreconstruction,zhang2025efficientlyreconstructingdynamicscenes} bypass test-time optimization via pointmap regression or persistent-state models but remain too compute-heavy for real-time use.

\textbf{Outdoor Dynamic Scenes and Sports Reconstruction.}
EmerNeRF~\cite{yang2023emernerf} decomposes Waymo sequences into static, dynamic, and flow fields; StreetSurf~\cite{guo2023streetsurf} and Street Gaussians~\cite{yan2024street} adapt implicit surfaces and 3DGS to forward-facing trajectories; SEED4D~\cite{kästingschäfer2025seed4dsyntheticegoexodynamic} adds exocentric views but remains urban and pedestrian-scale.
All assume ego-motion with strong forward parallax, which breaks in spectator settings where objects move faster than the camera baseline.
In sports, industrial systems~\cite{hawkeye,huang2019tracknetdeeplearningnetwork,Shishido2014TrajectoryEO,gossard2025tt3dtabletennis3d} reconstruct trajectories of compact objects (balls, shuttlecocks), not full 4D scenes.
Free-viewpoint sports video dates to Kanade et al.'s Virtualized Reality dome~\cite{10.5555/266989.267081}; AerialRecon~\cite{hong2025free} reconstructs outdoor athletes from a single drone, and LiveSplats~\cite{10.1145/3731214} demonstrates real-time Gaussian reconstruction of indoor arenas.
However, these systems operate under controlled lighting with subjects moving at up to 10\,m/s; neither the displacement regime nor the illumination conditions of outdoor motorsport have been addressed.

\begin{figure}[t]
    \centering
    \includegraphics[width=1\linewidth]{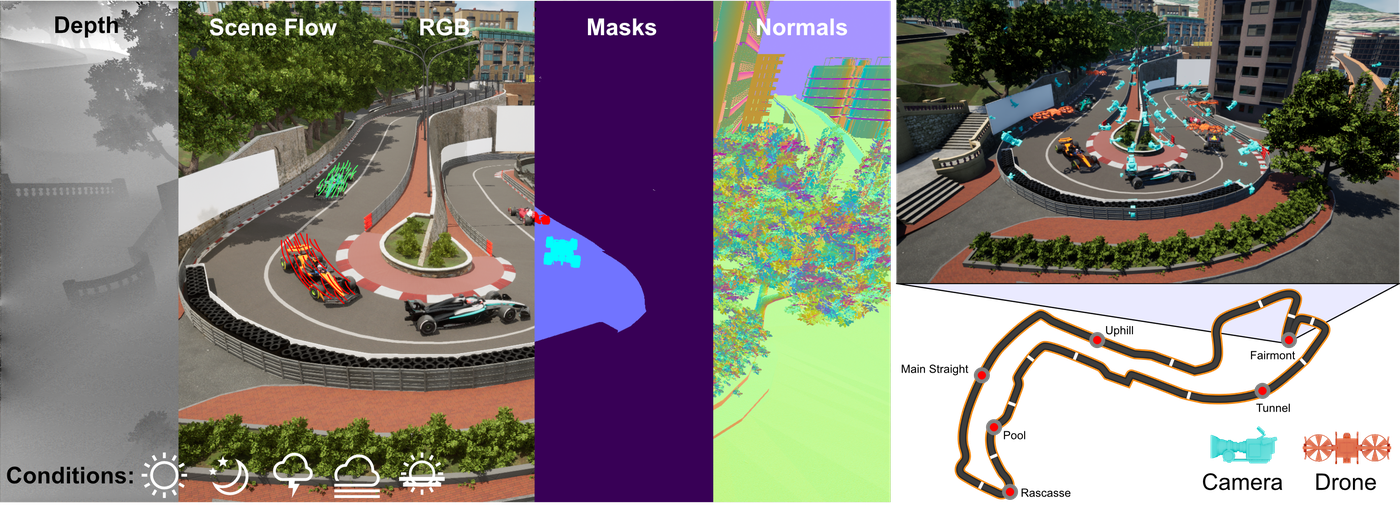}
    \caption{
        \textbf{Overview of the Monaco4D benchmark.}: We cover six sections of the Monaco circuit with dense multi-view camera coverage, multimodal ground-truth annotations (RGB, depth, normals, masks, scene flow), under five illumination conditions.
    }
    \label{fig:dataset_fig}
\end{figure}

\section{Monaco4D Benchmark}
\label{sec:dataset}

Monaco4D is a synthetic multi-view dataset rendered in Unreal Engine~5 with path-traced photorealism.
It comprises six sequences on a to-scale replica of the Monaco Grand Prix circuit, each capturing a distinct segment (shown in Fig. 4 in the supplemental) with single-car and multi-car variants that include realistic maneuvers such as overtaking, near-crashes, and formation drafting (Fig.~\ref{fig:dataset_fig}).

\paragraph{Camera rig.}
Each sequence provides fixed \emph{trackside cameras} that are positioned near the guardrails along the circuit, \emph{onboard cameras} that follow FIA specifications~\cite{fia2026} with seven per vehicle plus one helmet-mounted unit, covering forward, rear, lateral, and driver perspectives, and \emph{drone cameras} follow three trajectories that simulate broadcast operation, including subject tracking, rapid panning, and abrupt zoom changes.
% All cameras share known intrinsics and time-synchronized extrinsics. 
More details are in Sec. 2.1 of the supplemental.

\paragraph{Illumination conditions.}
Each sequence is rendered under five conditions (day, evening, night, fog, rain), producing 30 sequence-illumination combinations before car-count variations.
We also add within-clip transitions like tunnel entries, intermittent shade from buildings, and streetlamp pools at night, to challenge the slowly-varying illumination assumption of most photometric methods.

\paragraph{Ground-truth annotations.}
Each frame provides path-traced RGB, surface normals, metric depth, per-instance segmentation masks, and dense 3D scene flow.
\\

\clearpage

\section{FastFlowGS}
\label{sec:method}
\label{sec:overview}

\begin{figure}[t]
    \centering
    \includegraphics[width=1\linewidth]{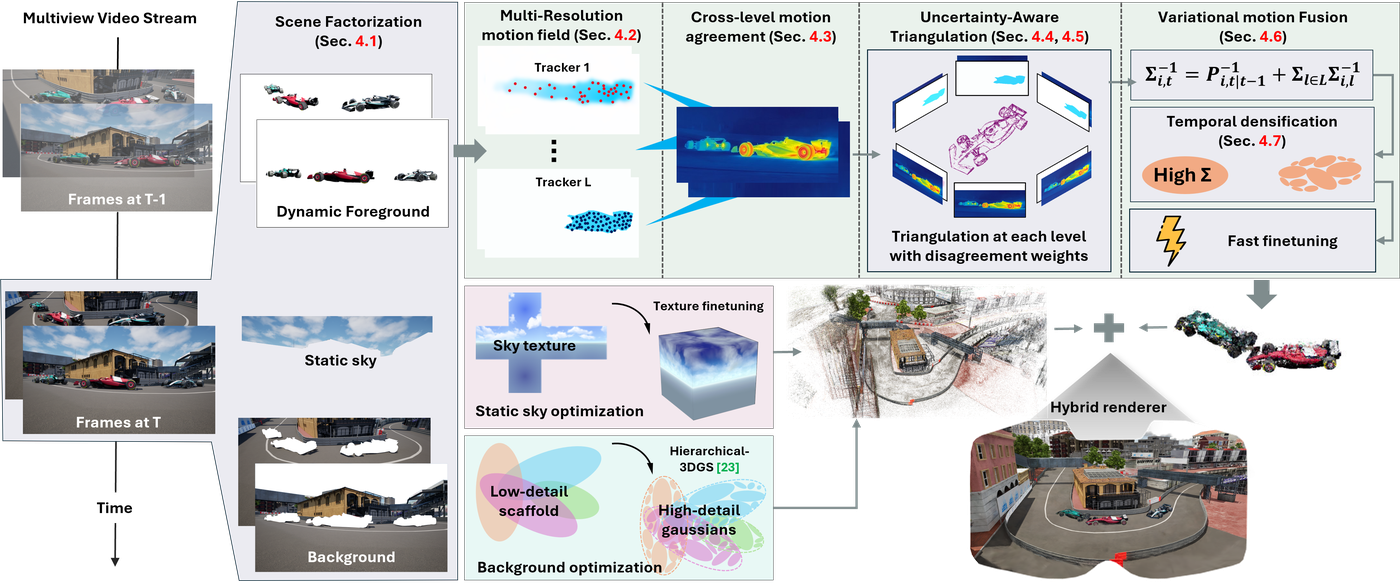}
    \caption{\textbf{Overview of FastFlowGS.} We decompose the scene into sky, background, and foreground (Sec.\,\ref{sec:decomp}).
        Sky and background are reconstructed via a skybox and Hierarchical 3DGS~\cite{kerbl2024hierarchical3dgaussianrepresentation}, respectively.
        Multi-resolution 2D motion fields are estimated per view (Sec.\,\ref{sec:tracks}) and scored for cross-level agreement (Sec.\,\ref{sec:crossconf}).
        Gaussians are associated with their contributing pixels (Sec.\,\ref{sec:assoc}) and each level's estimates are lifted to 3D via triangulation (Sec.\,\ref{sec:tri}).
        An uncertainty-based fusion merges all levels with a Kalman-based temporal prior into a single position estimate per Gaussian (Sec.\,\ref{sec:fusion}), which seeds fast per-frame optimization and densification (Sec.\,\ref{sec:opt}).
        The three layers are composited at render time. Refer Alg. 1 in the supplemental for pseudocode.
    }
    \label{fig:method_fig}
\end{figure}

We find that no single correspondence type is reliable across all surfaces and motions. Sparse features fail on textureless regions, dense flow degrades near occlusions, and any level may appear equally plausible where little motion occurs. FastFlowGS resolves this by measuring cross-level agreement as a proxy for reliability, lifting each level to 3D with a covariance estimate, and fusing all sources through a single information-form update where each contributes in proportion to its geometric precision.
 
\subsection{Scene Factorization for Dynamic Reconstruction}
\label{sec:decomp}
 
We represent the scene as three independently fitted layers composited at render time: static background, dynamic foreground, and sky.
The background uses Hierarchical 3DGS~\cite{kerbl2024hierarchical3dgaussianrepresentation}, optimized once at $t\!=\!0$. For subsequent frames, geometry is frozen and only SHs (color) are updated to account for illumination drift~\cite{10.1145/3731214}.
Dynamic Gaussians are initialized at $t\!=\!0$ by running 3DGS on a hemispherical rig around the cars with scale regularization~\cite{fang2024minisplattingrepresentingscenesconstrained, taming3dgs} For $t\!>\!0$, they are identified by projecting onto the previous frame's segmentation masks and repositioned via the fusion of Sec.~\ref{sec:fusion}.
The sky violates finite-depth assumptions and is rendered as a low-resolution skybox~\cite{yan2024street} composited behind other layers.

\subsection{Multi-Resolution Motion Field Estimation}
\label{sec:tracks}
 
FastFlowGS estimates per-view 2D motion fields by integrating correspondence sources at different spatial scales.
An arbitrary number of trackers with varying levels of sparsity comprise the set $L$; since triangulation (Sec.~\ref{sec:tri}) runs independently per level, all levels execute in parallel.
Sparse tracks are keypoint correspondences~\cite{lindenberger2023lightglue}, medium are semi-dense point tracks~\cite{karaev2024cotracker3,doersch2024bootstap}, and dense are per-pixel optical flow~\cite{teed2020raftrecurrentallpairsfield, moing2024denseopticaltrackingconnecting}.

\paragraph{Voronoi densification.}
Sparse and medium tracks are densified to full resolution via Gaussian-weighted Voronoi interpolation~\cite{moing2024denseopticaltrackingconnecting}.
For each pixel $\mathbf{p}=(u,v)$, the $K=8$ nearest track observations are interpolated with weights $w_i \propto \exp(-d_i^2 / 2\sigma^2)$ ($\sigma=0.05$), where $d_i$ is the distance to Voronoi center $i$. The nearest-neighbor kernel value defines confidence $\mathbf{C}^{\text{vor}}(\mathbf{p})$, which down-weights pixels far from any track during triangulation. More details are in Sec. 1.4 and Alg. 2 of the supplemental.

\subsection{Cross-Level Motion Agreement}
\label{sec:crossconf}
 
Since each correspondence tracker has its limitations, rather than selecting a preferred level, we estimate reliability by measuring agreement between them: pixels where motion estimates disagree are downweighted regardless of which estimate is correct.
 
Formally, let $\textbf{F} = \{\mathbf{F}_i\}$ where $i \in L$, $\textbf{F}_i \in \mathbb{R}^{V \times H \times W \times 2}$ denote the densified flow fields for all tracking levels over $V$ views.
Each level's per-pixel confidence is:
\begin{equation}
\textbf{C}_i = \psi(\textbf{C}^{\text{vor}}_i, \textbf{F}) = \textbf{C}_i^{\text{vor}}  \odot \exp\!\left(- \frac{1}{2} \cdot \tfrac{\sum_{k \neq i,\, k \in L} D_{ik}^2}{ (\sigma \bar{\textbf{F}})^2}\right)
\label{eq:conf}
\end{equation}
\begin{equation}
\textbf{C}_s = \psi(\textbf{C}^{\text{vor}}_s, \textbf{F}),\quad \textbf{C}_m = \psi(\textbf{C}^{\text{vor}}_m, \textbf{F}),\quad \textbf{C}_d = \psi(\textbf{C}^{\text{vor}}_d, \textbf{F}),
\label{eq:confd}
\end{equation}
where $D_{ab} = \|\mathbf{f}_a - \mathbf{f}_b\|_2$ for $\{a,b\}\in L$, and $\bar{\textbf{F}} = \tfrac{1}{|L|}\|\sum_{i \in L} \textbf{F}_i \|_2$ is the mean flow magnitude, clamped below at 1~px to avoid amplifying static regions.
Eqs.~\eqref{eq:conf}--\eqref{eq:confd} encode two complementary principles: a track far from any observation is unreliable (Voronoi term), and a track that disagrees with the other levels is unreliable regardless of its density (cross-level term).

\begin{wrapfigure}[10]{r}{4.5cm}
    \includegraphics[width=\linewidth]{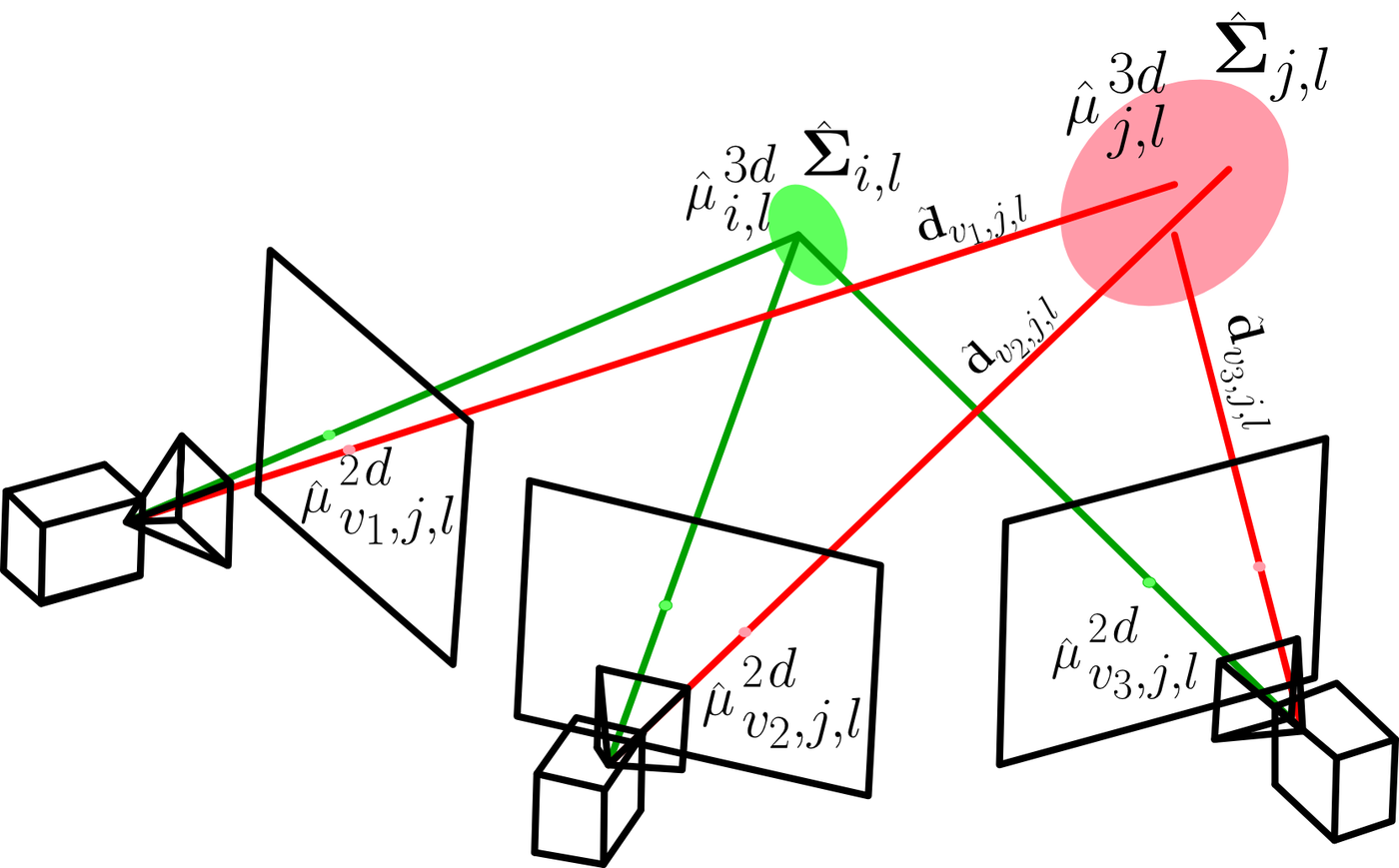}
    \caption{We use gaussian-pixel matching to estimate 3D flow of each level}
    \label{fig:dlt}
\end{wrapfigure}

\subsection{Gaussian-to-Pixel Association}
\label{sec:assoc}
 
To associate each Gaussian with its contributing pixels, we record the top-$K$ Gaussian indices and alpha-blending weights $\alpha_{v,\mathbf{p},k}$ for every pixel $\mathbf{p}$ and view $v\in V$, yielding weighted observations $\{i, v, \mathbf{p}, \alpha_{v,p,i}\}$ used to locate each Gaussian in image space and weight its triangulation contribution (Sec.~\ref{sec:tri}).
 
Each observation is weighted by:
\begin{equation}
w_{v, \textbf{p}, i, l} = \alpha_{v, \textbf{p},i} \cdot \textbf{C}_l(v, \mathbf{p}) \cdot \|\mathbf{F}_l(v, \textbf{p})\|_2.
\label{eq:weights}
\end{equation}
The product structure is such that, a Gaussian contributes to a position estimate only when it renders prominently ($\alpha$), the correspondence is geometrically consistent ($\mathbf{C}_l$), and there is actual motion to track ($\|\mathbf{F}_l\|$).
The projected mean of each \\ Gaussian $i$ is then updated per view $v$ and tracker $l$ as \\
$\hat{\mu}^{2d}_{v, i, l} = \mu^{2d}_{v, i} + \sum_\textbf{p} \alpha_{v, \textbf{p}, i} \textbf{F}_l(v, \textbf{p})$, to be used in triangulation.

\subsection{Uncertainty-Aware Multi-View Triangulation}
\label{sec:tri}

We estimate updated Gaussian positions $\hat{\mu}_{i,l}^{3d}$ as the weighted intersection of viewing rays, discarding views whose geometry is ambiguous.

\paragraph{Weighted DLT triangulation.}
Given projected observations $\{\hat{\mu}^{2d}_{v,i,l}\}$, we estimate each 3D position by minimizing the weighted reprojection error.
For unit ray direction $\hat{\mathbf{d}}_{v,i,l}$ lifted from $\hat{\mu}^{2d}_{v,i,l}$ via camera intrinsics and extrinsics, the DLT constraint is:
\begin{equation}
\overbrace{\left(\sum_\textbf{p} w_{v, \textbf{p}, i, l} \right)}^{\eta_{v,i,l}} \cdot  (\mathbf{I}_{3\times3} - \hat{\mathbf{d}}_{v,i, l}\hat{\mathbf{d}}_{v, i, l}^\top)(\hat{\mu}^{3d}_{i, l} - \mathbf{o}_{v}) = \mathbf{0}_{3\times1},
\label{eq:dlt}
\end{equation}
where $\mathbf{o}_v$ is the camera center.
For Gaussian $i$ visible from $N_i$ views, these constraints form a weighted least-squares system $\|\mathbf{W}[\mathbf{A}\hat{\mu}_{i,l}^{3d} - \mathbf{b}]\|^2_2 = \mathbf{0}$, where
$\mathbf{W} = \operatorname{diag}(\eta_1\mathbf{I}_{3\times3}, \ldots, \eta_{N_i}\mathbf{I}_{3\times3})$,
\[
\mathbf{A}
=
\begin{bmatrix}
\mathbf{I} - \hat{\mathbf{d}}_1 \hat{\mathbf{d}}_1^\top \\
\vdots \\
\mathbf{I} - \hat{\mathbf{d}}_{N_i} \hat{\mathbf{d}}_{N_i}^\top
\end{bmatrix},
\qquad
\mathbf{b}
=
\begin{bmatrix}
(\mathbf{I} - \hat{\mathbf{d}}_1 \hat{\mathbf{d}}_1^\top)\mathbf{o}_1 \\
\vdots \\
(\mathbf{I} - \hat{\mathbf{d}}_{N_i} \hat{\mathbf{d}}_{N_i}^\top)\mathbf{o}_{N_i}
\end{bmatrix},
\]

with closed-form solution $\hat{\mu}_{i,l}^{3d} = (\mathbf{A}^\top\mathbf{W}\mathbf{A})^{-1}\mathbf{A}^\top\mathbf{W}\mathbf{b}$, solved independently per tracker $l$. Fig. \ref{eq:dlt} demonstrates this.

Views whose motion direction is nearly collinear with the viewing ray (angle $<25^\circ$) are discarded: such configurations contribute little depth information and cause depth changes to appear as lateral motion. Geometrically, $\hat{\mu}^{3d}_{i,l}$ is the point minimizing perpendicular distance to all rays $\hat{\mathbf{d}}_{v,i,l}$.

\paragraph{Covariance estimation.}
Each triangulation yields not just a position but a measure of geometric reliability.
Under isotropic observation noise, the estimator covariance is:
\begin{equation}
\hat{\boldsymbol{\Sigma}}_{i,l} = \hat{\sigma}_{i,l}^{2} \cdot (\mathbf{A}^\top \mathbf{W} \mathbf{A})^{-1},
\qquad
\hat{\sigma}_{i,l}^{2} = \frac{\| \mathbf{W}[\mathbf{A}\hat{\mu}^{3d}_{i, l} - \mathbf{b}]\|_2^2}{N_i - 3},
\label{eq:cov}
\end{equation}
where $N_i - 3$ is the residual degrees of freedom.
Large eigenvalues of $\hat{\boldsymbol{\Sigma}}_{i,l}$ reflect poor geometry (near-parallel rays, few views, or inconsistent flow), while small eigenvalues reflect a reliable estimate.
We pass $\hat{\boldsymbol{\Sigma}}_{i,l}$ directly as measurement uncertainty into the fusion below.

\subsection{Variational Temporal Motion Fusion}
\label{sec:fusion}
 
We fuse the per-level triangulation estimates with a temporal prior through a single Kalman-style update, where each source contributes in proportion to its geometric precision.

\paragraph{Energy formulation.}
We seek the 3D position $\mu^{3d}_{i,t}$ of Gaussian $i$ at frame $t$ minimizing:
\begin{equation}
\mathcal{E}\!\left(\{\mu^{3d}_{i,\tau}\}_{\tau=1}^{t}\right)
= \sum_{\tau=1}^{t} \|\mu^{3d}_{i,\tau} - \mu^{3d}_{i,\tau-1}\|^{2}_{\mathbf{Q}_\tau^{-1}}
+ \sum_{l \in L} \|\mu^{3d}_{i,t} - \hat{\mu}_{i,l}^{3d}\|^{2}_{\hat{\boldsymbol{\Sigma}}_{i, l}^{-1}},
\label{eq:energy}
\end{equation}
where $\|\mathbf{v}\|_{\mathbf{M}}^2 = \mathbf{v}^\top \mathbf{M} \mathbf{v}$.
The first sum penalizes temporal deviations weighted by inverse process noise $\mathbf{Q}_\tau^{-1}$; the second penalizes disagreement with each tracker weighted by its triangulation precision $\hat{\boldsymbol{\Sigma}}_{i,l}^{-1}$.

\paragraph{Closed-form minimizer.}
Minimizing $\mathcal{E}$ with respect to $\mu^{3d}_{i,t}$ yields the linear system:
\begin{equation}
\left[\mathbf{P}_{i,t|t-1}^{-1} + \sum_{l \in L} \hat{\boldsymbol{\Sigma}}_{i, l}^{-1}\right]\mu^{3d}_{i,t}
= \mathbf{P}_{i,t|t-1}^{-1}\mu^{3d}_{i,t-1} + \sum_{l \in L} \hat{\boldsymbol{\Sigma}}_{i, l}^{-1}\hat{\mu}_{i,l}^{3d},
\label{eq:normal}
\end{equation}
where $\mathbf{P}_{i,t|t-1} = \mathbf{P}_{i,t-1} + \mathbf{Q}_t$ is the predicted covariance from the previous posterior.
This is the information-form Kalman update; the posterior mean and covariance are:
\begin{equation}
\boldsymbol{\mu}_{i,t} = \boldsymbol{\Sigma}_{i,t}\!\left[\mathbf{P}_{i,t|t-1}^{-1}\mu^{3d}_{i,t-1} + \sum_{l \in L} \hat{\boldsymbol{\Sigma}}_{i, l}^{-1}\hat{\mu}_{i,l}^{3d}\right],
\qquad
\boldsymbol{\Sigma}_{i,t}^{-1} = \mathbf{P}_{i,t|t-1}^{-1} + \sum_{l \in L} \hat{\boldsymbol{\Sigma}}_{i, l}^{-1}.
\end{equation}
Multi-resolution fusion and temporal propagation are therefore not independent design choices but two terms in the same objective, competing for influence in the precision accumulator $\boldsymbol{\Sigma}_{i,t}^{-1}$ purely on the basis of geometric reliability.
Gaussians for which all levels fail to triangulate receive only the temporal term $\mathbf{P}_{i,t|t-1}^{-1}$, anchoring them to their previous position with covariance that grows with each unanswered frame.
 
We model dynamics as a random walk $\mu^{3d}_{i,t} = \mu^{3d}_{i,t-1} + w_t$,
$w_t \sim \mathcal{N}(0,\, q_t^2\mathbf{I})$, where $q_t$ is the median 3D displacement of successfully triangulated Gaussians at frame $t$, bootstrapped from the previous frame at $t=1$.
This ties process noise directly to observed scene dynamics: fast frames relax the temporal prior and cede influence to the tracker measurements; near-static frames tighten it.
 
\paragraph{Optimization.}
\label{sec:opt}
Following~\cite{10.1145/3731214}, we optimize dynamic and static layers in parallel using a hybrid renderer that jointly rasterizes Gaussians and the skybox.
Dynamic Gaussians minimize a foreground-masked $\mathcal{L}_1$-SSIM loss over all parameters; background Gaussians update only their spherical harmonics to preserve geometric stability.
Because the variational initialization places Gaussians near their correct positions, convergence requires very few iterations.

\paragraph{Uncertainty-driven densification.}
Rather than waiting for rendering error to identify under-reconstructed regions after many gradient steps, we split or clone Gaussians with high triangulation uncertainty before optimization begins, front-loading capacity where multi-view geometry is weakest.
To the best of our knowledge, no prior streaming Gaussian method drives densification temporally (from geometric uncertainty) rather than rendering error.
\clearpage

\section{Experiments}
\label{sec:experiments}

We evaluate FastFlowGS on two datasets testing complementary regimes.
CMU-Panoptic~\cite{Joo_2017_TPAMI} is a standard indoor multi-view benchmark, included to confirm that solving the large-displacement problem does not degrade general indoor performance.
Monaco4D tests whether the approach holds when the assumptions underlying every prior method no longer do.
For CMU-Panoptic, we select six subsequences from \textit{161029\_sports1} and evaluate at the native frame rate and under $5\times$ subsampling to simulate larger inter-frame motion.

\paragraph{Baselines.}
We compare against six online reconstruction methods with public implementations: Dynamic-3DGS~\cite{dynamic3dgs}, 3DGStream~\cite{sun20243dgstreamontheflytraining3d}, HiCoM~\cite{gao2024hicomhierarchicalcoherentmotion}, QUEEN~\cite{girish2024queen}, ReCon-GS~\cite{fu2025recongscontinuumpreservedgaussianstreaming}, and TrackerSplat~\cite{trackersplat}.
Instant Gaussian Stream~\cite{igs} exceeded GPU memory on both datasets.

We additionally report \textbf{3DGS-Base}, a purposefully simple control that initializes each frame's dynamic Gaussians from the previous frame's reconstruction and finetunes with standard 3DGS, without predicting positions before optimization.
Every other method, including FastFlowGS, explicitly estimates where Gaussians should move before optimization begins; 3DGS-Base isolates exactly what that prediction step is worth.
On CMU-Panoptic, we report FastFlowGS which uses 300 iterations per frame, and FastFlowGS-faster which uses 200 iterations, trading a modest quality reduction for higher throughput.

\paragraph{Metrics.}
We report PSNR and VMAF averaged across all frames and views.
Neither captures reconstruction quality at dynamic regions, so we additionally compute Masked PSNR (M-PSNR), restricting evaluation to dynamic foreground regions.
A method that reconstructs static pixels accurately but smears fast-moving objects can post high full-image PSNR while failing at the actual task.

Most methods initialize from a high-quality first frame, inflating sequence-average PSNR while concealing temporal decay.
We introduce \textbf{$\Delta$-PSNR}, the difference between first- and last-frame PSNR, which is high for degrading methods, near zero for stable ones.
For efficiency, we report mean per-frame training time using only the dynamic component of FastFlowGS, since static and dynamic branches run in parallel.
VMAF Efficiency ($\mathrm{VE} = \mathrm{VMAF}/\mathrm{Time}$) and PSNR Efficiency ($\mathrm{PE} = \mathrm{PSNR}/\mathrm{Time}$) capture quality per unit of compute.
All experiments run on an NVIDIA RTX A4500 at $960\times540$.
Following~\cite{dynamic3dgs,sun20243dgstreamontheflytraining3d,trackersplat,10.1145/3731214}, we report training time only, as preprocessing pipelines differ across methods. We provide an entire wall-clock time decomposition of each component (including preprocessing time) in Sec. 1.3 of the accompanying supplemental.

\subsection{CMU-Panoptic}
\label{subsec:panoptic}

All methods run end-to-end with default pipelines on the selected subsequences.

% figure all
\begin{figure*}[t]
	\centering
	\setlength{\tabcolsep}{0.1pt}
	\renewcommand{\arraystretch}{0}
    
	\newcommand{\spyimg}[4]{%
		\begin{tikzpicture}[spy using outlines={green,magnification=2,size=0.85cm, connect spies}]
			\node[anchor=south west,inner sep=0] at (0,0) {\includegraphics[width=#1]{#2}};
			\spy[every spy on node/.append style={thick}] on (#3) in node [left] at (#4);
		\end{tikzpicture}%
	}

        \newcommand{\spyimgfast}[4]{%
		\begin{tikzpicture}[spy using outlines={blue,magnification=2,size=0.85cm, connect spies}]
			\node[anchor=south west,inner sep=0] at (0,0) {\includegraphics[width=#1]{#2}};
			\spy[every spy on node/.append style={thick}] on (#3) in node [left] at (#4);
		\end{tikzpicture}%
	}
	
	\begin{tabular}{ccccc}

	Ground truth & D-3DGS & QUEEN & TrackerSplat & \textbf{Ours} \\
	
	\spyimg{0.2\linewidth}{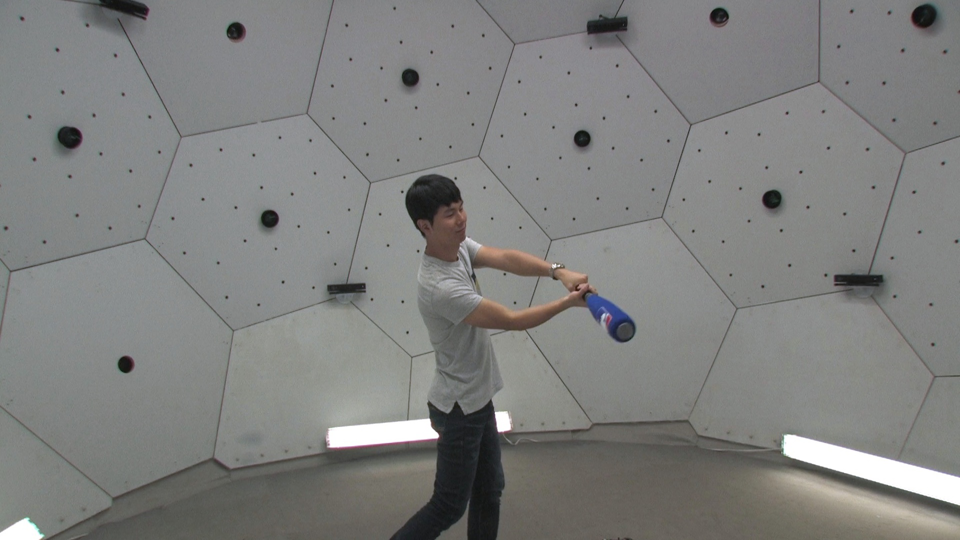}{2.05,0.8}{0.98,0.56} &
	\spyimg{0.2\linewidth}{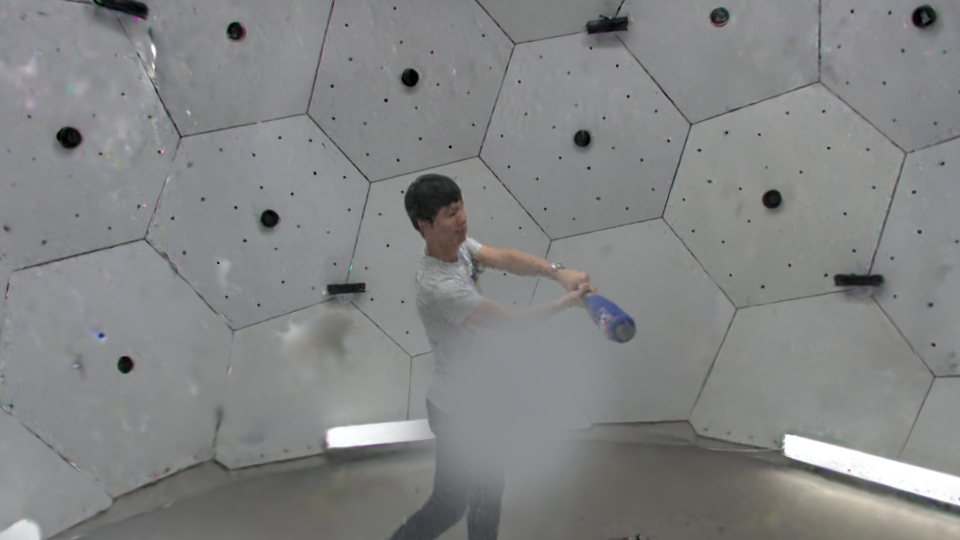}{2.05,0.8}{0.98,0.56} &
	\spyimg{0.2\linewidth}{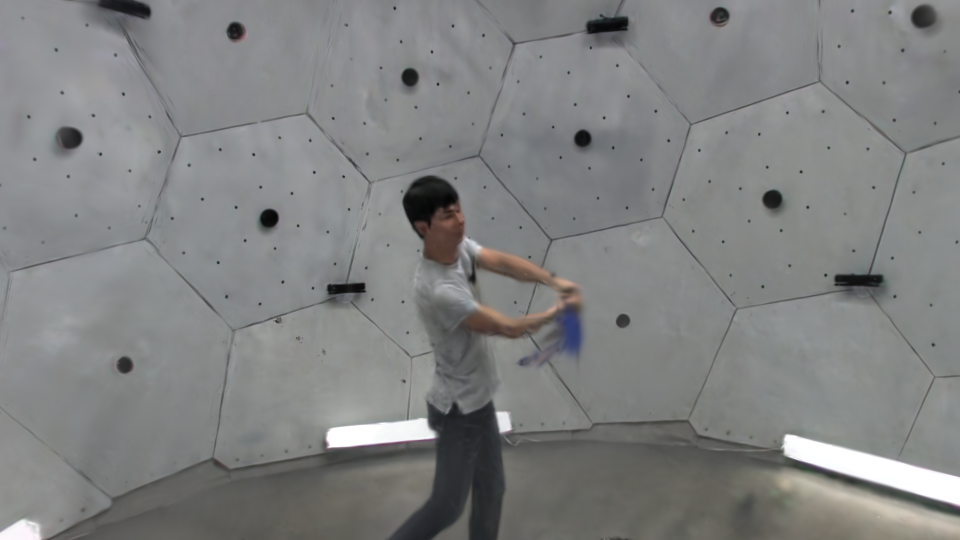}{2.05,0.8}{0.98,0.56} &
	\spyimg{0.2\linewidth}{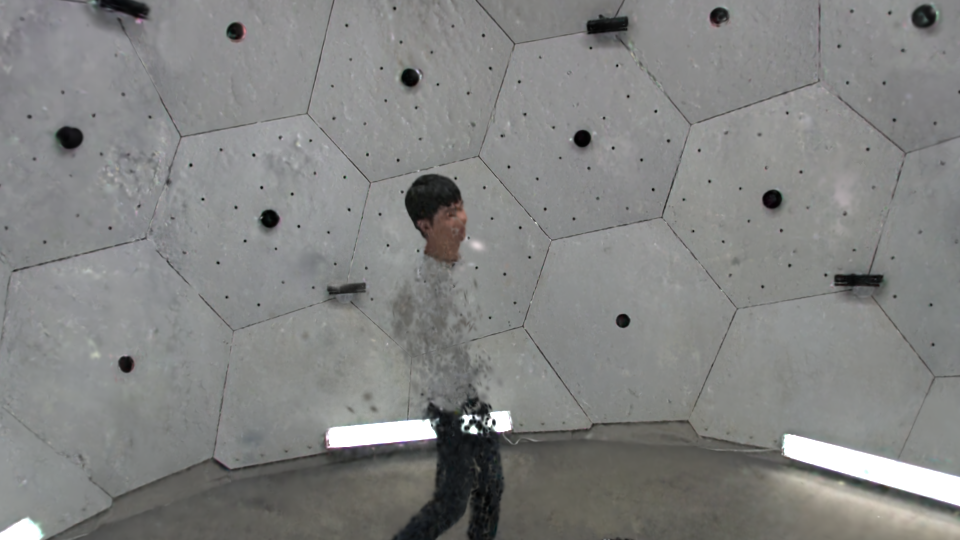}{2.05,0.8}{0.98,0.56} &
	\spyimg{0.2\linewidth}{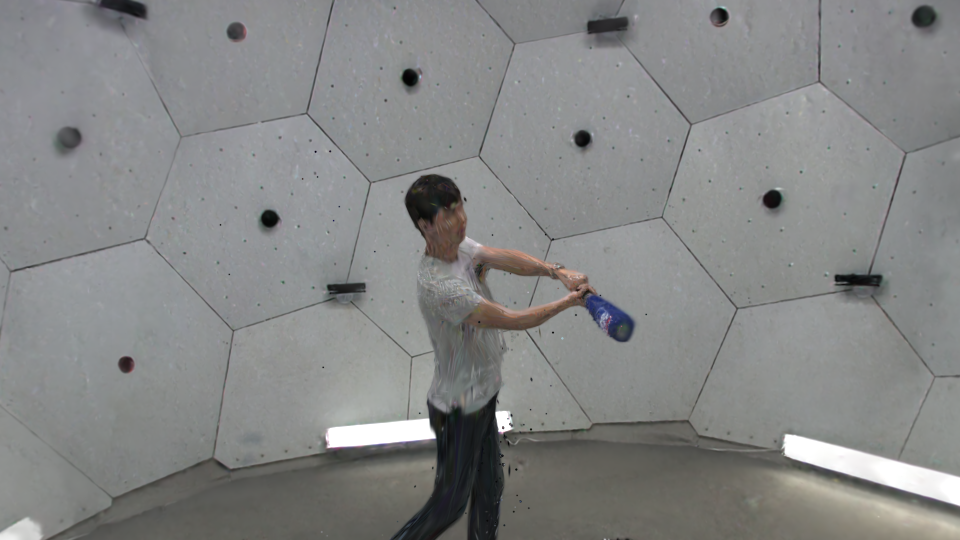}{2.05,0.8}{0.98,0.56} \\
	
	\spyimgfast{0.2\linewidth}{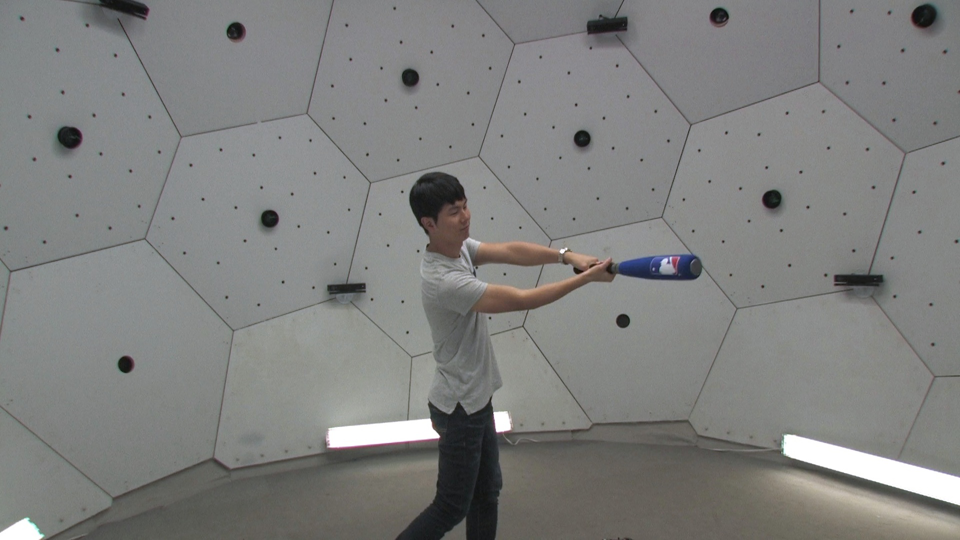}{2.25,0.9}{0.9,0.6} &
	\spyimgfast{0.2\linewidth}{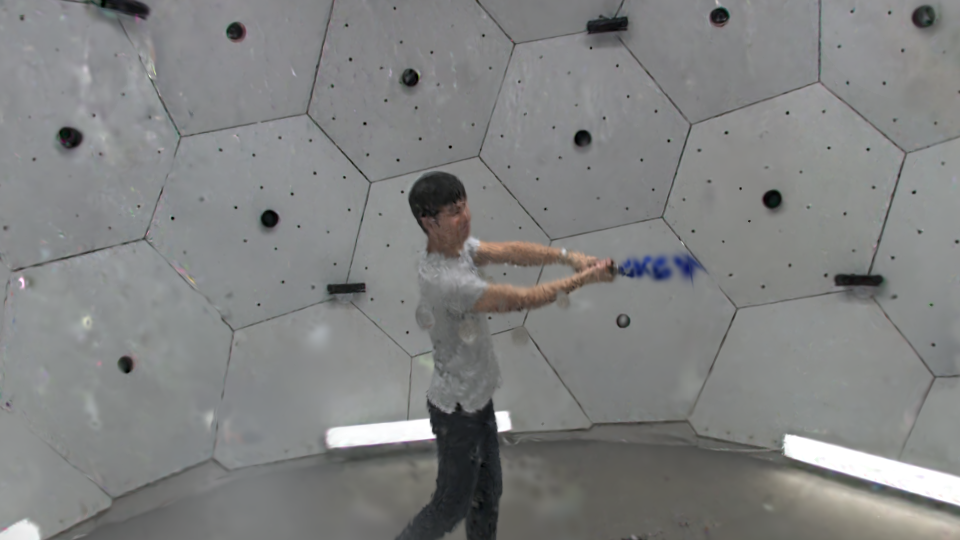}{2.25,0.9}{0.9,0.6} &
	\spyimgfast{0.2\linewidth}{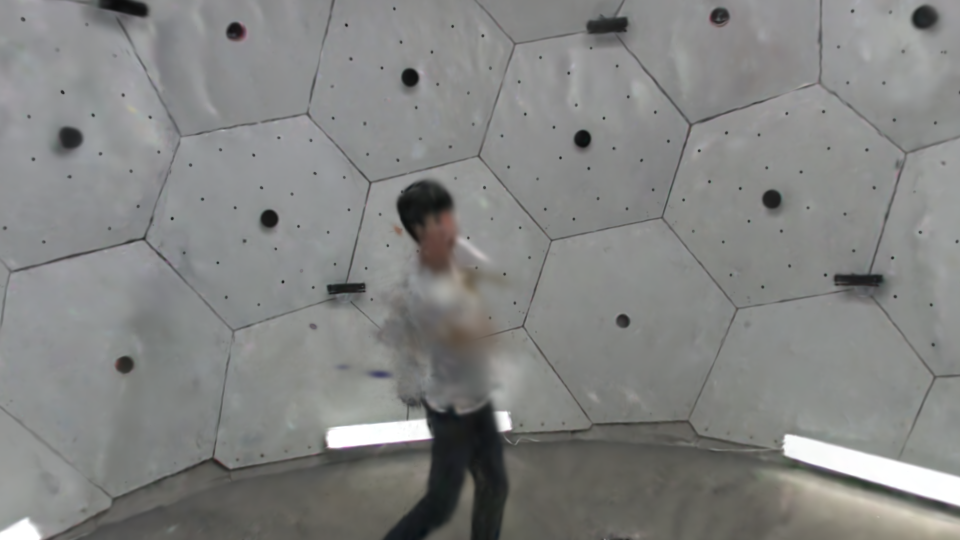}{2.25,0.9}{0.9,0.6} &
	\spyimgfast{0.2\linewidth}{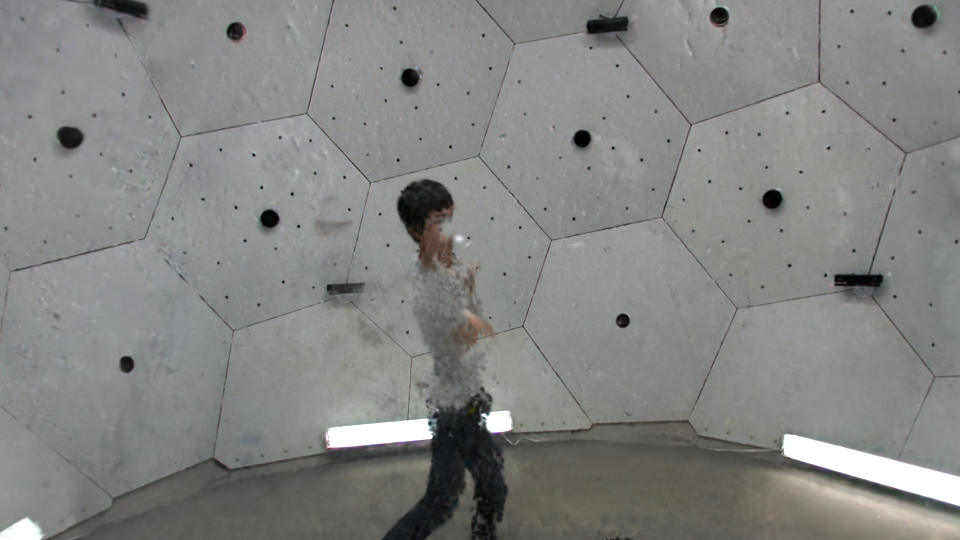}{2.25,0.9}{0.9,0.6} &
	\spyimgfast{0.2\linewidth}{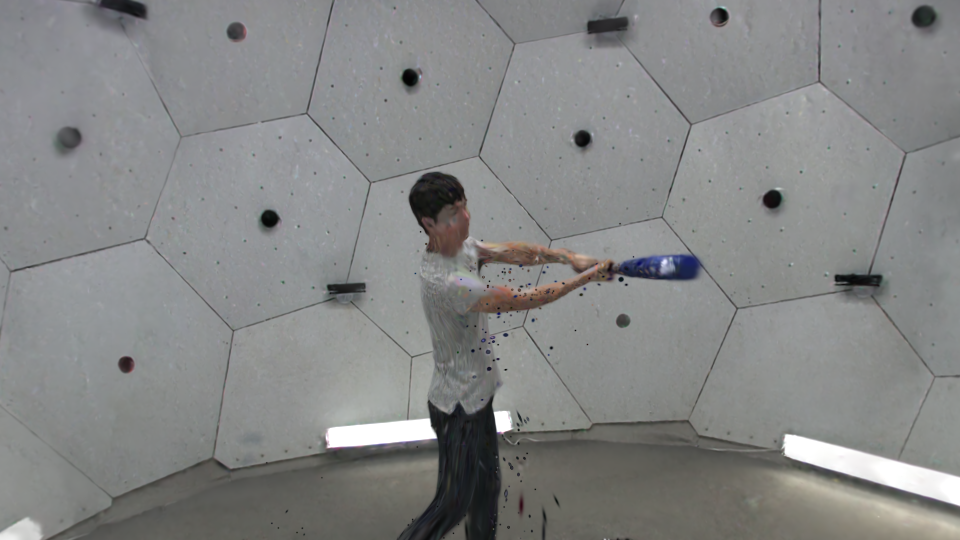}{2.25,0.9}{0.9,0.6} \\	
	
	\spyimg{0.2\linewidth}{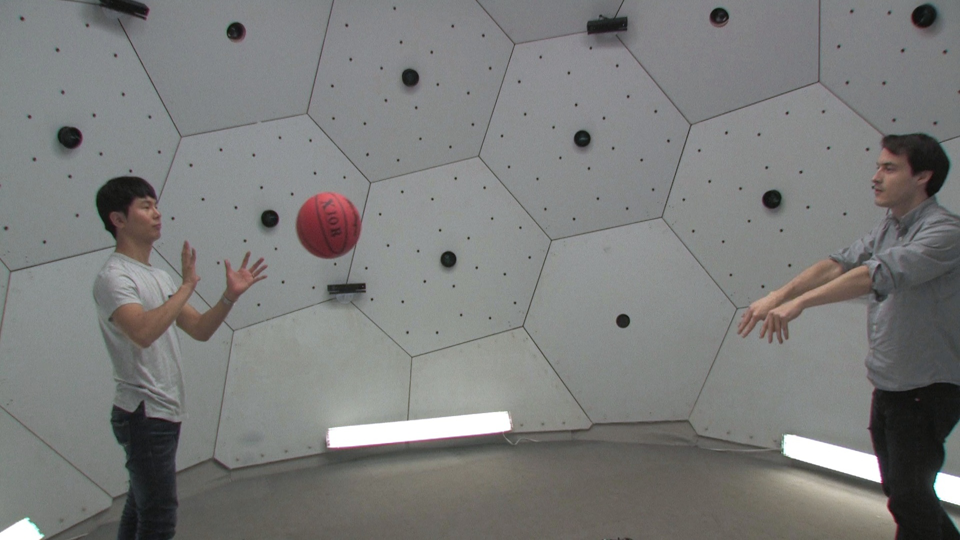}{1.15,1.1}{2.6,0.9} &
	\spyimg{0.2\linewidth}{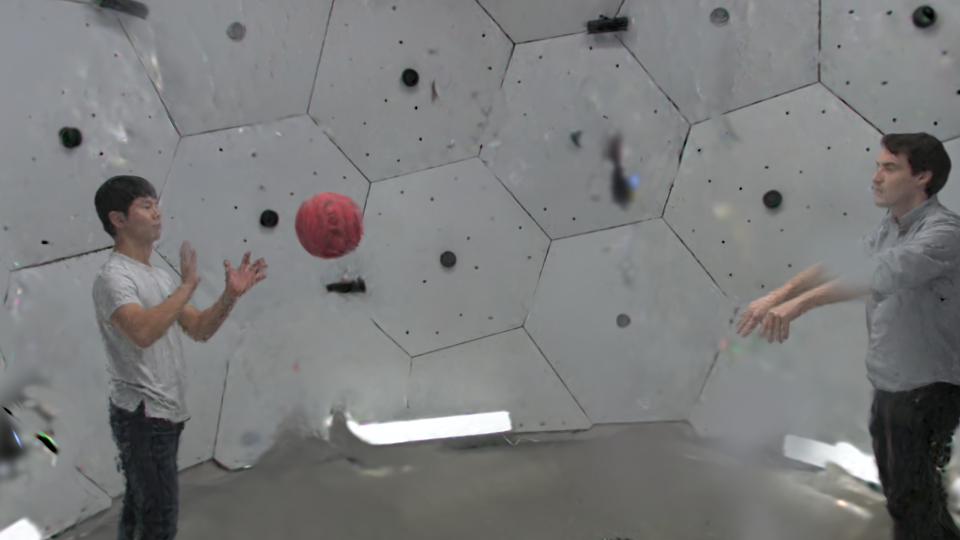}{1.15,1.1}{2.6,0.9} &
	\spyimg{0.2\linewidth}{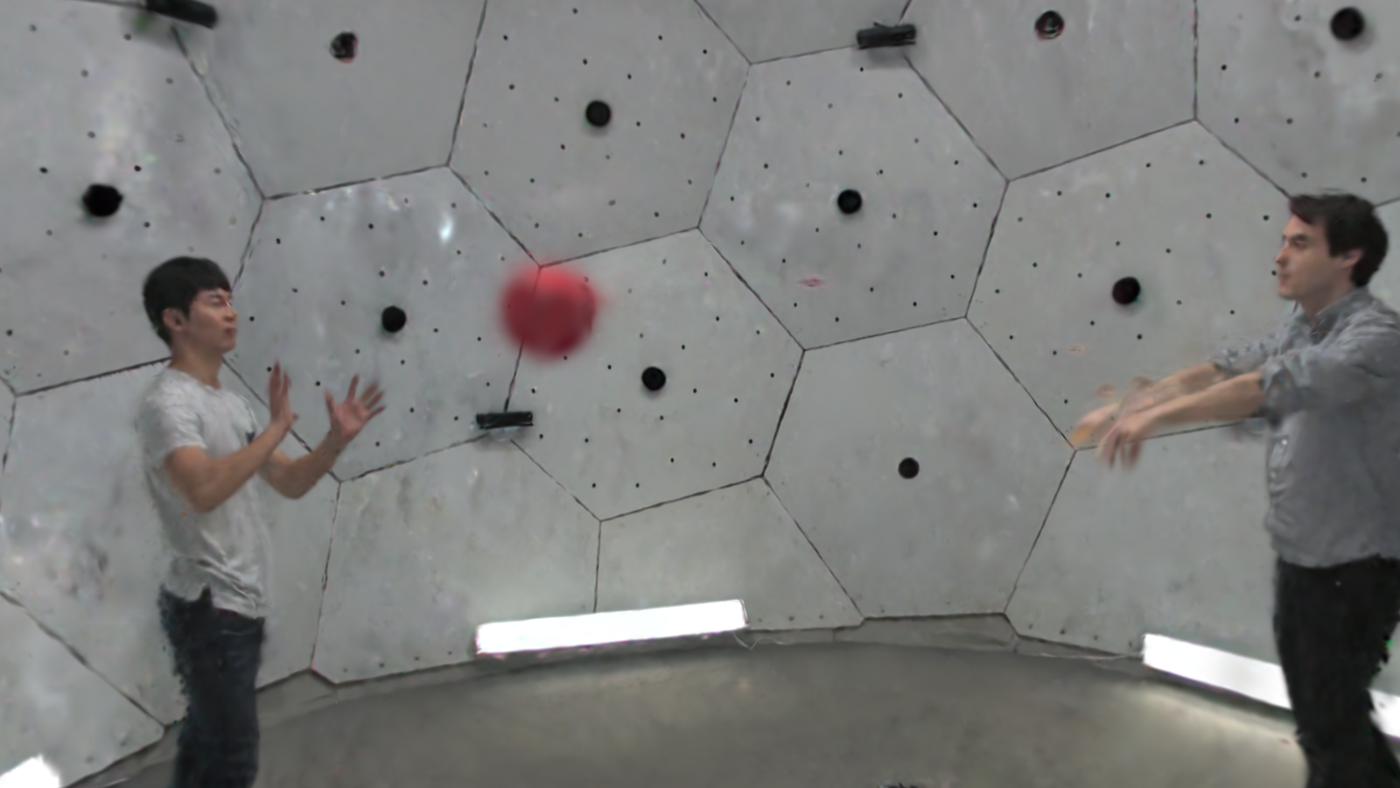}{1.15,1.1}{2.6,0.9} &
	\spyimg{0.2\linewidth}{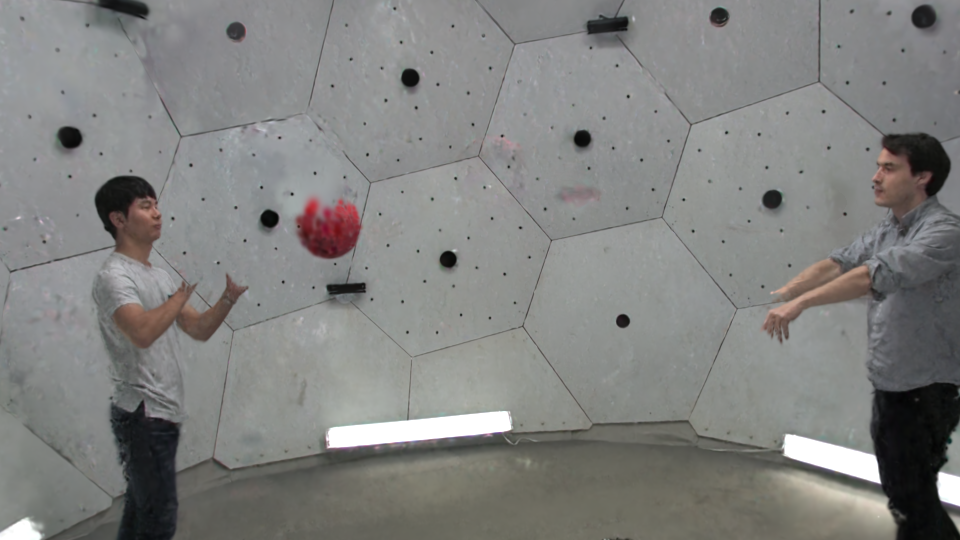}{1.15,1.1}{2.6,0.9} &
	\spyimg{0.2\linewidth}{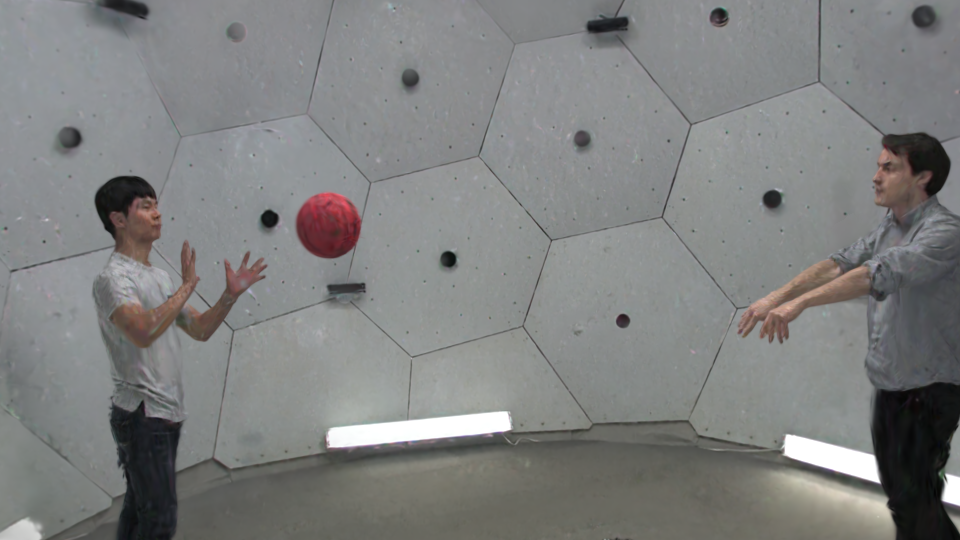}{1.15,1.1}{2.6,0.9} \\
	
	\spyimgfast{0.2\linewidth}{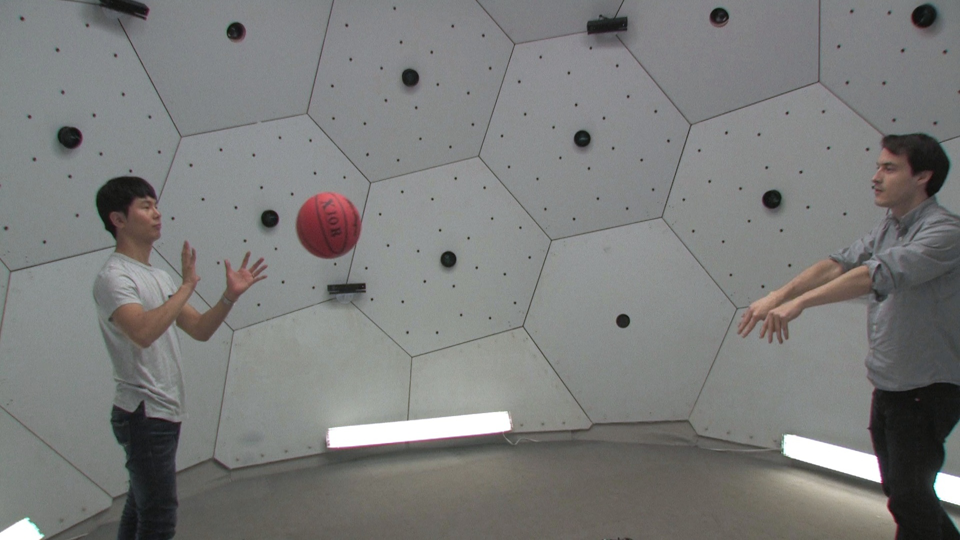}{1.15,1.1}{2.6,0.9} &
	\spyimgfast{0.2\linewidth}{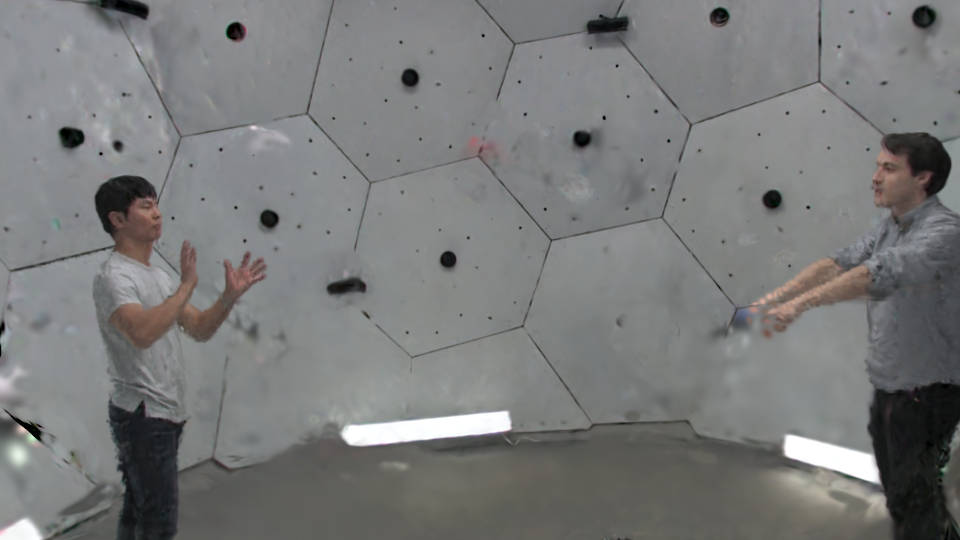}{1.15,1.1}{2.6,0.9} &
	\spyimgfast{0.2\linewidth}{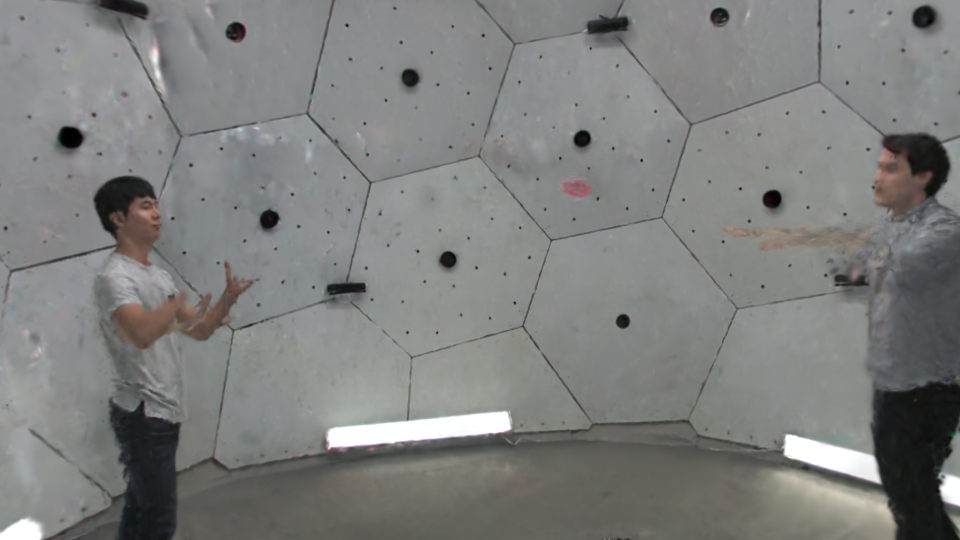}{1.15,1.1}{2.6,0.9} &
	\spyimgfast{0.2\linewidth}{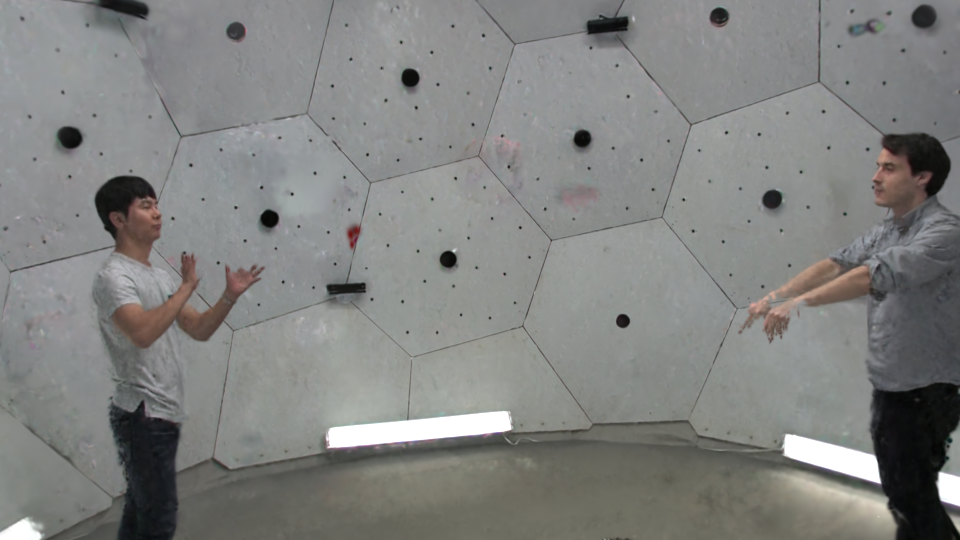}{1.15,1.1}{2.6,0.9} &
	\spyimgfast{0.2\linewidth}{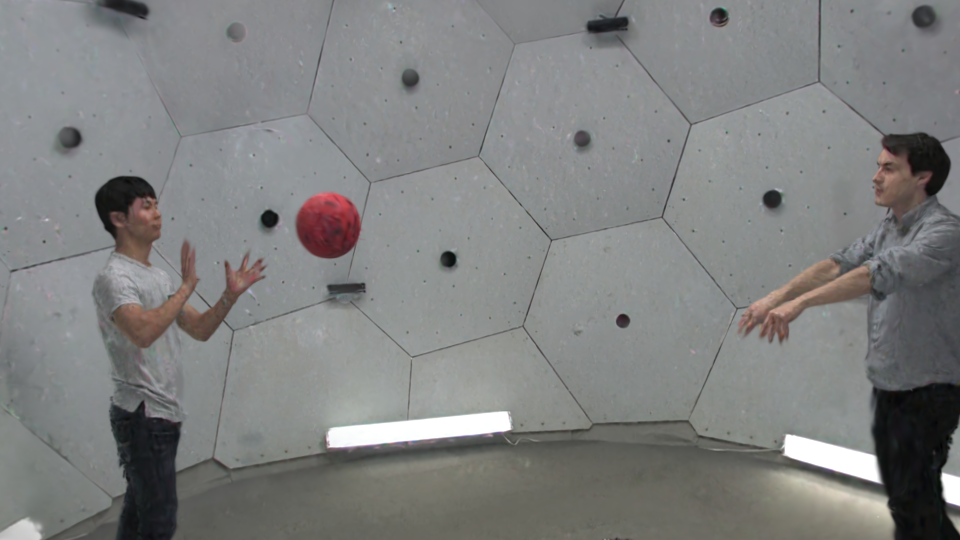}{1.15,1.1}{2.6,0.9} \\	

	\end{tabular}

    \caption{Qualitative Results for test cameras of CMU-Panoptic dataset. To avoid overcrowding, we show results only on the top 3 quality methods. \textcolor{green}{Green insets} mean the default frame rate and \textcolor{blue}{blue insets} mean 5 frame skip}
	\label{fig:results-panoptic}
    \vspace{-10pt}
    
\end{figure*}
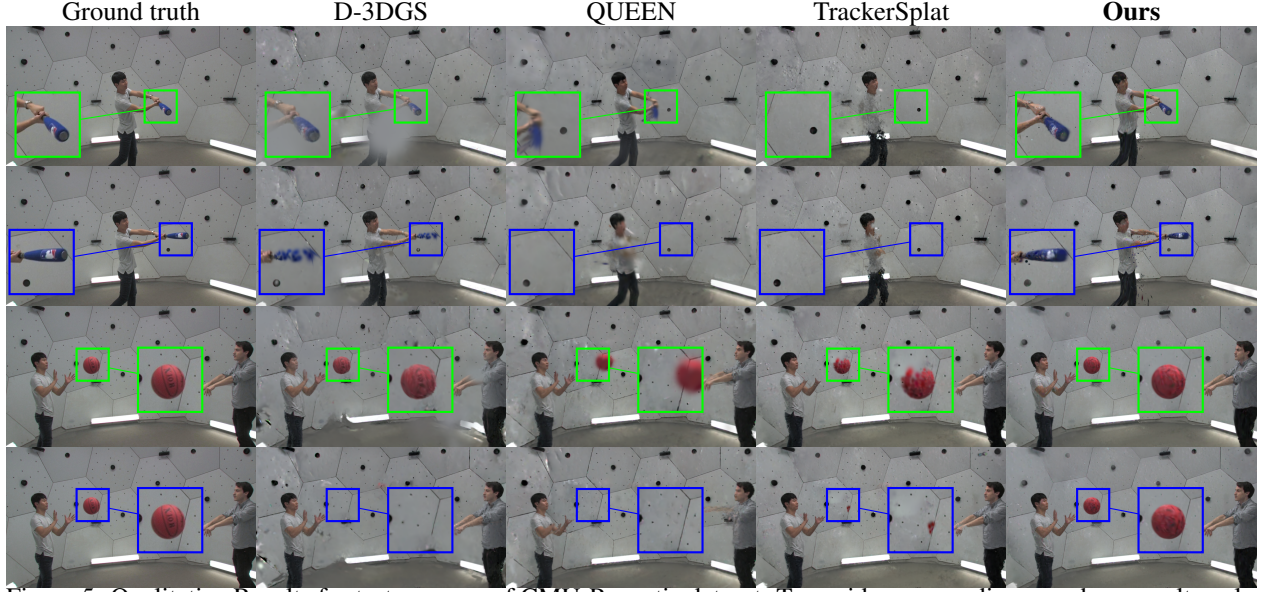

\setlength{\tabcolsep}{0.5pt}
\begin{table*}[b]
\centering
\caption{Quantitative results on CMU-Panoptic at normal frame rate (left) and $5\times$ frame skipping (right). \textit{Time} is in seconds.}
\label{tab:cmu_combined}
\scalebox{1}{
\begin{tabular}{@{}l|ccccc c|ccccc c@{}}
\toprule
& \multicolumn{6}{c|}{Normal Frame Rate} & \multicolumn{6}{c}{5$\times$ Faster} \\
\cmidrule(lr){2-7} \cmidrule(l){8-13}
& VMAF $\uparrow$ & PSNR $\uparrow$ & $\Delta$PSNR $\downarrow$ & MPSNR $\uparrow$ & Time $\downarrow$ & VE $\uparrow$
& VMAF $\uparrow$ & PSNR $\uparrow$ & $\Delta$PSNR $\downarrow$ & MPSNR $\uparrow$ & Time $\downarrow$ & VE $\uparrow$ \\ \midrule
D-3DGS
    & 45.45  & 25.981 & 0.397  & 24.070  & 120.54 & 0.38
    & 45.93  & 25.941 & 0.631  & 24.394  & 120.54 & 0.38 \\
3DGStream
    & 48.39  & 27.189 & \cellcolor[HTML]{FD8488}0.006 & 21.563 & 6.75 & 7.17
    & 32.95  & 21.68  & 3.71   & 13.20   & 6.75   & 4.88 \\
HiCoM
    & 53.93  & 27.033 & 2.349  & 23.784  & 5.89   & 9.16
    & 36.66  & 21.105 & 3.042  & 13.16   & 5.89   & 6.22 \\
QUEEN
    & 49.21  & 27.046 & 1.321  & 25.006  & 8.81   & 5.59
    & \cellcolor[HTML]{FD8488}51.16 & 26.448 & 2.188 & 25.39 & 8.81 & 5.81 \\
ReConGs
    & 42.94  & 25.475 & 1.484  & 22.665  & \cellcolor[HTML]{FEC088}3.82 & 11.24
    & 18.91  & 20.749 & 2.473  & 12.71   & \cellcolor[HTML]{FEC088}3.82 & 4.95 \\
Trackersplat
    & 46.24  & 24.719 & 3.825  & 22.536  & 14.14  & 3.27
    & 44.36  & 24.364 & 2.928  & 19.791  & 14.14  & 3.14 \\ \midrule
\textbf{Ours}
    & \cellcolor[HTML]{FD8488}60.71 & \cellcolor[HTML]{FD8488}27.893 & 0.133 & \cellcolor[HTML]{FD8488}26.141 & 4 & \cellcolor[HTML]{FEC088}15.18
    & \cellcolor[HTML]{FEC088}50.66 & \cellcolor[HTML]{FD8488}27.571 & \cellcolor[HTML]{FD8488}0.192 & \cellcolor[HTML]{FD8488}27.19 & 4 & \cellcolor[HTML]{FEC088}12.67 \\
\textbf{Ours-faster}
    & \cellcolor[HTML]{FEC088}55.79 & \cellcolor[HTML]{FEC088}27.262 & \cellcolor[HTML]{FEC088}0.147 & \cellcolor[HTML]{FEC088}25.722 & \cellcolor[HTML]{FD8488}3 & \cellcolor[HTML]{FD8488}27.9
    & 48.36  & \cellcolor[HTML]{FEC088}26.894 & \cellcolor[HTML]{FEC088}0.241 & \cellcolor[HTML]{FEC088}26.84 & \cellcolor[HTML]{FD8488}3 & \cellcolor[HTML]{FD8488}24.18 \\ \bottomrule
\end{tabular}
}
\end{table*}

FastFlowGS leads every baseline on every metric at both frame rates, but the more revealing result is temporal stability, not the absolute quality gap.
Most baselines accumulate more than 1\,dB of drift over a sequence; FastFlowGS holds $\Delta$-PSNR near zero at both frame rates (Tab.~\ref{tab:cmu_combined}, Fig.~\ref{fig:results-panoptic}).

The $5\times$ subsampled setting reveals which motion models hold under stress.
3DGStream, HiCoM and ReCon-GS suffer the sharpest drops, because they all use (neural or hierarchical) hash grids which impose a finest-cell resolution that caps representable displacements. 
FastFlowGS retains highest PSNR, M-PSNR, and lowest $\Delta$-PSNR under faster motion, confirming that multi-scale correspondence fusion does not require smooth inter-frame transitions to work.
FastFlowGS-faster maintains the highest VE across both settings at 3s per frame.

\begin{table}[]
\centering
\caption{Results on Monaco4D under the full-quality protocol. All streaming baselines fail on this setting (Fig.~\ref{fig:mask_failure}), hence only 3DGS-Base is reported. \textit{Time} is in seconds.}
\resizebox{\textwidth}{!}{
\begin{tabular}{@{}l|cccc|cccc|cccc@{}}
\toprule
\multicolumn{1}{c|}{}                   & \multicolumn{4}{c|}{Fairmont (5 cars)}                                                                                       & \multicolumn{4}{c|}{Rascasse (1 car)}                                                                                        & \multicolumn{4}{c}{Main Straight (3 cars)}                                                                                   \\ \cmidrule(l){2-13} 
\multicolumn{1}{c|}{\multirow{-2}{*}{}} & VMAF$\uparrow$ & PSNR $\uparrow$ & MPSNR $\uparrow$ & Time $\downarrow$ & VMAF $\uparrow$ & PSNR$\uparrow$ & MPSNR$\uparrow$ & Time$\downarrow$ & VMAF$\uparrow$ & PSNR$\uparrow$ & MPSNR$\uparrow$ & Time$\downarrow$ \\ \midrule
Base                                    & 41.87                         & 21.06                         & 16.4                          & 11.82                        & 39.87                         & 21.26                         & 15.66                         & 6.42                         & 55.89                         & 21.43                         & 15.99                         & 8.51                         \\
\textbf{Ours}                           & \cellcolor[HTML]{FD8488}43.93 & \cellcolor[HTML]{FD8488}21.53 & \cellcolor[HTML]{FD8488}18.88 & \cellcolor[HTML]{FD8488}8.47 & \cellcolor[HTML]{FD8488}40.81 & \cellcolor[HTML]{FD8488}21.48 & \cellcolor[HTML]{FD8488}17.47 & \cellcolor[HTML]{FD8488}6.12 & \cellcolor[HTML]{FD8488}56.24 & \cellcolor[HTML]{FD8488}22.22 & \cellcolor[HTML]{FD8488}18.96 & \cellcolor[HTML]{FD8488}7.11 \\ \bottomrule
\end{tabular}
}
\label{tab:monaco}
\end{table}

\subsection{Monaco4D}
\label{subsec:monaco}

Monaco4D is not simply a harder version of CMU-Panoptic, because it tests a regime where the assumptions underlying existing methods break rather than bend.
Both evaluation protocols below share the same scene factorization: the static background uses hierarchical 3DGS~\cite{lin2024vastgaussian}, the skybox is rendered from a cubemap~\cite{yan2024street}, and the dynamic foreground is optimized per frame.
This decomposition is necessary because end-to-end optimization at full resolution exceeds the GPU memory of every method we evaluate.
All methods use the same static and sky layers, so the comparison isolates each method's dynamic reconstruction.

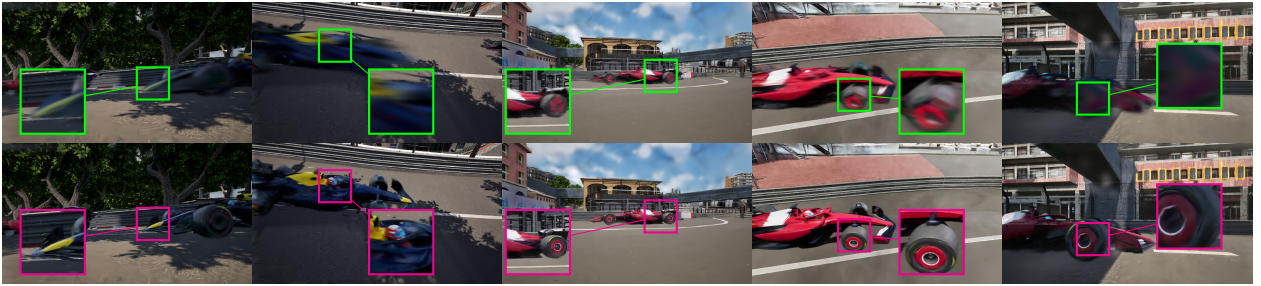
\begin{figure*}[b]
	\centering
	\setlength{\tabcolsep}{0.1pt}
	\renewcommand{\arraystretch}{0}
    
	\newcommand{\spyimg}[4]{%
		\begin{tikzpicture}[spy using outlines={green,magnification=2,size=0.85cm, connect spies}]
			\node[anchor=south west,inner sep=0] at (0,0) {\includegraphics[width=#1]{#2}};
			\spy[every spy on node/.append style={thick}] on (#3) in node [left] at (#4);
		\end{tikzpicture}%
	}

        \newcommand{\spyimgfast}[4]{%
		\begin{tikzpicture}[spy using outlines={magenta,magnification=2,size=0.85cm, connect spies}]
			\node[anchor=south west,inner sep=0] at (0,0) {\includegraphics[width=#1]{#2}};
			\spy[every spy on node/.append style={thick}] on (#3) in node [left] at (#4);
		\end{tikzpicture}%
	}
	
	\begin{tabular}{ccccc}

	\spyimg{0.2\linewidth}{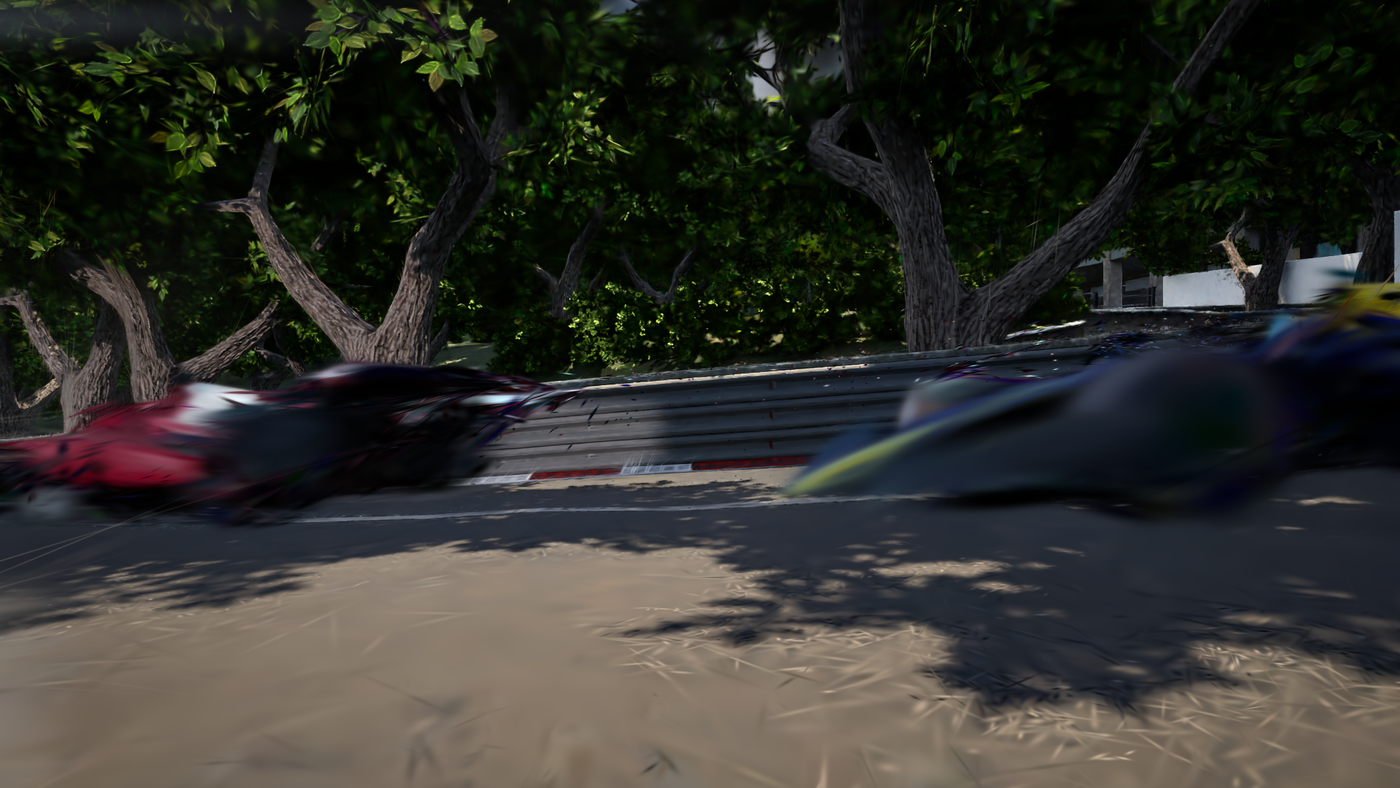}{2.0,0.8}{1.1,0.56} &
	\spyimg{0.2\linewidth}{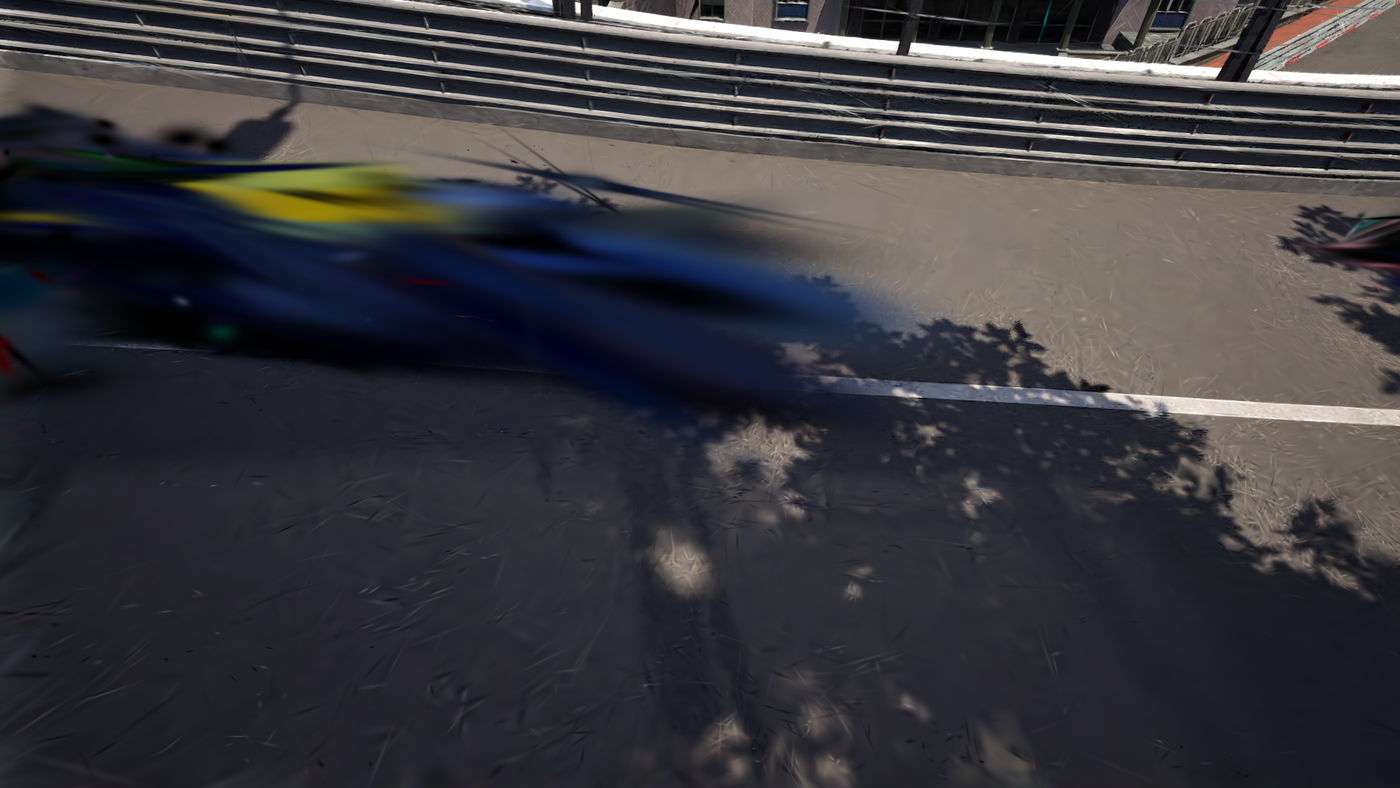}{1.1,1.3}{2.4,0.56} &
	\spyimg{0.2\linewidth}{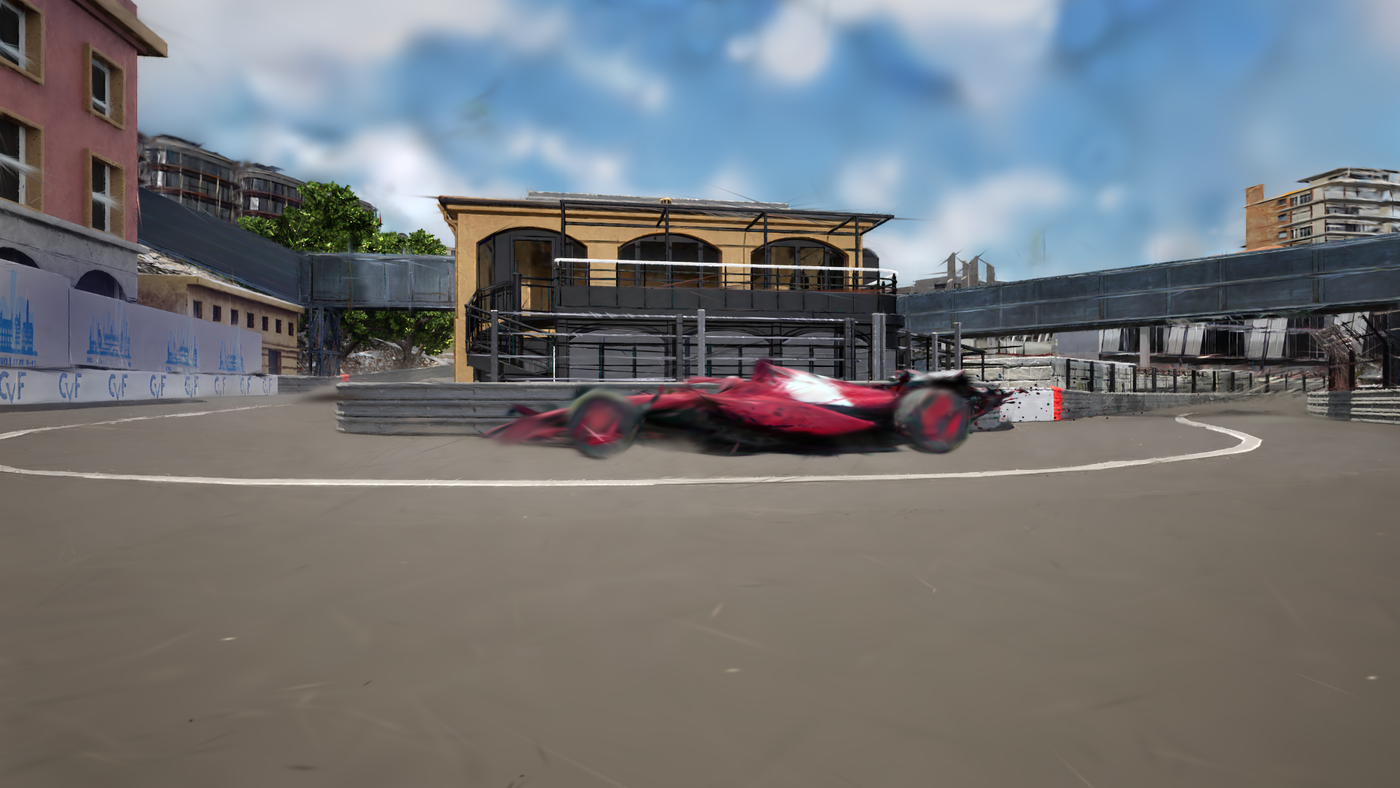}{2.1,0.9}{0.9,0.56} &
	\spyimg{0.2\linewidth}{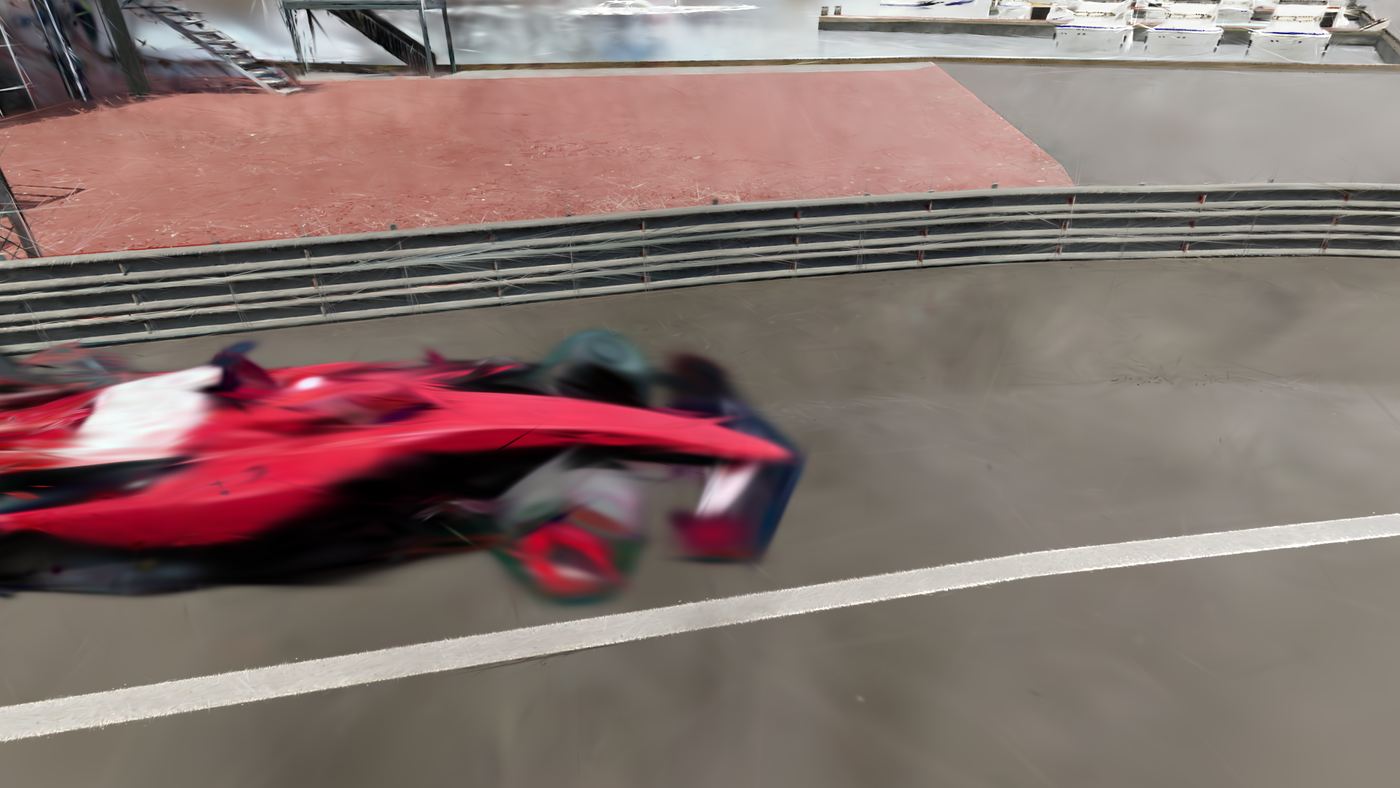}{1.35,0.65}{2.8,0.56} &
	\spyimg{0.2\linewidth}{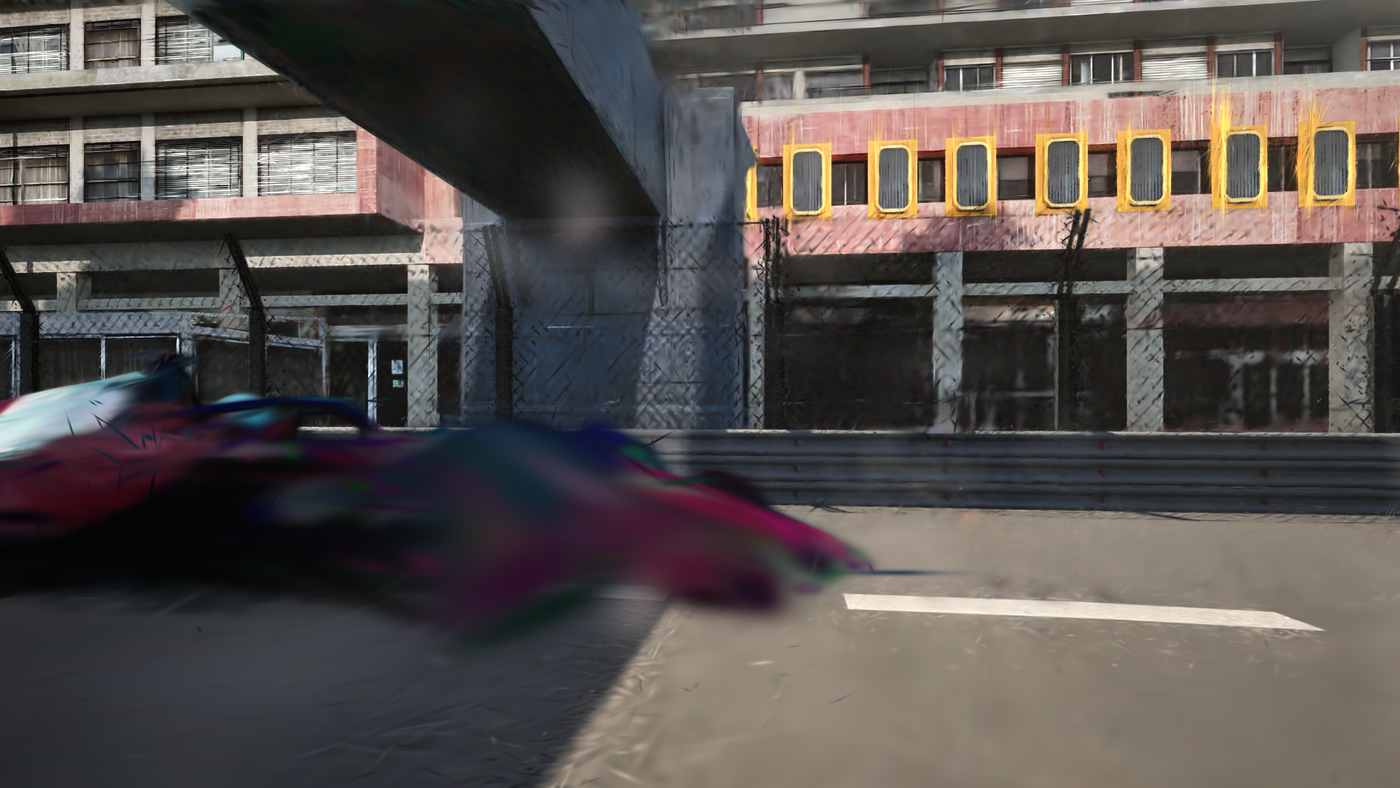}{1.2,0.6}{2.9,0.9} \\

	\spyimgfast{0.2\linewidth}{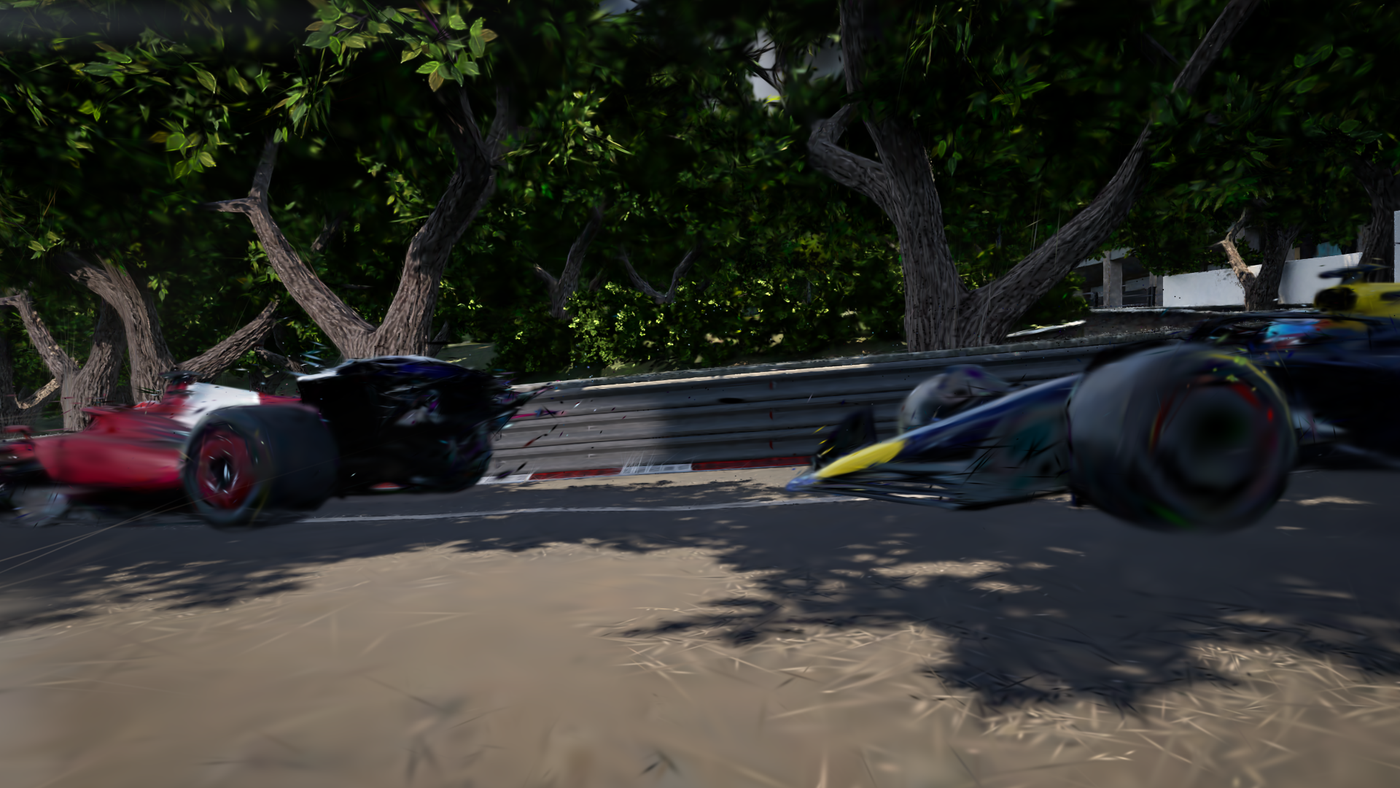}{2.0,0.8}{1.1,0.56} &
	\spyimgfast{0.2\linewidth}{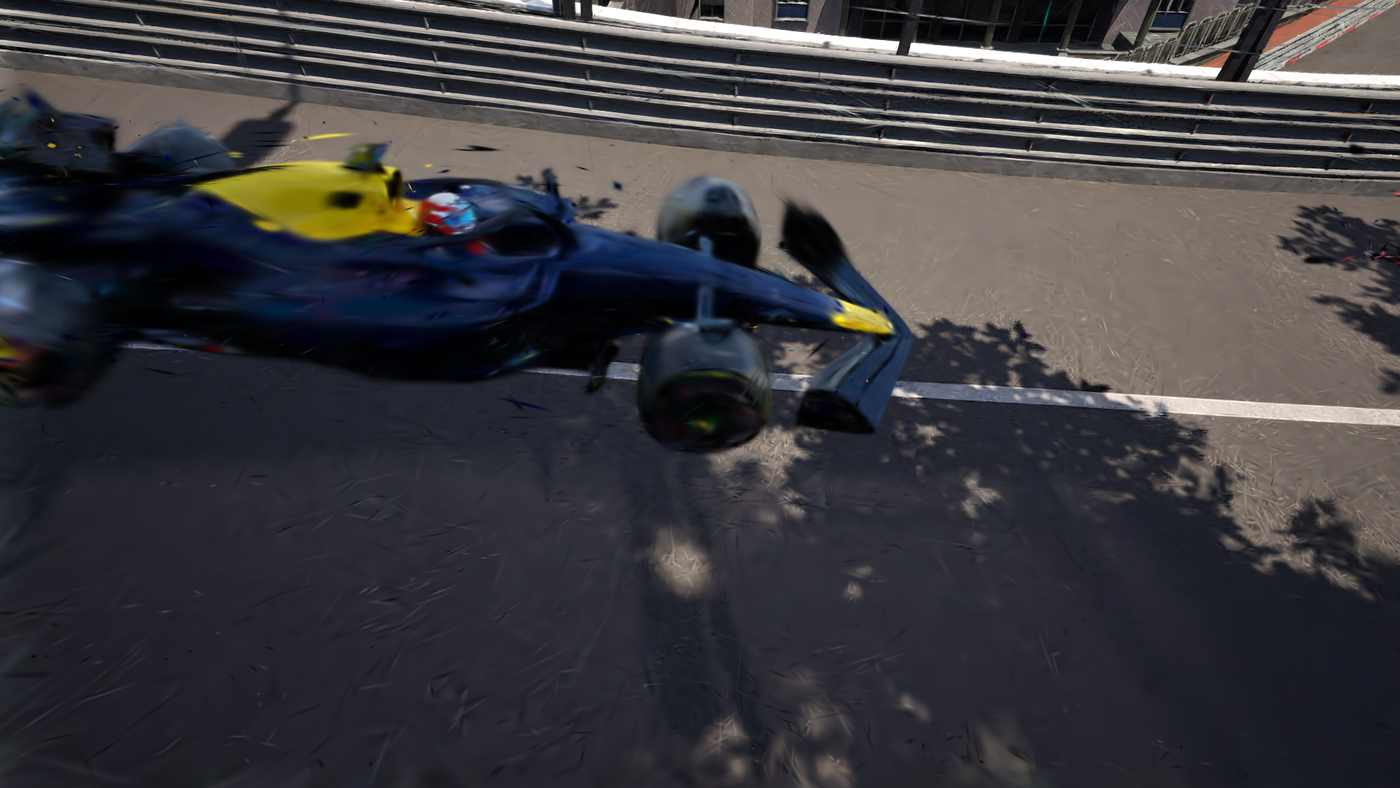}{1.1,1.3}{2.4,0.56} &
	\spyimgfast{0.2\linewidth}{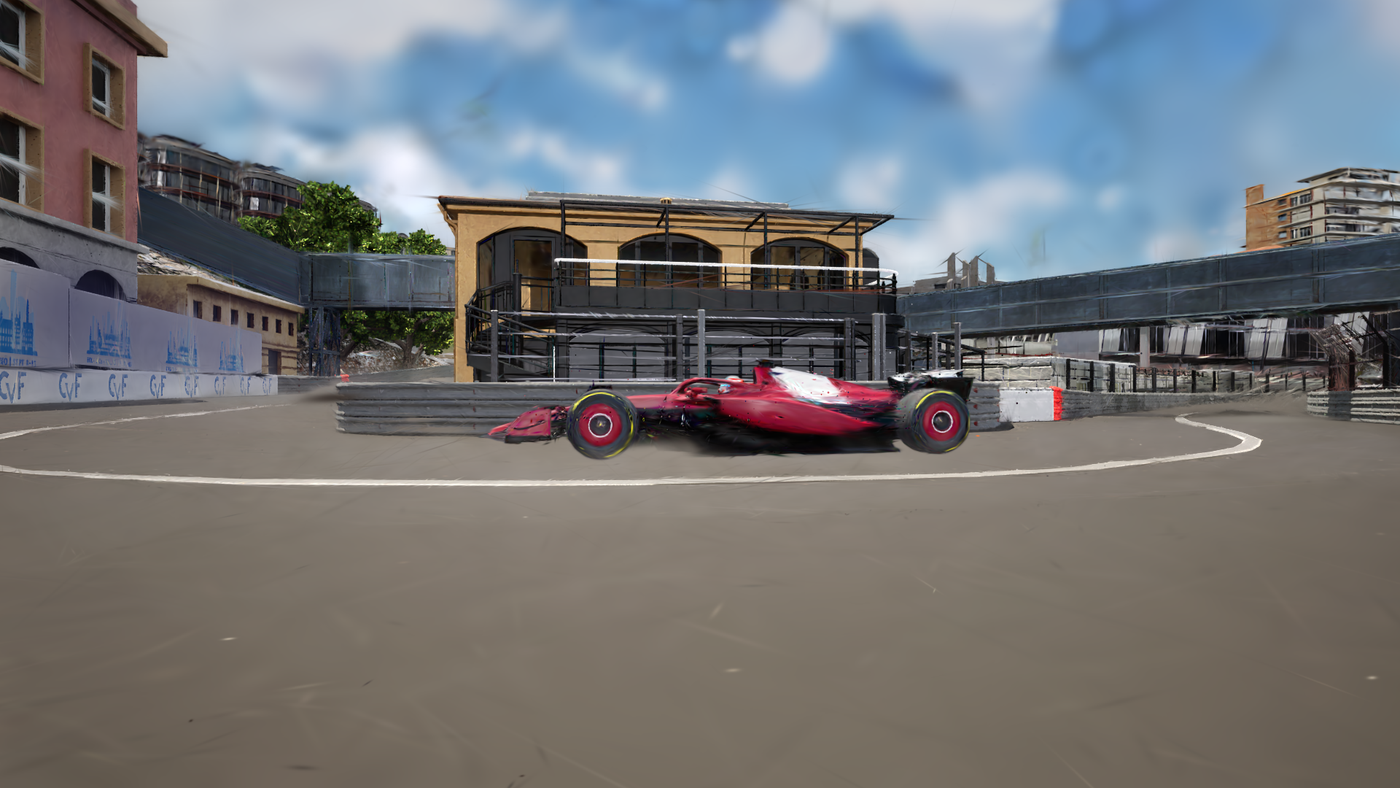}{2.1,0.9}{0.9,0.56} &
	\spyimgfast{0.2\linewidth}{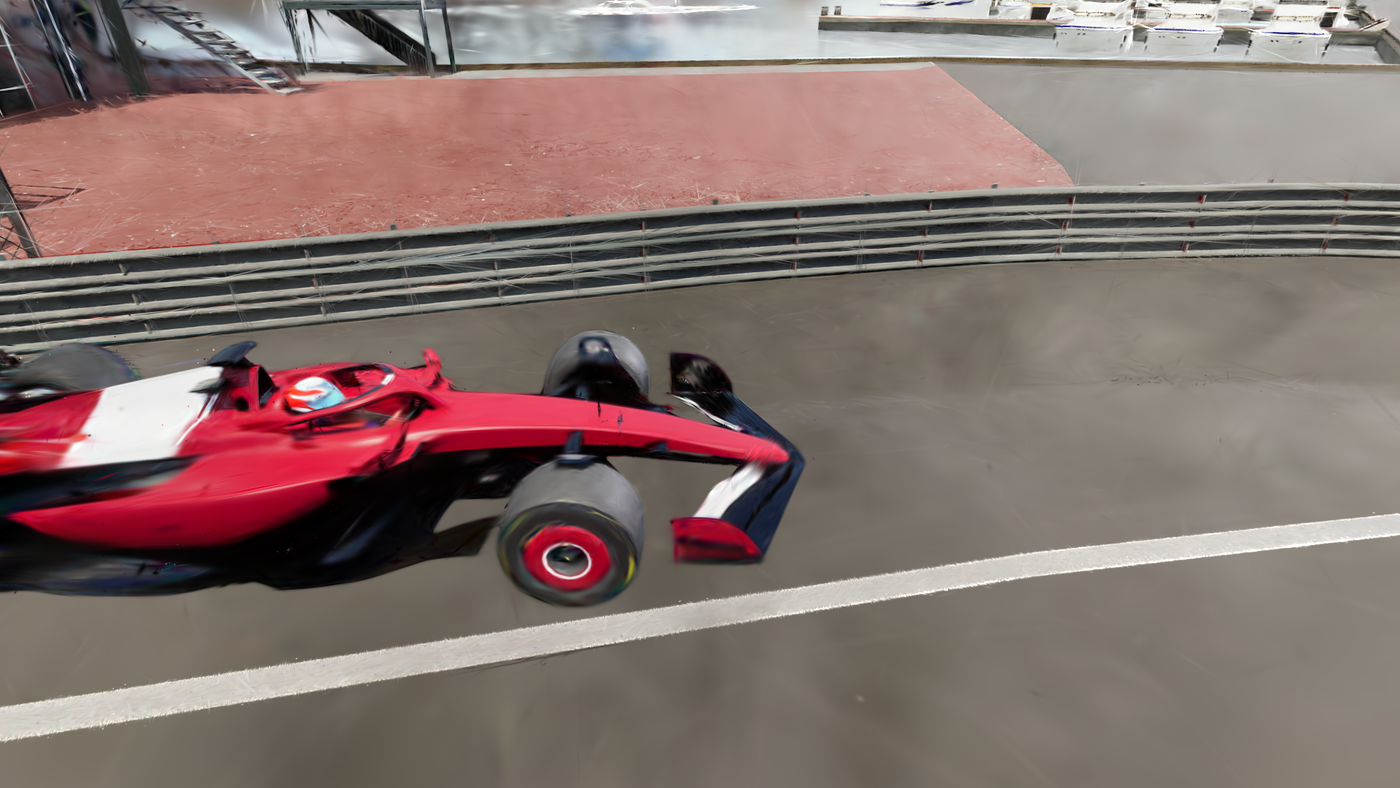}{1.35,0.65}{2.8,0.56} &
	\spyimgfast{0.2\linewidth}{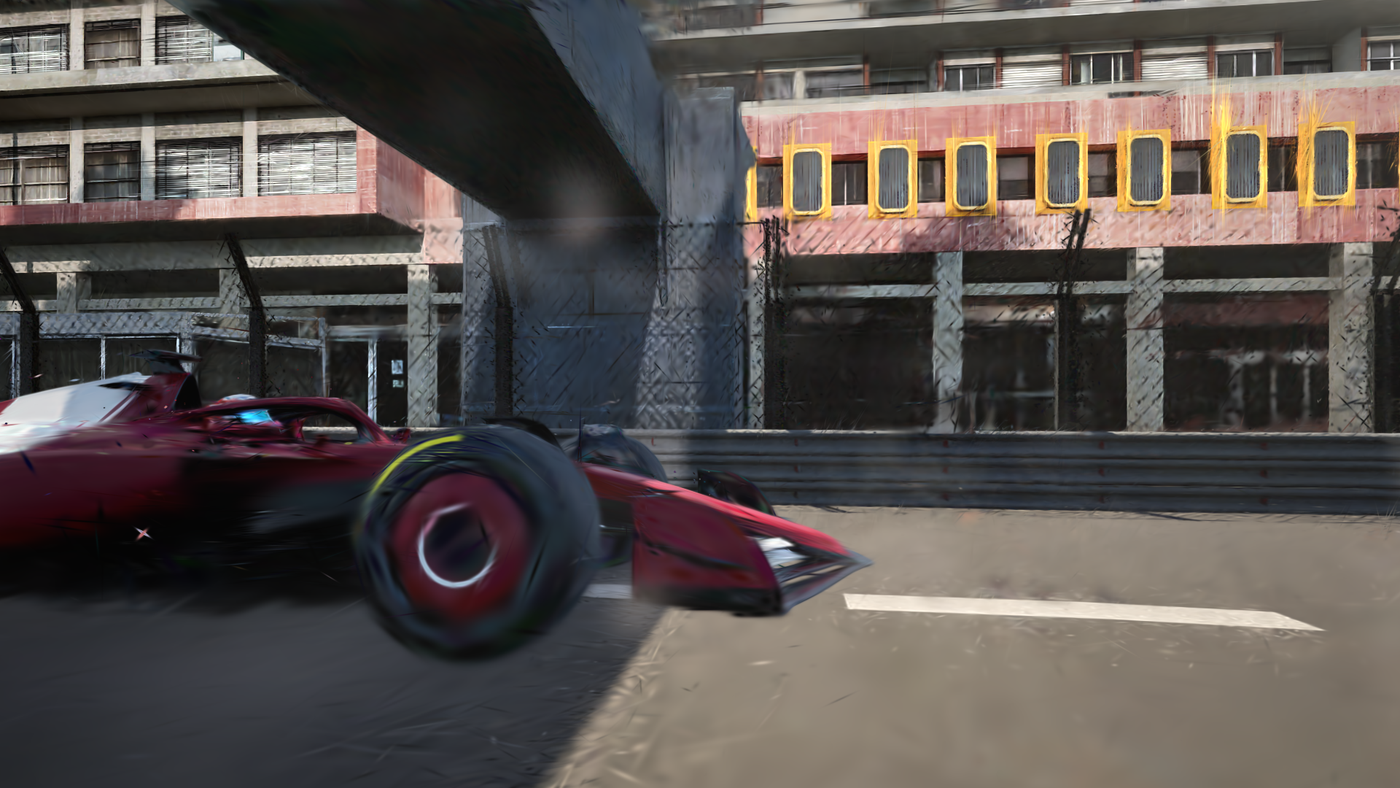}{1.2,0.6}{2.9,0.9} \\	

	\end{tabular}

    \caption{Qualitative Results for test cameras of Monaco4D. \textcolor{green}{Insets} depict the renders from our 3DGS baseline and \textcolor{magenta}{insets} are from FastFlowGS.}
	\label{fig:results-monaco}
    % \vspace{-10pt}
\end{figure*}

\paragraph{Full-quality evaluation.}
\label{para:monaco_full}

Under this protocol the static background uses the full hierarchical reconstruction at roughly 8\,million Gaussians, exceeding the memory budget of all streaming baselines; the only valid comparison is therefore against 3DGS-Base.
That this comparison reduces to a single baseline is not a limitation of the experimental design rather, the central empirical finding of this section, and we return to it directly with the reduced-memory experiment below.

We evaluate on three sequences spanning increasing scene complexity: Fairmont puts five mutually occluding cars through a tight hairpin, Main Straight runs three cars at maximum speed, Rascasse, the simplest, has a single car.

FastFlowGS outperforms 3DGS-Base on every sequence and metric (Tab.~\ref{tab:monaco}).
The gains are correlated with scene complexity, largest on Fairmont and smallest on Rascasse.
Full-image PSNR understates the difference because the static background dominates the per-pixel average. 3DGS-Base produces visibly smeared car surfaces with lost high-frequency detail that appear clearly in Fig.~\ref{fig:results-monaco} but are diluted by full-image metrics.

\paragraph{Why existing streaming methods fail.}

The failure of streaming baselines on Monaco4D is not a matter of degree.
Trained on masked dynamic regions under the full-quality protocol, all five methods lose the majority of their Gaussians by frame~2 and produce empty reconstructions by frame~5.
Three conditions, absent in combination from every prior streaming benchmark, compound to produce this outcome:
\begin{enumerate}
    \item Inter-frame displacements are an order of magnitude beyond what any existing method was designed or evaluated for.
    \item Cars routinely leave and re-enter individual camera frustums between consecutive frames, breaking the assumption of continuous multi-view coverage that underpins every baseline.
    \item At most 7--8 cameras observe the dynamic region at once, providing far sparser per-object supervision than any indoor benchmark.
\end{enumerate}

\begin{figure*}[]
	\centering
	\setlength{\tabcolsep}{0.1pt}
	\renewcommand{\arraystretch}{0}
        
	\begin{tabular}{ccccc}

    Ground truth & Ours & D3DGS & QUEEN & ReConGS \\
	
	\includegraphics[width=0.2\linewidth]{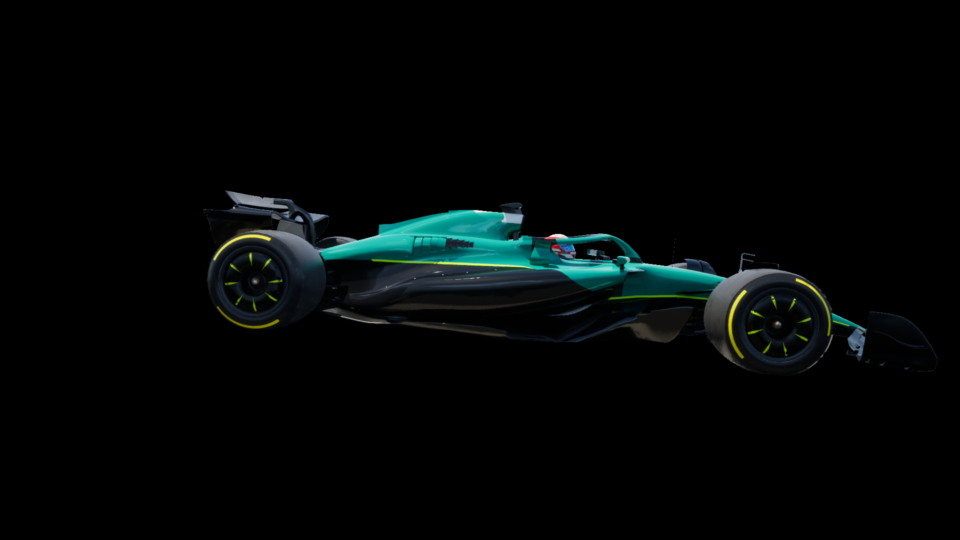} &
	\includegraphics[width=0.2\linewidth]{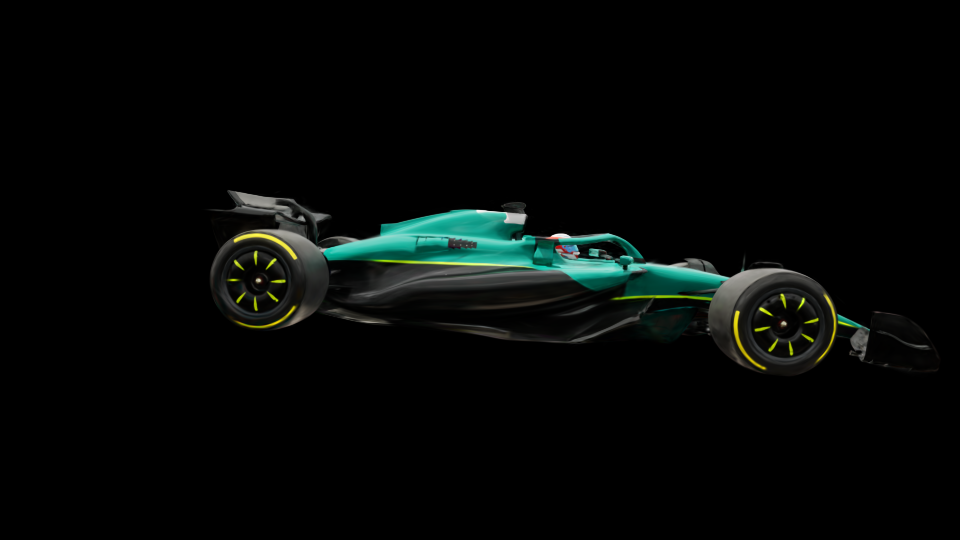} &
	\includegraphics[width=0.2\linewidth]{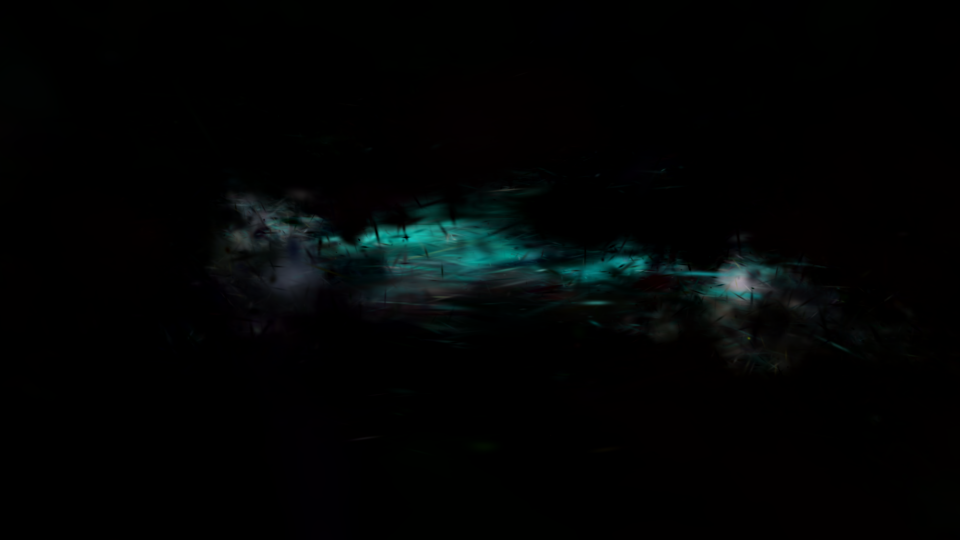} &
	\includegraphics[width=0.2\linewidth]{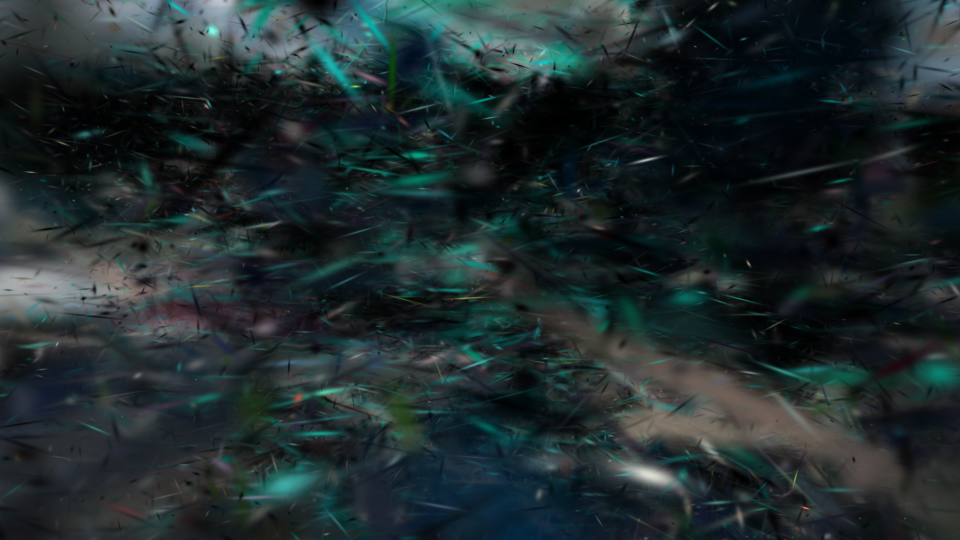} &
	\includegraphics[width=0.2\linewidth]{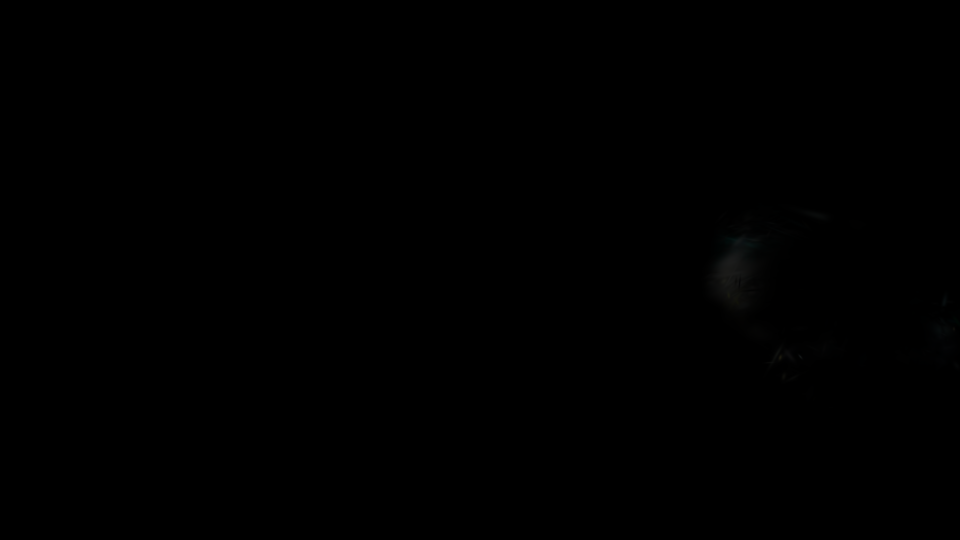}\\
	
	\end{tabular}

    \caption{Renders from a training view at frame~2. All streaming baselines initialized on frame~1 lose their Gaussians within one step. Large displacements push primitives outside the dynamic mask, severing gradient flow before optimization can recover.}
	\label{fig:mask_failure}
\end{figure*}

Methods without densification, D-3DGS and TrackerSplat, degrade monotonically because Gaussians displaced beyond the dynamic mask lose their gradient signal and are never recovered.
Methods with densification, 3DGStream, HiCoM, ReCon-GS, and QUEEN, eventually regenerate an approximate shape, but the initial estimate collapses so completely that recovery is indistinguishable from restarting 3DGS from scratch on each frame (Fig.~\ref{fig:mask_failure}).

\paragraph{Reduced-memory evaluation.}
\label{para:monaco_reduced}

To test whether this failure is intrinsic to the displacement regime or merely a consequence of memory constraints, we replace the hierarchical background with a standard 3DGS at one-eighth the point count, roughly 1\,million Gaussians, bringing every method within budget.
Per-timestep iterations are increased to allow convergence, so FastFlowGS numbers in Tab.~\ref{tab:all_seq_ind} differ slightly from the full-quality protocol.

The answer is unambiguous (Tab.~\ref{tab:all_seq_ind}).
Despite operating against a background eight times denser than any baseline, FastFlowGS achieves the highest VE on every sequence.
The gap is not primarily about raw timing: on Fairmont, HiCoM finishes within two seconds of FastFlowGS (27\,s vs.\ 25\,s), yet produces VMAF~0.89 against FastFlowGS's 46.89.
D-3DGS is the only baseline that achieves recognizable reconstruction quality (VMAF 32--41 across sequences), but at 720--740 seconds per frame it is two orders of magnitude slower than FastFlowGS.
QUEEN shows its best result on Pool, where its VE of 1.68 is the closest any baseline comes to FastFlowGS's 2.36, but its VMAF of 38.75 trails by 13 points and it collapses on every other sequence.
HiCoM, ReCon-GS, and TrackerSplat fall below VMAF~10 on most sequences.
Relaxing the memory constraint does not resolve the failure, which confirms that the problem is a fundamental mismatch between the displacement regime Monaco4D introduces and the motion models these methods rely on. Qualitative results in Fig. 1 in supplemental.

\begin{table*}[t]
\caption{\textbf{Quantitative results with reduced-memory background, per sequence.} \textit{Time} is in seconds. QUEEN ran out of memory on Main Straight (--). Full results including PSNR and MPSNR are in the supplemental. (Tab. \ref{tab:all_seq_ind})}
\label{tab:all_seq_ind}
\resizebox{\textwidth}{!}{
\begin{tabular}{@{}l|ccc|ccc|ccc|ccc|ccc|ccc@{}}
\toprule
 & \multicolumn{3}{c|}{Fairmont} & \multicolumn{3}{c|}{Main Straight} & \multicolumn{3}{c|}{Rascasse} & \multicolumn{3}{c|}{Tunnel} & \multicolumn{3}{c|}{Uphill} & \multicolumn{3}{c}{Pool} \\
\cmidrule(lr){2-4}\cmidrule(lr){5-7}\cmidrule(lr){8-10}\cmidrule(lr){11-13}\cmidrule(lr){14-16}\cmidrule(lr){17-19}
Method & VMAF & Time & VE & VMAF & Time & VE & VMAF & Time & VE & VMAF & Time & VE & VMAF & Time & VE & VMAF & Time & VE \\
\midrule
D-3DGS
 & \cellcolor[HTML]{FEC088}40.47 & 738 & 0.05
 & \cellcolor[HTML]{FEC088}35.83 & 724 & \cellcolor[HTML]{FEC088}0.05
 & \cellcolor[HTML]{FEC088}38.73 & 729 & 0.05
 & \cellcolor[HTML]{FEC088}32.11 & 732 & 0.04
 & \cellcolor[HTML]{FEC088}41.17 & 740 & \cellcolor[HTML]{FEC088}0.06
 & \cellcolor[HTML]{FEC088}34.36 & 727 & 0.05 \\
HiCoM
 & 0.89 & \cellcolor[HTML]{FEC088}27 & 0.03
 & 1.09 & \cellcolor[HTML]{FEC088}24 & \cellcolor[HTML]{FEC088}0.05
 & 2.01 & 25 & 0.08
 & 0.25 & \cellcolor[HTML]{FEC088}24 & 0.01
 & 0.97 & \cellcolor[HTML]{FEC088}29 & 0.03
 & 1.16 & 26 & 0.04 \\
QUEEN
 & 5.39 & 61 & \cellcolor[HTML]{FEC088}0.09
 & -- & -- & --
 & 12.67 & \cellcolor[HTML]{FEC088}23 & \cellcolor[HTML]{FEC088}0.58
 & 0.39 & 69 & 0.01
 & 0.22 & 70 & 0
 & 38.75 & \cellcolor[HTML]{FEC088}23 & \cellcolor[HTML]{FEC088}1.68 \\
ReConGs
 & 0.71 & 48 & 0.01
 & 1.21 & 48 & 0.03
 & 1.66 & 35 & 0.05
 & 0.36 & 39 & 0.01
 & 0.89 & 39 & 0.02
 & 1.24 & 25 & 0.05 \\
Trackersplat
 & 2.72 & 72 & 0.04
 & 0.78 & 95 & 0.01
 & 9.67 & 70 & 0.14
 & 7.15 & 94 & \cellcolor[HTML]{FEC088}0.08
 & 1.59 & 60 & 0.03
 & 6.77 & 58 & 0.12 \\
\midrule
FastFlowGS
 & \cellcolor[HTML]{FD8488}46.89 & \cellcolor[HTML]{FD8488}25 & \cellcolor[HTML]{FD8488}1.88
 & \cellcolor[HTML]{FD8488}50.1  & \cellcolor[HTML]{FD8488}21 & \cellcolor[HTML]{FD8488}2.39
 & \cellcolor[HTML]{FD8488}53.84 & \cellcolor[HTML]{FD8488}23 & \cellcolor[HTML]{FD8488}2.34
 & \cellcolor[HTML]{FD8488}48.9  & \cellcolor[HTML]{FD8488}19 & \cellcolor[HTML]{FD8488}2.57
 & \cellcolor[HTML]{FD8488}55.84 & \cellcolor[HTML]{FD8488}27 & \cellcolor[HTML]{FD8488}2.07
 & \cellcolor[HTML]{FD8488}51.87 & \cellcolor[HTML]{FD8488}22 & \cellcolor[HTML]{FD8488}2.36 \\
\bottomrule
\end{tabular}
}
\end{table*}

\subsection{Ablations}
\label{subsec:ablations}

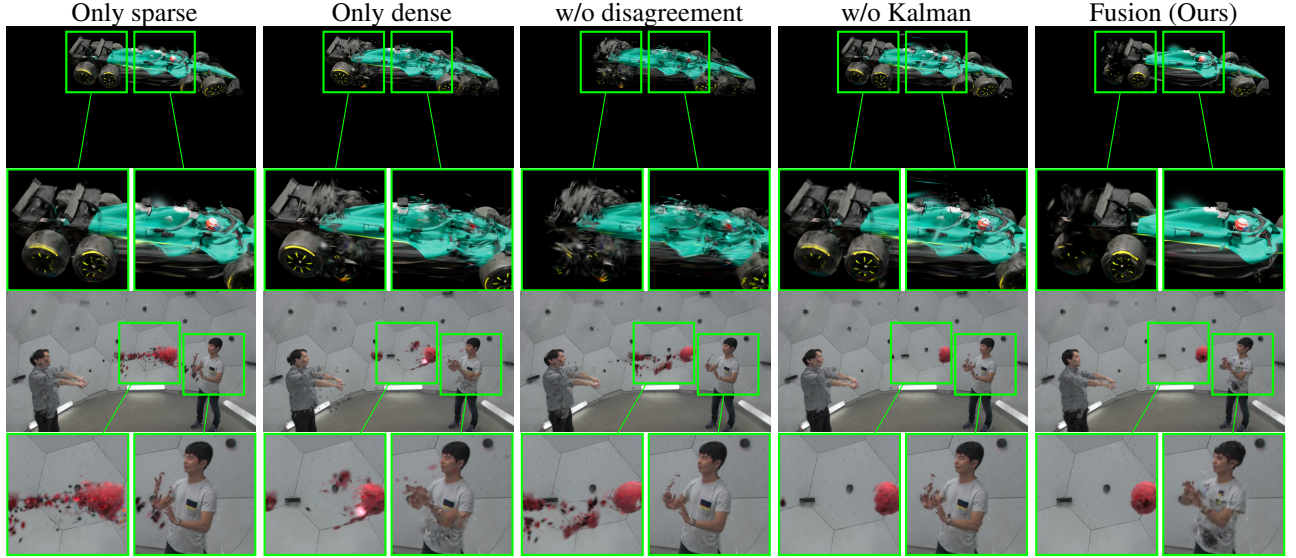
\begin{figure*}[]
	\centering
	\setlength{\tabcolsep}{0.1pt}
	\renewcommand{\arraystretch}{0}
    
	\newcommand{\spyimg}[4]{
		\begin{tikzpicture}[spy using outlines={green,magnification=2,size=1.2cm, connect spies}]
			\node[anchor=south west,inner sep=0] at (0,0) {\includegraphics[width=#1]{#2}};
			\spy[every spy on node/.append style={thick}] on (#3) in node [left] at (#4);
		\end{tikzpicture}
	}

	\newcommand{\spyimgtwo}[6]{
		\begin{tikzpicture}[spy using outlines={green,magnification=2,size=1.6cm, connect spies}]
			\node[anchor=south west,inner sep=0] at (0,0) {\includegraphics[width=#1]{#2}};
			\begin{scope}
				\spy[every spy on node/.append style={thick}] on (#3) in node [left] at (#4);
			\end{scope}
			\begin{scope}
				\spy[every spy on node/.append style={thick}] on (#5) in node [left] at (#6);
			\end{scope}
		\end{tikzpicture}
	}
    
	\begin{tabular}{ccccc}

    Only sparse & Only dense & w/o disagreement & w/o Kalman & Fusion (Ours) \\
	
	\spyimgtwo{0.2\linewidth}{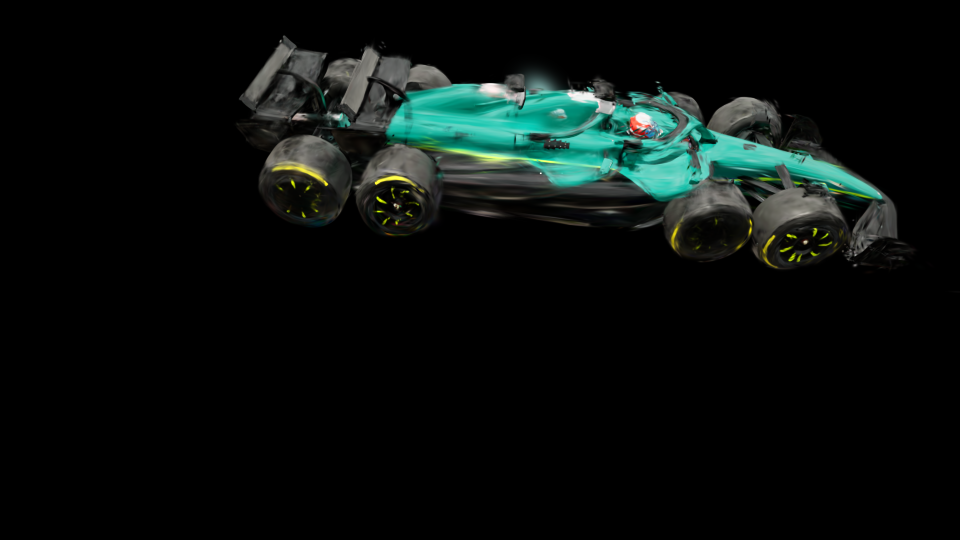}{1.2,1.4}{1.62,-0.82}{2.1,1.4}{3.3,-0.82} &
	\spyimgtwo{0.2\linewidth}{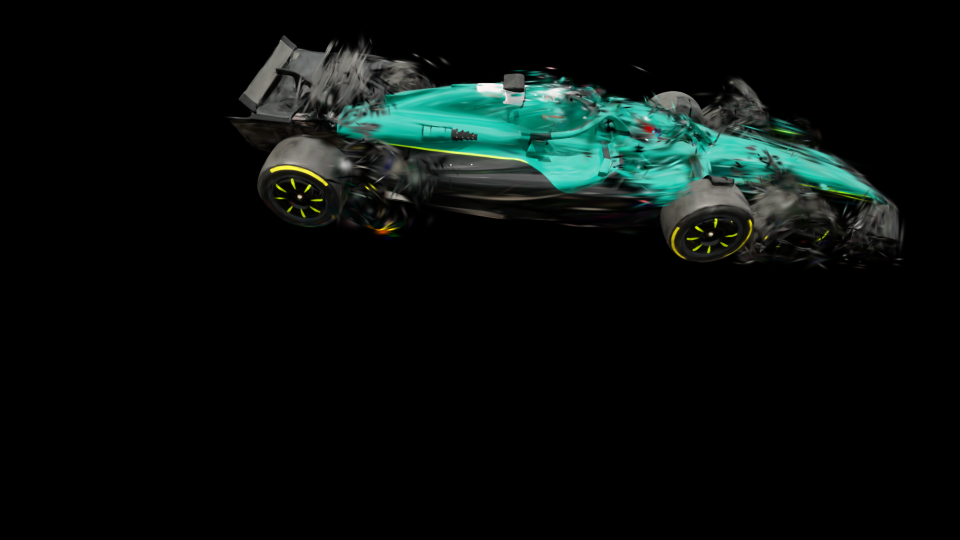}{1.2,1.4}{1.62,-0.82}{2.1,1.4}{3.3,-0.82} &
	\spyimgtwo{0.2\linewidth}{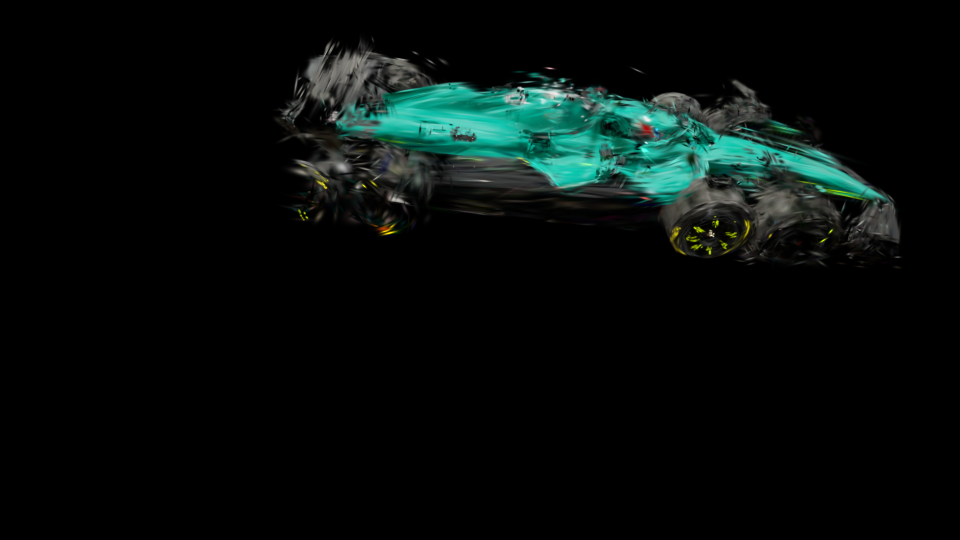}{1.2,1.4}{1.62,-0.82}{2.1,1.4}{3.3,-0.82} &
	\spyimgtwo{0.2\linewidth}{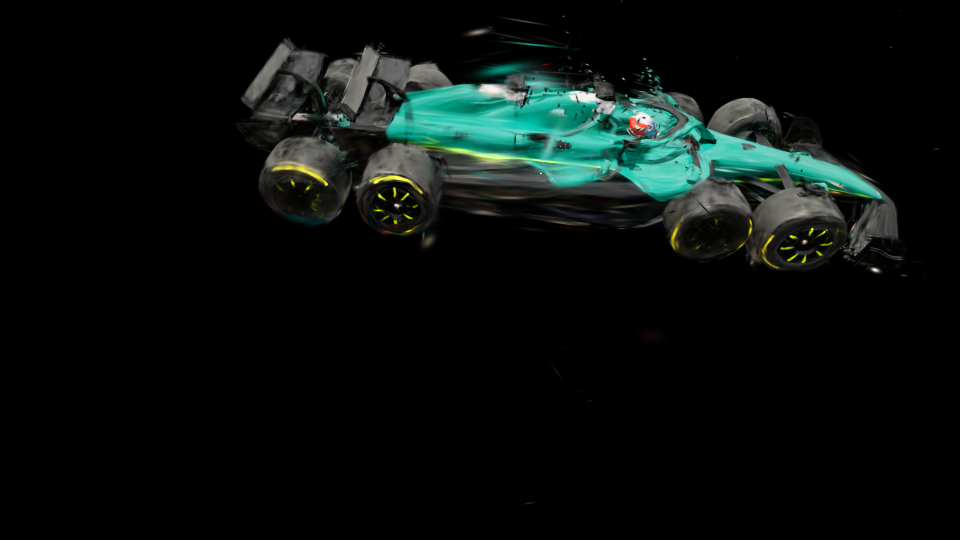}{1.2,1.4}{1.62,-0.82}{2.1,1.4}{3.3,-0.82} &
	\spyimgtwo{0.2\linewidth}{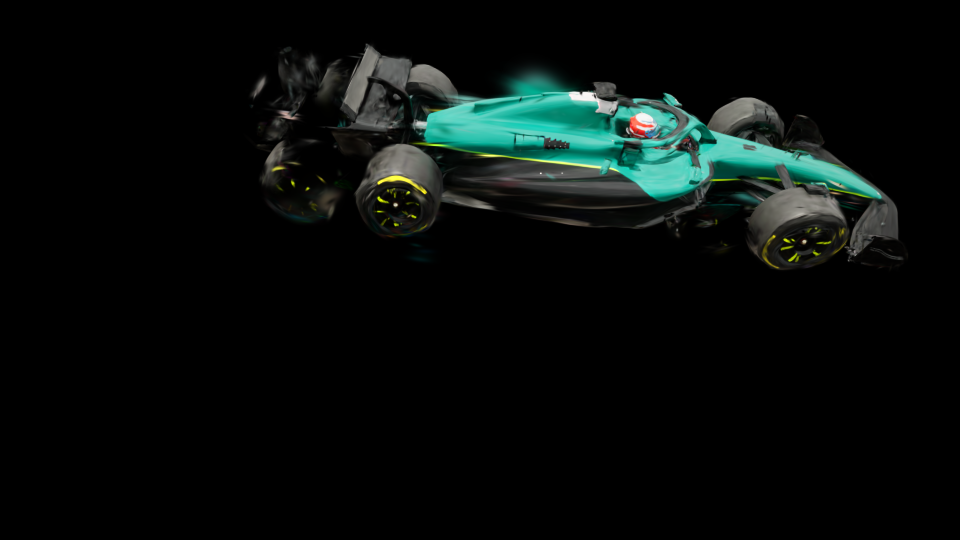}{1.2,1.4}{1.62,-0.82}{2.1,1.4}{3.3,-0.82} \\
    
	\spyimgtwo{0.2\linewidth}{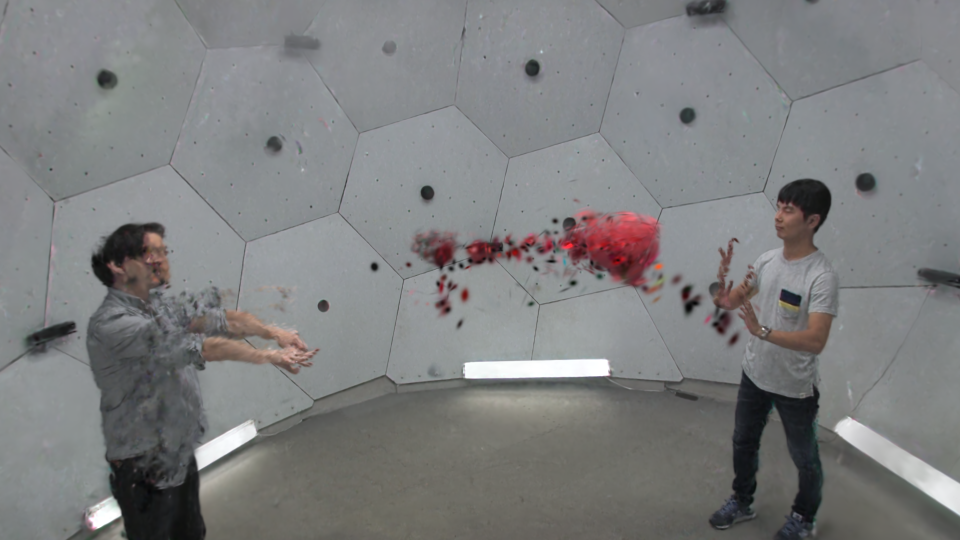}{1.9,1.05}{1.62,-0.82}{2.75,0.9}{3.3,-0.82} &
	\spyimgtwo{0.2\linewidth}{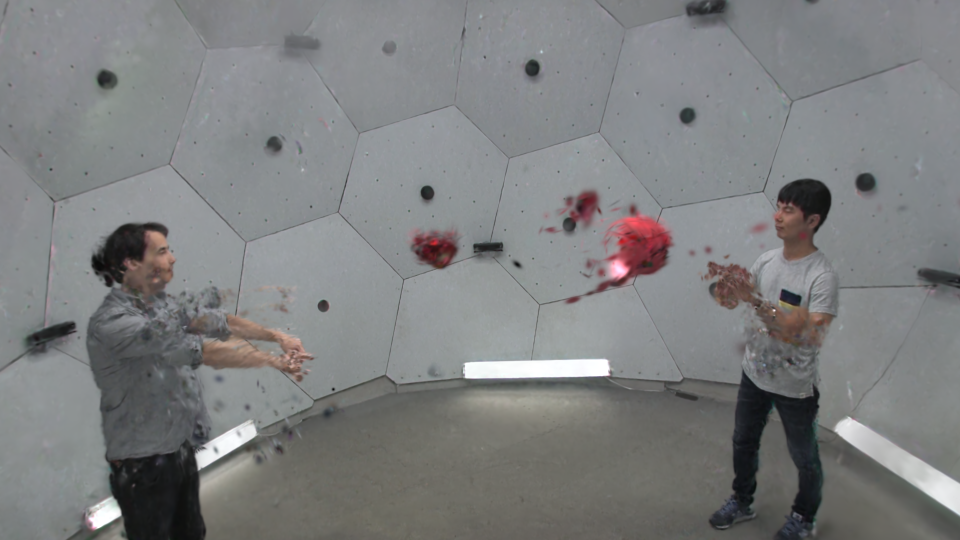}{1.9,1.05}{1.62,-0.82}{2.75,0.9}{3.3,-0.82} &
	\spyimgtwo{0.2\linewidth}{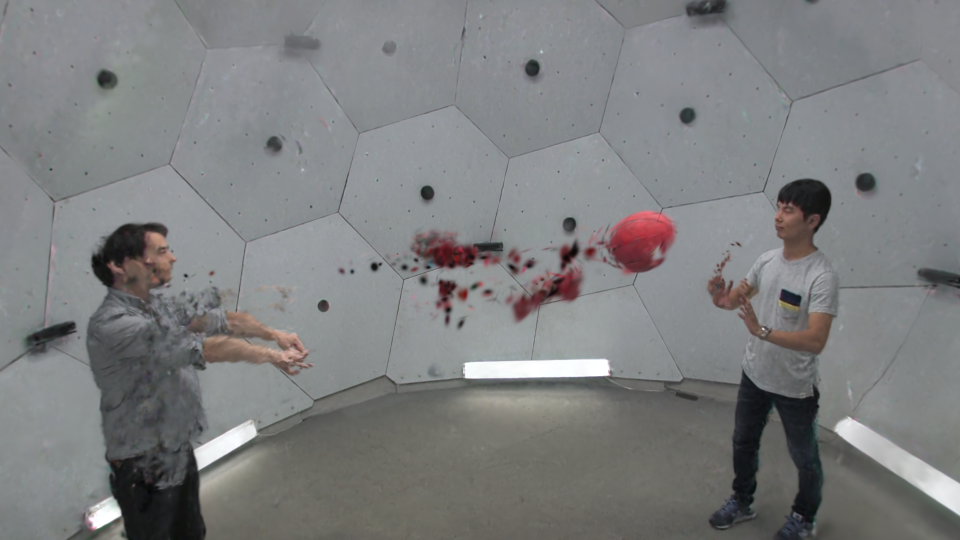}{1.9,1.05}{1.62,-0.82}{2.75,0.9}{3.3,-0.82} &
	\spyimgtwo{0.2\linewidth}{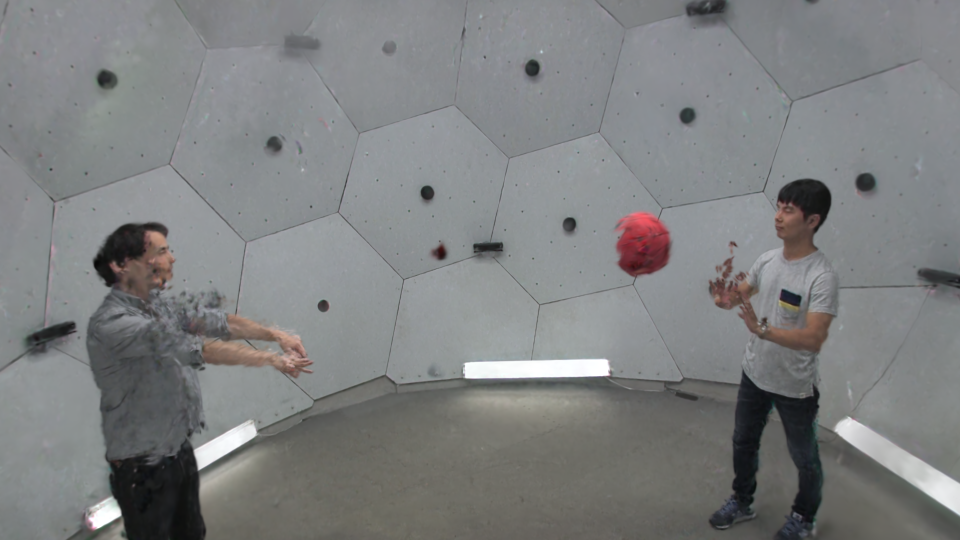}{1.9,1.05}{1.62,-0.82}{2.75,0.9}{3.3,-0.82} &
	\spyimgtwo{0.2\linewidth}{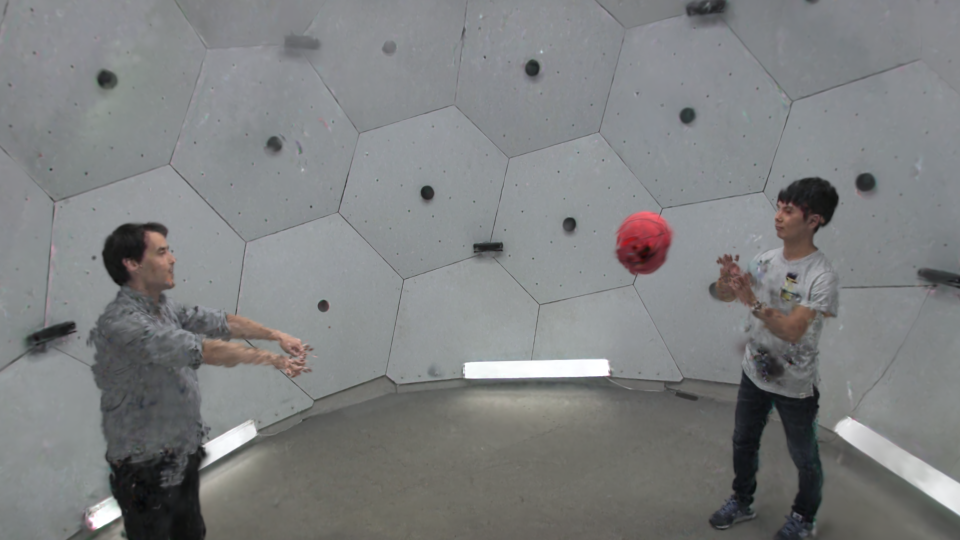}{1.9,1.05}{1.62,-0.82}{2.75,0.9}{3.3,-0.82} \\

	\end{tabular}

    \caption{\textbf{Ablation study on Monaco4D (top) and CMU-Panoptic (bottom).} We additionally report \#iters: the number of iterations to reach 24+ PSNR on CMU-Panoptic and 18.5+ PSNR on Monaco4D.}
    
	\label{fig:ablations-panoptic}
\end{figure*}

Tab.~\ref{tab:ablations} and Fig.~\ref{fig:ablations-panoptic} show that every component contributes, and the gains compound rather than overlap.

Sparse and dense correspondences alone reach similar quality on CMU-Panoptic (M-PSNR 25.31 vs.\ 25.29), with dense converging modestly faster.
On Monaco4D, the advantage of dense correspondences is clearer (VMAF 40.76 vs.\ 39.33), consistent with the greater prevalence of textureless surfaces and specular bodywork that makes sparse feature matching unreliable outdoors.
Fusing both outperforms either in isolation on every metric, confirming that the two signals cover different failure modes rather than providing redundant coverage.

The disagreement filter's primary contribution is to convergence rather than final quality, removing 40 iterations on CMU-Panoptic and 80 on Monaco4D before the quality threshold is reached.
The larger effect outdoors is consistent with the higher rate of flow failures on Monaco4D, where filtering unreliable correspondences before they enter the optimizer has more noise to suppress.

The Kalman temporal prior delivers the single largest improvement across all components and both datasets: VMAF rises by 2.55 points on CMU-Panoptic and 2.76 on Monaco4D, and convergence drops from 260 to 200 iterations and from 760 to 700 respectively.
We attribute this to the prior constraining Gaussian positions to a temporally smooth trajectory, reducing the search space the optimizer must explore at each new frame.

\noindent\textit{Component interactions.}
The disagreement filter operates on the fused correspondence set, which already combines sparse and dense coverage; applied to either signal alone, it would identify fewer cross-signal disagreements and suppress less noise.
The Kalman prior then acts on a well-filtered set, and its gain is largest precisely because upstream filtering has removed the noise that would otherwise corrupt the temporal smoothness assumption.

\begin{table}[]
\centering
\setlength{\tabcolsep}{2pt}
\caption{Per-component ablation. \emph{\#iters} is iterations to reach 24+ PSNR on CMU-Panoptic and 18.5+ PSNR on Monaco4D (averaged over training views).}
\begin{tabular}{cccc|ccc|ccc}
\hline
\multicolumn{4}{c|}{}                  & \multicolumn{3}{c|}{CMU Panoptic (Football)} & \multicolumn{3}{c}{Monaco4D (Fairmont)} \\ \hline
Sparse & Dense & Disag. & Kalman & VMAF$\uparrow$ & MPSNR$\uparrow$ & \#iters$\downarrow$    & VMAF$\uparrow$ & MPSNR$\uparrow$ & \#iters$\downarrow$ \\ \hline
\textcolor{green}{\cmark}      & \textcolor{red}{\xmark}      &  \textcolor{red}{\xmark}            & \textcolor{red}{\xmark}       & 46.14& 25.31& 350&         39.33&         18.65&                     1000
\\
  \textcolor{red}{\xmark}     & \textcolor{green}{\cmark}     & \textcolor{red}{\xmark}             & \textcolor{red}{\xmark}       & 45.32& 25.29& 320&         40.76&         18.64&                     940
\\
\textcolor{green}{\cmark}      & \textcolor{green}{\cmark}     &  \textcolor{red}{\xmark}            &  \textcolor{red}{\xmark}      & 46.49& 25.37& 300&         40.84&         18.71&                     840
\\
\textcolor{green}{\cmark}      & \textcolor{green}{\cmark}     & \textcolor{green}{\cmark}            &  \textcolor{red}{\xmark}      & 46.82& 25.43& 260&         41.17&         18.76&                     760
\\
\textcolor{green}{\cmark}      & \textcolor{green}{\cmark}     & \textcolor{green}{\cmark}            & \textcolor{green}{\cmark}      & 49.37& 25.58& 200&         43.93&         18.88&                     700\\ \hline
\end{tabular}
\label{tab:ablations}
\end{table}

\section{Conclusion}
\label{sec:conclusion}

The central finding of this work is that initialization quality, not optimization budget, is the bottleneck in streaming Gaussian reconstruction. By fusing multi-resolution correspondences with a Kalman temporal prior yields initializations that converge faster and remain stable over long sequences.
FastFlowGS instantiates this principle and outperforms all streaming baselines on CMU-Panoptic. On Monaco4D, the first outdoor multi-view dataset in the large-displacement regime released publicly alongside this work, it is the only method to produce coherent reconstructions where all existing approaches fail entirely.
We hope Monaco4D shifts the community's attention toward the displacement and sparsity regimes that outdoor broadcast applications actually require, rather than the smooth, dense-view conditions that current benchmarks favor.

\section{Limitations and Failure Cases}
\label{sec:limitations}

The motion model updates positions but not rotations or scales, so deforming or rapidly rotating objects must recover shape from scratch, eroding the convergence advantage that good initialization provides. We also acknowledge that correspondence quality degrades on textureless surfaces, distant objects, and depth-dominant motion, where disagreement scoring helps but cannot fully compensate. Variational fusion treats tracker levels as independent, making the fused covariance $\boldsymbol{\Sigma}_{i,t}$ overconfident when levels share image evidence and reducing the optimizer's ability to correct under tight iteration budgets. Both FastFlowGS and Monaco4D assume a static background, so moving spectators or trackside elements produce localized ghosting, and the method has no recovery path when segmentation masks fail across multiple views simultaneously. Monaco4D is synthetic, and while FastFlowGS transfers to real capture on CMU-Panoptic, performance on real outdoor footage at F1-scale displacements remains uncharacterized. Preprocessing is dominated by the point tracker at 727\,s per frame and is excluded from timing comparisons following standard practice~\cite{dynamic3dgs,sun20243dgstreamontheflytraining3d,trackersplat,10.1145/3731214}, with the full breakdown in Sec. 1.3 of the supplemental.
\\

\textbf{Acknowledgments}
We thank Javier Sevilla, Juanfran Ramírez, Arturo Javier Loza, and Daniel Fernandez from BionicApe for their guidance and advice during dataset creation, and Aviral Chharia for helpful suggestions during the review process.

\clearpage

\bibliographystyle{unsrt}
\bibliography{main}

\section{FastFlowGS: Additional Details}

\subsection{Summary of method}
The overview of our method is presented in Alg. \,\ref{alg:fastflowgs}. FastFlowGS represents a dynamic scene as a composition of multiple Gaussian layers rendered together by alpha blending in depth order: a static background modeled once and updated only in appearance, a dynamic foreground updated geometrically at each frame, and in outdoor environments, a distant sky layer modeled by a skybox.
At the core of FastFlowGS is a variational formulation of per-frame Gaussian position initialization.
We treat the true 3D position of each Gaussian as an optimization variable and seek the sequence of positions across time that minimizes a quadratic cost combining two terms: 1) a temporal dynamics term that penalizes deviations from the previous frame's position scaled by accumulated uncertainty, and 2) a multi-source measurement term that penalizes disagreement with each independent tracker estimates scaled by their triangulation covariance.
The closed-form minimizer of this objective is a precision-weighted fusion that subsumes both the Kalman temporal prior and multi-resolution tracker fusion as special cases, and it produces a calibrated per-Gaussian posterior covariance that downstream processes can exploit directly.
The practical motivation is efficiency: at frame $t\!+\!1$, the Gaussian representation from frame $t$ already approximates the scene geometry closely.
A strong geometric initialization that relocates Gaussians to their frame $t\!+\!1$ positions reduces the required finetuning to $N_{\text{dyn}}$ iterations rather than the thousands needed when optimizing from scratch, enabling immersive streaming at interactive rates.

\begin{table}[]
    \centering
    \begin{tabular}{c}
         \includegraphics[width=1\linewidth]{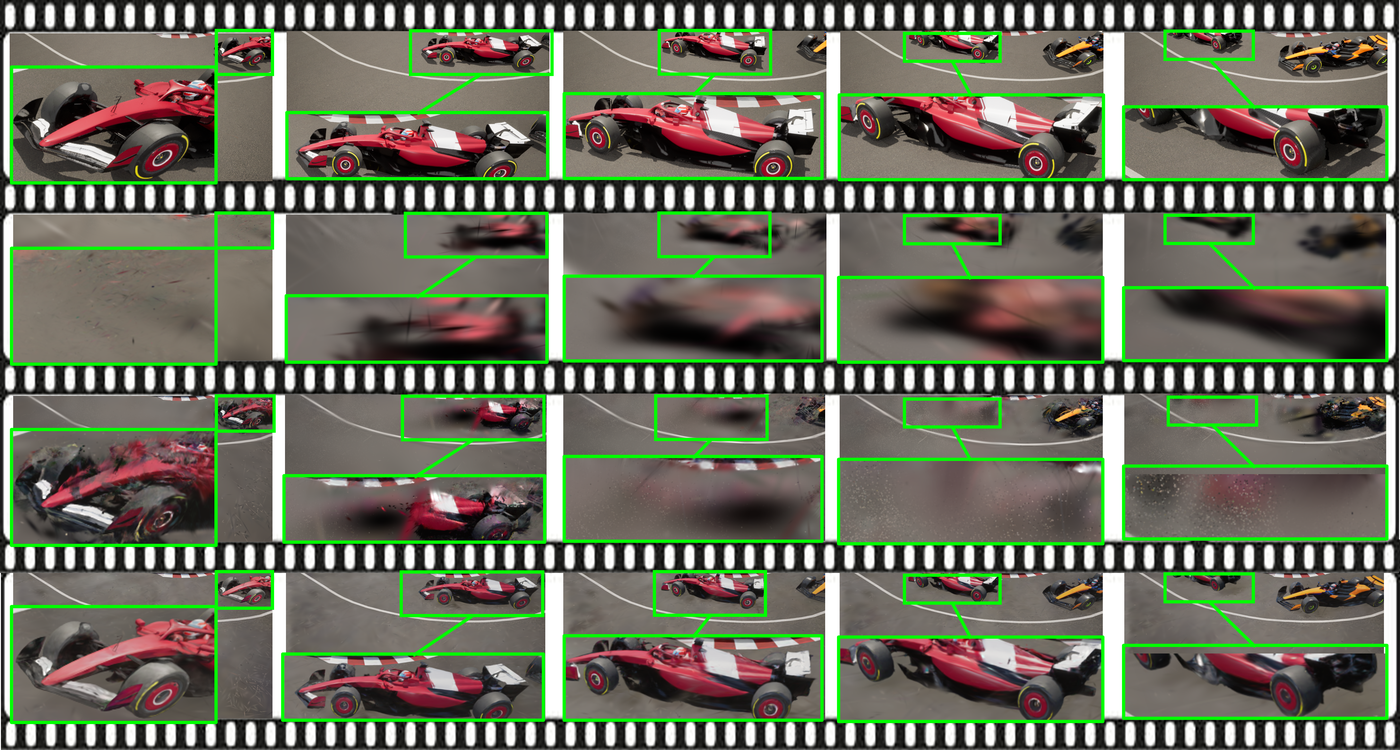} \\ 
         \midrule
         \includegraphics[width=1\linewidth]{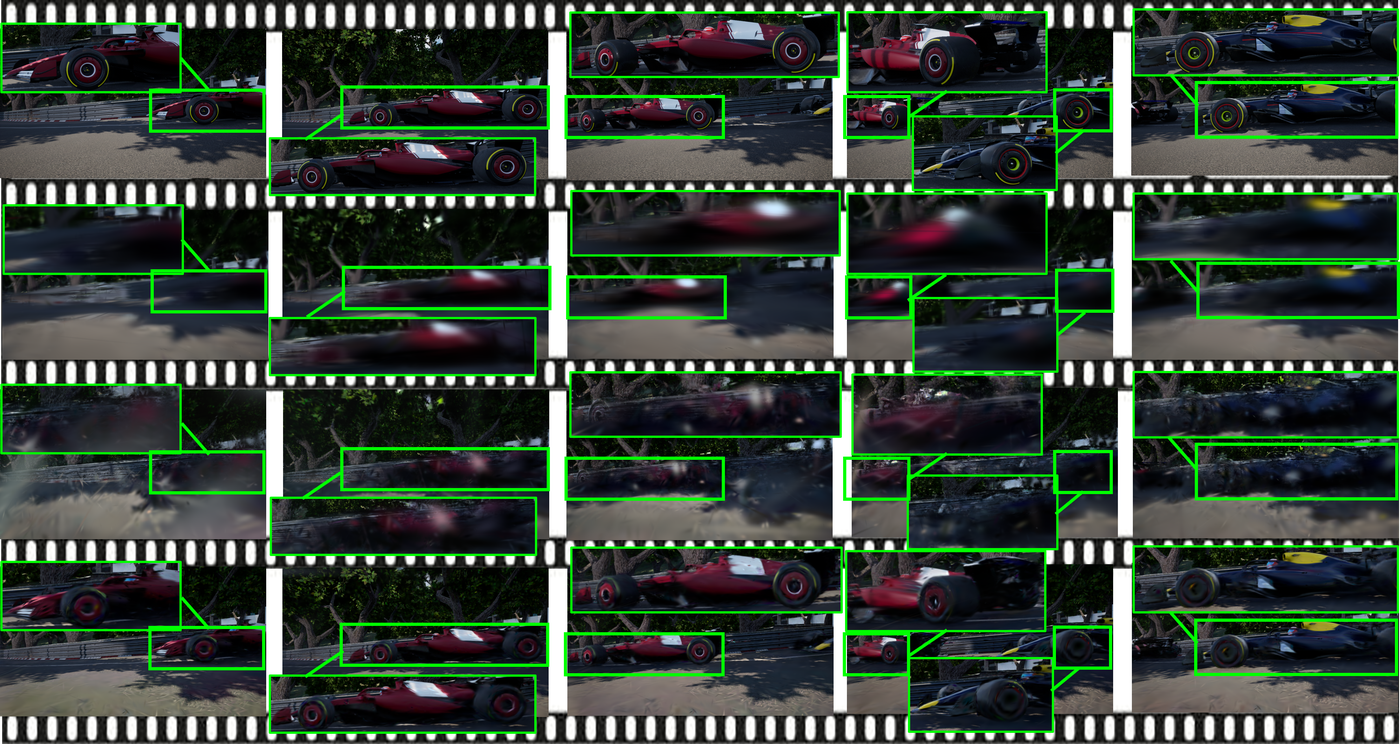} \\ 
    \end{tabular}
    \caption{\textbf{Qualitative comparison on consecutive frames on the best performing methods: } First row is \texttt{Ground truth}, second row is \texttt{QUEEN}, third row is \texttt{TrackerSplat}, fourth row is \texttt{FastFlowGS}}
    \label{fig:qual_video}
\end{table}

\subsection{Additional experiments with competing methods}
We report full metrics (PSNR, M\-PSNR) for the low-memory experiments outlined in Sec. 5.2 of the main text and Tab. \ref{tab:all_seq_ind}.

\begin{table}
    \caption{\textbf{Qualitative results with fewer primitives in background: } FastFlowGS achieves comparable quality with enhanced convergence speed (Time is in seconds)}\label{tab:all_seq}
    \centering
    \scalebox{1}{
        \begin{tabular}{l|ccc|ccc}
            \hline
            & VMAF                          & PSNR                          & MPSNR                         & Time                          & VE                           & PE                           \\ \hline
            D-3DGS       & \cellcolor[HTML]{FEC088}37.11 & \cellcolor[HTML]{FEC088}18.58 & 16.8                          & 731.67                        & 0.05                         & 0.03                         \\
            HiCoM        & 1.06                          & 10.18                         & 10.61                         & 25.83                         & 0.04                         & \cellcolor[HTML]{FEC088}0.40 \\
            QUEEN        & 9.57                          & 15.13                         & 15.07                         & 40.83                         & \cellcolor[HTML]{FEC088}0.39 & \cellcolor[HTML]{FEC088}0.40 \\
            ReConGs      & 1.01                          & 10.47                         & 10.29                         & \cellcolor[HTML]{FEC088}39.00 & 0.03                         & 0.28                         \\
            Trackersplat & 4.78                          & 15.53                         & \cellcolor[HTML]{FD8488}18.78 & 74.83                         & 0.07                         & 0.22                         \\
            Ours         & \cellcolor[HTML]{FD8488}51.24 & \cellcolor[HTML]{FD8488}23.03 & \cellcolor[HTML]{FEC088}17.59 & \cellcolor[HTML]{FD8488}22.83 & \cellcolor[HTML]{FD8488}2.27 & \cellcolor[HTML]{FD8488}1.02 \\ \hline
        \end{tabular}
    }
\end{table}
   
The motivation is that this setting now lets these methods distinguish between foreground and background, which helps them attempt corrections over time, instead of just losing the dynamic Gaussians that fall outside the segmentation mask and are never recovered.

FastFlowGS sometimes loses its advantage in MPSNR, which we justify by a limitation of the metric itself. Fig.\,\ref{fig:qual_video} shows qualitative comparisons of our method and the two highest-ranked competing methods in terms of MPSNR (QUEEN and TrackerSplat). The competing methods' blurriness produces cars with (on-average) closer texture to the groundtruth, but lacking detail or sharp features. In contrast, our method produces visually better images, at the cost of some (on-average) noisier texture. MPSNR prefers blur because the squared error penalizes sharp deviations heavily and a  single sharp edge misalignment hurts MSE more than global blur. As such, though MPSNR is sometimes higher for other methods, we claim FastFlowGS is able to produce better reconstructions.

Overall, FastFlowGS is typically either best or second-best in all metrics, being able to reconstruct scenes with a larger number of Gaussians, faster, and better quality. 

\subsection{Timing breakdown of FastFlowGS}
\label{sec:timing}
Table~\ref{tab:timing} reports the per-frame timing breakdown for FastFlowGS on Fairmont (variation 3).
Static background, dynamic foreground, and skybox optimize on separate GPUs. We report the wall-clock time which equals the slowest lane.

Within the dynamic lane, the initialization pipeline (Voronoi interpolation through fusion) adds less than 1\,s and finetuning dominates at ${\sim}90\%$ of the lane cost.
On Monaco4D, the static SH update is the overall bottleneck due to the large scene scale. However on CMU Panoptic, static and dynamic finish in roughly equal time.
Tracks are precomputed offline. LightGlue (feature tracks) takes ${\sim}$40\,ms/view, GMFlow (dense tracks) takes ${\sim}$65\,ms/view and DOT (point tracks+dense) takes ${\sim}$0.9\,s/frame amortized over 8-frame windows in general. We use LightGlue+GMFlow on Monaco4D and LightGlue+DOT in CMU-Panoptic.
All baselines similarly exclude equivalent preprocessing from their reported times (TrackerSplat excludes point tracks, D-3DGS excludes segmentation masks, QUEEN excluded depth estimation).

\begin{table*}
\centering
\caption{\textbf{Per-frame timing breakdown.} Times in seconds.
  \emph{Offline} stages are run once per sequence (precomputed).
  For our method, \emph{online} stages are parallelized into independent lanes
  executing on separate GPUs; wall-clock time equals the slowest lane.}
\label{tab:timing}
\scalebox{1}{
\begin{tabular}{@{}lccccc@{}}
\toprule
\multicolumn{1}{l|}{Component}                                 & \multicolumn{1}{c|}{Parallelism}      & \multicolumn{1}{c|}{Ours (Monaco4D)}                                         & \multicolumn{1}{c|}{Dynamic-3DGS} & \multicolumn{1}{c|}{TrackerSplat}               & Ours (CMU Panoptic) \\ \midrule
\multicolumn{6}{l}{Offline pre-processing}                                                                                                                                                                                                                                                        \\ \midrule
\multicolumn{1}{l|}{Feature tracker}                           & \multicolumn{1}{c|}{views × trackers} & \multicolumn{1}{c|}{0.031}                                                   & \multicolumn{1}{c|}{-}            & \multicolumn{1}{c|}{-}                          & 0.031               \\
\multicolumn{1}{l|}{Point tracker}                             & \multicolumn{1}{c|}{views × trackers} & \multicolumn{1}{c|}{727.38}                                                  & \multicolumn{1}{c|}{-}            & \multicolumn{1}{c|}{727.38}                     & 727.38              \\
\multicolumn{1}{l|}{Optical flow}                              & \multicolumn{1}{c|}{views}            & \multicolumn{1}{c|}{0.035}                                                   & \multicolumn{1}{c|}{-}            & \multicolumn{1}{c|}{-}                          & 0.035               \\
\multicolumn{1}{l|}{Segmentation}                              & \multicolumn{1}{c|}{views}            & \multicolumn{1}{c|}{0.015}                                                   & \multicolumn{1}{c|}{0.015}        & \multicolumn{1}{c|}{0.015}                      & 0.015               \\ \midrule
\multicolumn{1}{l|}{Pre-proc. total}                           & \multicolumn{1}{c|}{}                 & \multicolumn{1}{c|}{727.461}                                                 & \multicolumn{1}{c|}{0.015}        & \multicolumn{1}{c|}{727.395}                    & 727.461             \\ \midrule
\multicolumn{6}{l}{Online — Dynamic lane}                                                                                                                                                                                                                                                         \\ \midrule
\multicolumn{1}{l|}{Trackers (load)}                           & \multicolumn{1}{c|}{views × trackers} & \multicolumn{1}{c|}{\textless{}0.1}                                          & \multicolumn{1}{c|}{-}            & \multicolumn{1}{c|}{-}                          & \textless{}0.1      \\
\multicolumn{1}{l|}{Voronoi interpolation}                     & \multicolumn{1}{c|}{views × trackers} & \multicolumn{1}{c|}{0.5}                                                     & \multicolumn{1}{c|}{-}            & \multicolumn{1}{c|}{-}                          & 0.5                 \\
\multicolumn{1}{l|}{Cross-level disagreement}                  & \multicolumn{1}{c|}{views}            & \multicolumn{1}{c|}{\textless{}0.1}                                          & \multicolumn{1}{c|}{-}            & \multicolumn{1}{c|}{-}                          & \textless{}0.1      \\
\multicolumn{1}{l|}{Triangulation}                             & \multicolumn{1}{c|}{per tracker}      & \multicolumn{1}{c|}{0.2}                                                     & \multicolumn{1}{c|}{-}            & \multicolumn{1}{c|}{-}                          & 0.2                 \\
\multicolumn{1}{l|}{Fusion}                                    & \multicolumn{1}{c|}{-}                & \multicolumn{1}{c|}{0.15}                                                    & \multicolumn{1}{c|}{-}            & \multicolumn{1}{c|}{-}                          & 0.15                \\
\multicolumn{1}{l|}{Initialization}                            & \multicolumn{1}{c|}{-}                & \multicolumn{1}{c|}{0.85}                                                    & \multicolumn{1}{c|}{-}            & \multicolumn{1}{c|}{-}                          & -                   \\
\multicolumn{1}{l|}{Finetuning}                                & \multicolumn{1}{c|}{-}                & \multicolumn{1}{c|}{$\sim$9.5}                                               & \multicolumn{1}{c|}{738}          & \multicolumn{1}{c|}{72}                         & $\sim$3.0           \\ \midrule
\multicolumn{1}{l|}{Dynamic total}                             & \multicolumn{1}{c|}{}                 & \multicolumn{1}{c|}{\cellcolor[HTML]{FD8488}$\sim$10.5}                      & \multicolumn{1}{c|}{738}          & \multicolumn{1}{c|}{\cellcolor[HTML]{FEC088}72} & $\sim$4.0           \\ \midrule
\multicolumn{6}{l}{Online — Static and Skybox lanes (ours only)}                                                                                                                                                                                                                                  \\ \midrule
\multicolumn{1}{l|}{SH finetuning (500 iter.)}                 & \multicolumn{1}{c|}{-}                & \multicolumn{1}{c|}{$\sim$25}                                                & \multicolumn{1}{c|}{-}            & \multicolumn{1}{c|}{-}                          & $\sim$1.0           \\
\multicolumn{1}{l|}{Texture optimization}                      & \multicolumn{1}{c|}{-}                & \multicolumn{1}{c|}{$\sim$1.0}                                               & \multicolumn{1}{c|}{-}            & \multicolumn{1}{c|}{-}                          & -                   \\ \midrule
\multicolumn{1}{l|}{\textbf{Wall-clock online (max of lanes)}} & \multicolumn{1}{c|}{}                 & \multicolumn{1}{c|}{\cellcolor[HTML]{FD8488}{\color[HTML]{333333} $\sim$25}} & \multicolumn{1}{c|}{738}          & \multicolumn{1}{c|}{\cellcolor[HTML]{FEC088}72} & $\sim$4.0           \\
\multicolumn{1}{l|}{\textbf{Wall-clock total (+ offline)}}     & \multicolumn{1}{c|}{}                 & \multicolumn{1}{c|}{$\sim$752}                                               & \multicolumn{1}{c|}{738.015}      & \multicolumn{1}{c|}{799.595}                    & $\sim$731           \\ \bottomrule
\end{tabular}
}
\end{table*}

While pre-processing is dominated by the point tracker, it runs in parallel across frames and views. The practical bottleneck is \textbf{sequential} finetuning, which FastFlowGS reduces from 738\,s to 9.5\,s/frame (\textbf{78$\times$}). Notably, TrackerSplat uses the same point tracker yet requires 72\,s for finetuning (7.6$\times$ more), confirming that the reduction stems from our multi-resolution fusion, not from the external priors themselves.

\begin{table}[]
\caption{\textbf{Qualitative results with fewer primitives in background: Per sequence} FastFlowGS achieves comparable quality with enhanced convergence speed (Time is measured in seconds). QUEEN ran out-of-memory for Sequence Main Straight, noted with --}
\label{tab:all_seq_ind}
\begin{tabular}{@{}l|c|ccccc|c@{}}
\toprule
\multicolumn{1}{c|}{Sequence}   & Metrics & D-3DGS                        & HiCoM                         & QUEEN                         & ReConGs & Trackersplat                  & FastFlowGS                    \\ \midrule
                                & VMAF    & \cellcolor[HTML]{FEC088}40.47 & 0.89                          & 5.39                          & 0.71    & 2.72                          & \cellcolor[HTML]{FD8488}46.89 \\
                                & PSNR    & 20.49                         & \cellcolor[HTML]{FEC088}10.35 & 21.73                         & 10.27   & 15.31                         & \cellcolor[HTML]{FD8488}24.5  \\
                                & MPSNR   & 17.49                         & 10.75                         & \cellcolor[HTML]{FD8488}23.54 & 10.31   & 15.86                         & \cellcolor[HTML]{FEC088}19.78 \\
                                & Time    & 738                           & \cellcolor[HTML]{FEC088}27    & 61                            & 48      & 72                            & \cellcolor[HTML]{FD8488}25    \\
                                & VE      & 0.05                          & 0.03                          & \cellcolor[HTML]{FEC088}0.09  & 0.01    & 0.04                          & \cellcolor[HTML]{FD8488}1.88  \\
\multirow{-6}{*}{Fairmont}      & PE      & 0.03                          & \cellcolor[HTML]{FEC088}0.38  & 0.36                          & 0.21    & 0.21                          & \cellcolor[HTML]{FD8488}0.98  \\ \midrule
                                & VMAF    & \cellcolor[HTML]{FEC088}35.83 & 1.09                          & --                          & 1.21    & 0.78                          & \cellcolor[HTML]{FD8488}50.1  \\
                                & PSNR    & \cellcolor[HTML]{FEC088}17.34 & 9.82                          & --                         & 9.62    & 14.57                         & \cellcolor[HTML]{FD8488}21.97 \\
                                & MPSNR   & 15.99                         & 9.79                          & -- & 9.81    & \cellcolor[HTML]{FEC088}16.99                         & \cellcolor[HTML]{FD8488}18.91 \\
                                & Time    & 724                           & \cellcolor[HTML]{FEC088}24    & --                            & 48      & 95                            & \cellcolor[HTML]{FD8488}21    \\
                                & VE      & \cellcolor[HTML]{FEC088}0.05  & \cellcolor[HTML]{FEC088}0.05  & --                          & 0.03    & 0.01                          & \cellcolor[HTML]{FD8488}2.39  \\
\multirow{-6}{*}{Main Straight} & PE      & 0.02                          & \cellcolor[HTML]{FEC088}0.41  & --                           & 0.2     & 0.15                          & \cellcolor[HTML]{FD8488}1.05  \\ \midrule
                                & VMAF    & \cellcolor[HTML]{FEC088}38.73 & 2.01                          & 12.67                         & 1.66    & 9.67                          & \cellcolor[HTML]{FD8488}53.84 \\
                                & PSNR    & \cellcolor[HTML]{FEC088}17.43 & 10.64                         & 12.02                         & 10.76   & 16.83                         & \cellcolor[HTML]{FD8488}23    \\
                                & MPSNR   & 15.11                         & 11.14                         & 9.62                          & 11.19   & \cellcolor[HTML]{FD8488}21.34 & \cellcolor[HTML]{FEC088}17.53 \\
                                & Time    & 729                           & 25                            & \cellcolor[HTML]{FEC088}23    & 35      & 70                            & \cellcolor[HTML]{FD8488}23    \\
                                & VE      & 0.05                          & 0.08                          & \cellcolor[HTML]{FEC088}0.58  & 0.05    & 0.14                          & \cellcolor[HTML]{FD8488}2.34  \\
\multirow{-6}{*}{Rascasse}      & PE      & 0.02                          & 0.43                          & \cellcolor[HTML]{FEC088}0.55  & 0.31    & 0.24                          & \cellcolor[HTML]{FD8488}1     \\ \midrule
                                & VMAF    & \cellcolor[HTML]{FEC088}32.11 & 0.25                          & 0.39                          & 0.36    & 7.15                          & \cellcolor[HTML]{FD8488}48.9  \\
                                & PSNR    & \cellcolor[HTML]{FEC088}17.76 & 10.39                         & 20.15                         & 10.56   & 15.57                         & \cellcolor[HTML]{FD8488}21.83 \\
                                & MPSNR   & 16.81                         & 11.62                         & 23.35                         & 10.11   & \cellcolor[HTML]{FD8488}19.91 & \cellcolor[HTML]{FEC088}16.92 \\
                                & Time    & 732                           & \cellcolor[HTML]{FEC088}24    & 69                            & 39      & 94                            & \cellcolor[HTML]{FD8488}19    \\
                                & VE      & 0.04                          & 0.01                          & 0.01                          & 0.01    & \cellcolor[HTML]{FEC088}0.08  & \cellcolor[HTML]{FD8488}2.57  \\
\multirow{-6}{*}{Tunnel}        & PE      & 0.02                          & \cellcolor[HTML]{FEC088}0.43  & 0.29                          & 0.27    & 0.17                          & \cellcolor[HTML]{FD8488}1.15  \\ \midrule
                                & VMAF    & \cellcolor[HTML]{FEC088}41.17 & 0.97                          & 0.219                         & 0.89    & 1.59                          & \cellcolor[HTML]{FD8488}55.84 \\
                                & PSNR    & \cellcolor[HTML]{FEC088}18.66 & 10.22                         & 13.73                         & 11.41   & 15.64                         & \cellcolor[HTML]{FD8488}22.28 \\
                                & MPSNR   & 17.92                         & 10.19                         & 9.71                          & 10.72   & \cellcolor[HTML]{FEC088}20.92 & \cellcolor[HTML]{FD8488}20.97 \\
                                & Time    & 740                           & \cellcolor[HTML]{FEC088}29    & 70                            & 39      & 60                            & \cellcolor[HTML]{FD8488}27    \\
                                & VE      & \cellcolor[HTML]{FEC088}0.06  & 0.03                          & 0                             & 0.02    & 0.03                          & \cellcolor[HTML]{FD8488}2.07  \\
\multirow{-6}{*}{Uphill}        & PE      & 0.03                          & \cellcolor[HTML]{FEC088}0.35  & 0.2                           & 0.29    & 0.26                          & \cellcolor[HTML]{FD8488}0.83  \\ \midrule
                                & VMAF    & \cellcolor[HTML]{FEC088}34.36 & 1.16                          & 38.75                         & 1.24    & 6.77                          & \cellcolor[HTML]{FD8488}51.87 \\
                                & PSNR    & \cellcolor[HTML]{FEC088}19.82 & 9.64                          & 23.16                         & 10.19   & 15.24                         & \cellcolor[HTML]{FD8488}24.6  \\
                                & MPSNR   & 17.47                         & 10.19                         & \cellcolor[HTML]{FD8488} 24.22                         & 9.62    & \cellcolor[HTML]{FEC088} 17.64 & 11.43 \\
                                & Time    & 727                           & 26                            & \cellcolor[HTML]{FEC088}23    & 25      & 58                            & \cellcolor[HTML]{FD8488}22    \\
                                & VE      & 0.05                          & 0.04                          & \cellcolor[HTML]{FEC088}1.68  & 0.05    & 0.12                          & \cellcolor[HTML]{FD8488}2.36  \\
\multirow{-6}{*}{Pool}          & PE      & 0.03                          & 0.37                          & \cellcolor[HTML]{FEC088}1.01  & 0.41    & 0.26                          & \cellcolor[HTML]{FD8488}1.12  \\ \bottomrule
\end{tabular}
\end{table}

\begin{algorithm}[t]
\caption{FastFlowGS: Per-Frame Streaming Update ($t > 0$)}
\label{alg:fastflowgs}
\begin{algorithmic}[1]

\REQUIRE Gaussians from frame $t{-}1$ with positions $\mu_{i}$ and covariances $\mathbf{P}_{i}$;\\ multi-view images $\{I_v\}_{v=1}^{V}$; foreground masks $\{M_v\}$; tracker set $L$
\ENSURE Updated Gaussian positions $\mu_{i}$ and covariances $\boldsymbol{\Sigma}_{i}$

\STATE Project each $\mu_{i}$ into views; mark as dynamic if inside previous mask

\STATE \textit{// Estimate multi-resolution 2D motion fields}
\FOR{each tracker $l \in L$ \textbf{in parallel}}
    \STATE Compute 2D flow $\mathbf{F}_l$ between frames $t{-}1$ and $t$ for all views
    \IF{$l$ is sparse or medium}
        \STATE Voronoi-densify to full resolution; record distance confidence $C^{\text{vor}}_l$
    \ENDIF
\ENDFOR

\STATE \textit{// Score cross-level agreement}
\FOR{each tracker $l \in L$}
    \STATE $C_l \leftarrow C^{\text{vor}}_l \times \exp(-\text{pairwise flow disagreement} \;/\; \text{mean flow magnitude})$
\ENDFOR

\STATE \textit{// Associate Gaussians with pixels}
\STATE Render previous Gaussians; record top-$K$ indices and blending weights $\alpha$ per pixel
\FOR{each Gaussian $i$, view $v$, tracker $l$}
    \STATE Weight $w \leftarrow \alpha \times C_l \times \|\text{flow}\|$; \; 
    \STATE displaced mean $\hat{x}^{2d} \leftarrow$ projected mean $+$ $\alpha$-weighted flow
\ENDFOR

\STATE \textit{// Triangulate each tracker independently}
\FOR{each tracker $l \in L$, each Gaussian $i$ \textbf{in parallel}}
    \STATE Lift displaced 2D positions to viewing rays; discard views with motion nearly collinear to ray ($< 25^{\circ}$)
    \STATE Solve weighted least-squares for 3D position $\hat{\mu}_{i,l}$ and covariance $\hat{\boldsymbol{\Sigma}}_{i,l}$
\ENDFOR

\STATE \textit{// Fuse tracker estimates with temporal prior (Kalman update)}
\STATE $\mathbf{Q} \leftarrow q^2 \mathbf{I}$, where $q =$ median 3D displacement across triangulated Gaussians
\FOR{each Gaussian $i$}
    \STATE $\mathbf{P}^{-} \leftarrow \mathbf{P}_{i} + \mathbf{Q}$ \hfill $\rhd$ Predicted covariance
    \STATE $\boldsymbol{\Sigma}_{i}^{-1} \leftarrow (\mathbf{P}^{-})^{-1} + \sum_{l} \hat{\boldsymbol{\Sigma}}_{i,l}^{-1}$ \hfill $\rhd$ Fused precision
    \STATE $\mu_{i} \leftarrow \boldsymbol{\Sigma}_{i}\bigl[(\mathbf{P}^{-})^{-1} \mu_{i}^{\text{prev}} + \sum_{l} \hat{\boldsymbol{\Sigma}}_{i,l}^{-1}\, \hat{\mu}_{i,l}\bigr]$
    \IF{all tracker levels failed}
        \STATE $\mu_{i} \leftarrow \mu_{i}^{\text{prev}}$; \; $\boldsymbol{\Sigma}_{i} \leftarrow \mathbf{P}^{-}$
    \ENDIF
\ENDFOR

\STATE \textit{// Geometric validation}
\FOR{each Gaussian $i$}
    \IF{$\mu_{i}$ projects inside mask in fewer than $V{-}2$ views}
        \STATE Mark as invalid
    \ENDIF
\ENDFOR

\STATE \textit{// Finetuning and densification}
\STATE Split or clone Gaussians with high $\mathrm{tr}(\boldsymbol{\Sigma}_{i})$
\FOR{Gaussian state}
    \STATE Dynamic: optimize all parameters (foreground-masked $\mathcal{L}_1$-SSIM) \hfill $\rhd$ In parallel
    \STATE Static: optimize spherical harmonics only \hfill $\rhd$ In parallel
    \ENDFOR

\STATE Composite static, dynamic, and skybox via alpha blending

\end{algorithmic}
\end{algorithm}
\clearpage

\subsection{Confidence-Weighted Voronoi Interpolation}
\label{sec:supp_voronoi}

Sparse and medium-density trackers produce correspondences at a subset of pixels.
We densify each tracker's output to full resolution via Gaussian-weighted $K$-nearest-neighbor interpolation, adapted from DOT~\cite{moing2024denseopticaltrackingconnecting}.
Algorithm~\ref{algo:voronoi} details the procedure, which jointly produces a dense flow field, interpolated target visibility, and a per-pixel distance confidence $\mathbf{C}^{\text{vor}}$.

This confidence enters the cross-level agreement and propagates into the triangulation weights, ensuring that pixels far from any observed track are downweighted throughout the pipeline.
We use $K\!=\!8$ and $\sigma\!=\!0.05$ in all experiments.

\begin{algorithm}[t]
\caption{Gaussian-Weighted Voronoi Interpolation (adapted from DOT~\cite{moing2024denseopticaltrackingconnecting})}
\label{algo:voronoi}
\begin{algorithmic}[1]
    \REQUIRE \\ Sparse track source positions $\{\mathbf{p}^{\text{src}}_n\}_{n=1}^{N} \subset \mathbb{R}^2$ (pixels) \\Target positions $\{\mathbf{p}^{\text{tgt}}_n\}_{n=1}^{N} \subset \mathbb{R}^2$ (pixels) \\Target visibility $\{\alpha^{\text{tgt}}_n\}_{n=1}^{N} \subset [0,1]$ \\ Image size $H \times W$ \\ Number of neighbors $K$ \\Gaussian kernel scale $\sigma$
    \ENSURE \\ Dense flow $\mathbf{F} \in \mathbb{R}^{H \times W \times 2}$ \\ Interpolated target visibility $\hat{\alpha} \in \mathbb{R}^{H \times W}$ \\ Distance confidence $\mathbf{C}^{\text{vor}} \in \mathbb{R}^{H \times W}$
    \FOR{each pixel $\mathbf{p} = (u, v)$ in $[0, W\!-\!1] \times [0, H\!-\!1]$}
        \STATE Find $K$ nearest source tracks by Euclidean distance in normalized coordinates: $\bar{\mathbf{p}} = \left(\frac{u}{W-1},\, \frac{v}{H-1}\right)$,\; $\bar{\mathbf{p}}^{\text{src}}_n = \left(\frac{p^{\text{src}}_{n,x}}{W-1},\, \frac{p^{\text{src}}_{n,y}}{H-1}\right)$
        \STATE Let $\{d^2_k\}_{k=1}^{K}$ be the squared distances to the $K$ neighbors (sorted ascending)
        \STATE Gaussian weights:\; $w_k = \exp\!\left(-\,d^2_k \,/\, 2\sigma^2\right)$
        \STATE Distance confidence:\; $\mathbf{C}^{\text{vor}}(\mathbf{p}) = w_1$ \hfill $\rhd$ Nearest-neighbor weight
        \STATE Normalize:\; $\hat{w}_k = w_k \,/\, \sum_{j=1}^{K} w_j$
        \STATE Dense flow:\; $\mathbf{F}(\mathbf{p}) = \sum_{k=1}^{K} \hat{w}_k \left(\mathbf{p}^{\text{tgt}}_k - \mathbf{p}^{\text{src}}_k\right)$
        \STATE Target visibility:\; $\hat{\alpha}(\mathbf{p}) = \sum_{k=1}^{K} \hat{w}_k \,\alpha^{\text{tgt}}_k$
    \ENDFOR
\end{algorithmic}
\end{algorithm}
\clearpage

\section{Monaco4D: Additional Details}
\label{sec:supp_dataset}
This section provides the rendering and scene configuration details needed to reproduce Monaco4D.
All sequences are built in Unreal Engine~5~\cite{unrealengine5} using path-traced rendering at real-world scale (1 Unreal Unit\,=\,1\,cm).

\subsection{Camera Configuration}
\label{sec:supp_cameras}

\begin{wrapfigure}{r}{5.5cm}
\includegraphics[width=5.5cm]{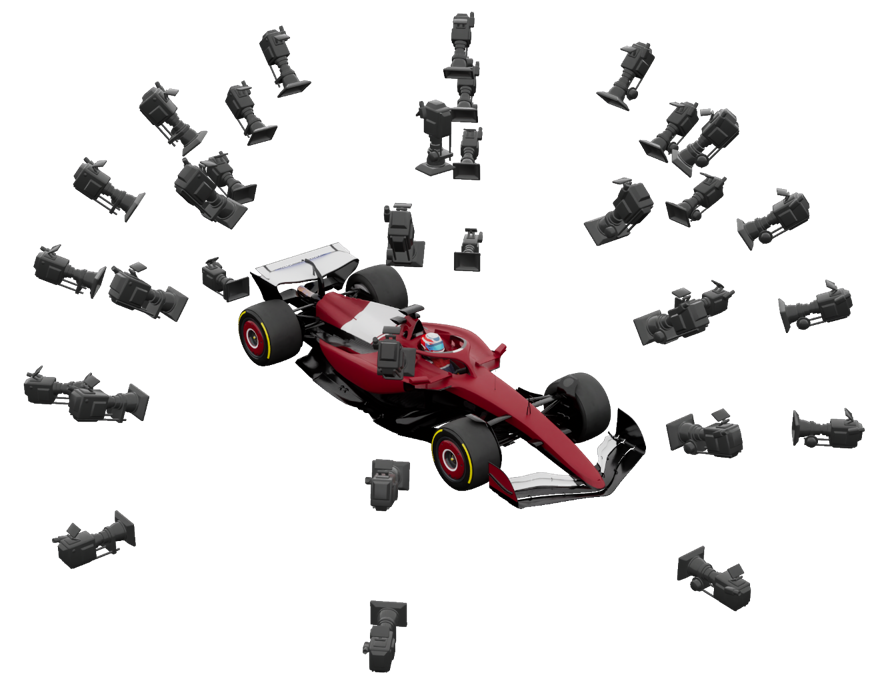}
\caption{\textbf{Hemispherical initialization rig.} At $t\!=\!0$, a dense hemisphere of cameras surrounds each vehicle to bootstrap a high-quality first-frame Gaussian reconstruction. This provides the initial 3D representation that streaming methods then propagate forward in time. Hemisphere views are used only for initialization and are excluded from all evaluation metrics.}
\label{fig:binding}
\end{wrapfigure}

 \paragraph{Intrinsics.}
All cameras use Unreal Engine's \texttt{CineCameraActor} with a 16:9 digital filmback, a fixed 12\,mm prime lens at $f$/2.8, and no autofocus, yielding a horizontal field of view of approximately 89.4\textdegree.
Intrinsic parameters are identical across every camera in the dataset. Hence, differences between viewpoints arise solely from position and orientation.

\paragraph{Initialization hemispheres.}
\label{supp_hemisphere}
To provide a high-quality first-frame reconstruction for each sequence, we additionally place a dense hemispherical camera grid around each vehicle at $t\!=\!0$ (Fig.~\ref{fig:binding}).
These views are used only for initialization and are excluded from evaluation.

\paragraph{Trackside cameras.}
Cameras are distributed along the Monaco circuit along the guardrails, as shown in the track overview of Fig.~\ref{fig:track_layout}.
The number varies by track section: Main Straight (478), Uphill (342), Fairmont (232), Tunnel (198), Pool (135), and Rascasse (92).
Frame counts vary per sequence with the duration of the captured motion.

\begin{figure}[t]
    \centering
    \includegraphics[width=1\linewidth]{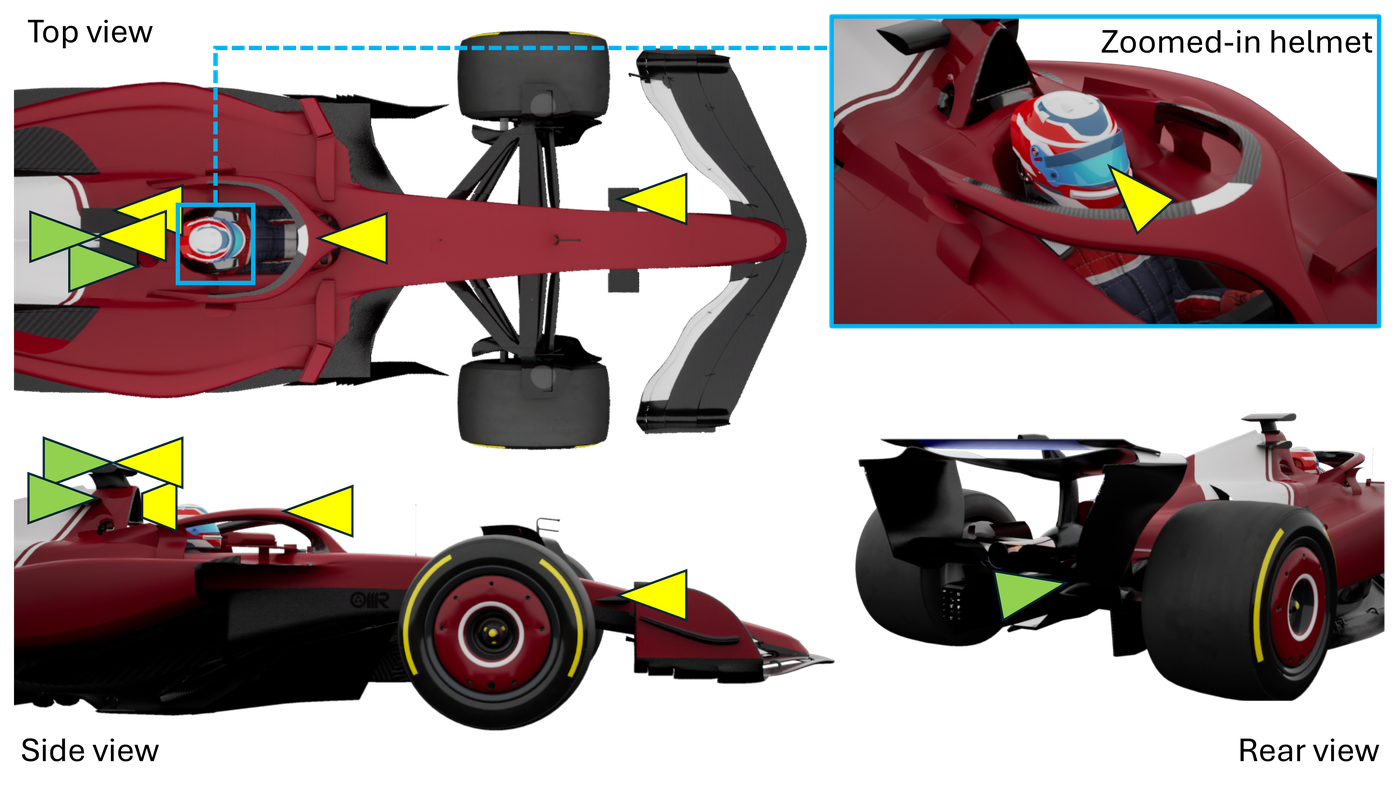}
    \caption{\textbf{Onboard camera placement.} Four viewpoints of a single vehicle illustrating the seven per-car camera positions that follow FIA broadcast regulations~\cite{fia2026}. \textcolor{yellow}{triangles} depict forward facing cameras while \textcolor{green}{triangles} depict rear facing cameras.}
    \label{fig:camera_placement}
\end{figure}

\paragraph{Onboard cameras.}
Each vehicle carries seven onboard cameras that replicate standard Formula~1 broadcast mounting positions per FIA regulations~\cite{fia2026}: forward, rear, lateral, and driver perspectives (Fig.~\ref{fig:camera_placement}).
Onboard cameras share the same intrinsic configuration as trackside cameras.

\begin{wraptable}{R}{5.5cm}
\caption{\textbf{Distribution of cameras:} The number of cameras and frames per sequence, per variation. We also present the average number of views that are dynamic (have atleast $1$ car) per frame}\label{wrap-tab:1}
\scalebox{0.65}{
\begin{tabular}{@{}l|c|c|c|c@{}}
\toprule
\multicolumn{1}{c|}{Sequence}  & Variation & Frames & \begin{tabular}[c]{@{}c@{}}Trackside\\ Cameras\end{tabular} & \begin{tabular}[c]{@{}c@{}}Frames with \\ car  per \\ camera\end{tabular} \\ \midrule
\multirow{3}{*}{Main Straight} & V1        & 214    & 478                                                         & 17                                                                   \\
                               & V2        & 210    & 478                                                         & 11                                                                   \\
                               & V3        & 224    & 478                                                         & 25                                                                   \\ \midrule
\multirow{2}{*}{Rascasse}      & V1        & 147    & 92                                                          & 28                                                                   \\
                               & V2        & 147    & 92                                                          & 52                                                                   \\ \midrule
\multirow{2}{*}{Uphill}        & V1        & 197    & 342                                                         & 20                                                                   \\
                               & V2        & 233    & 342                                                         & 49                                                                   \\ \midrule
Pool                           & V1        & 146    & 135                                                         & 16                                                                   \\ \midrule
\multirow{2}{*}{Tunnel}        & V1        & 320    & 198                                                         & 19                                                                   \\
                               & V2        & 335    & 198                                                         & 32                                                                   \\ \midrule
\multirow{3}{*}{Fairmont}      & V1        & 241    & 232                                                         & 23                                                                   \\
                               & V2        & 270    & 232                                                         & 41                                                                   \\
                               & V3        & 262    & 232                                                         & 71                                                                   \\ \bottomrule
\end{tabular}
}
\end{wraptable}

\paragraph{Drone cameras.} \raggedright
Three aerial trajectories per sequence simulate broadcast drone operation: subject tracking, rapid panning, and abrupt zoom changes. Their flight paths are overlaid on the track layout in Fig.~\ref{fig:track_layout} in color.
\\

\subsection{Environment and Track Geometry}
\label{sec:supp_environment}
 
The surrounding urban structures (buildings, grandstands, background elements) originate from a commercially available Monaco circuit asset (Fab).
We reconstructed the track surface and several structural components from scratch to ensure accurate geometry, consistent topology, and full control over materials.
The drivable track geometry and camera placements are metric (1 UE unit\,=\,1\,cm), so vehicle speeds, inter-camera baselines, and camera-to-track distances correspond to real-world values.
Surrounding city structures are optimized for visual realism and occlusion fidelity rather than survey-grade geometric accuracy. Hence, while they preserve the characteristic layout and proportions of Monaco, they should not be treated as a precise urban reconstruction.

\subsection{Weather and Illumination Layers}
To generate environmental variability in the rendered dataset, we use the Ultra Dynamic Sky plugin in Unreal Engine. This plugin provides a procedural sky, lighting, cloud, fog, and weather system that can be controlled directly inside the scene. Instead of creating a single fixed lighting setup, we organize the environment into separate illumination and weather layers, each corresponding to a specific capture condition.

In our setup, we define individual layers for daytime illumination, night, foggy conditions, and rainy weather. Each layer stores the required scene configuration for that condition, including sky appearance, sun or moon contribution, atmospheric fog, cloud coverage, weather intensity, and the corresponding visual effects. This allows us to switch between different environmental states while keeping the geometry, camera setup, and scene layout unchanged.

We use the \texttt{daytime} layer to simulate clear or naturally lit outdoor conditions, with stronger directional illumination from the sun and brighter sky contribution. The \texttt{night} layer reduces the global illumination and changes the sky and atmospheric response to produce low-light captures. The \texttt{fog} layer increases atmospheric scattering and reduces scene visibility, creating images with lower contrast and stronger depth-dependent haze. The \texttt{rain} layer activates precipitation effects and modifies the overall appearance of the scene to represent wet weather conditions.

\begin{figure}[t]
    \centering
    \includegraphics[width=1\linewidth]{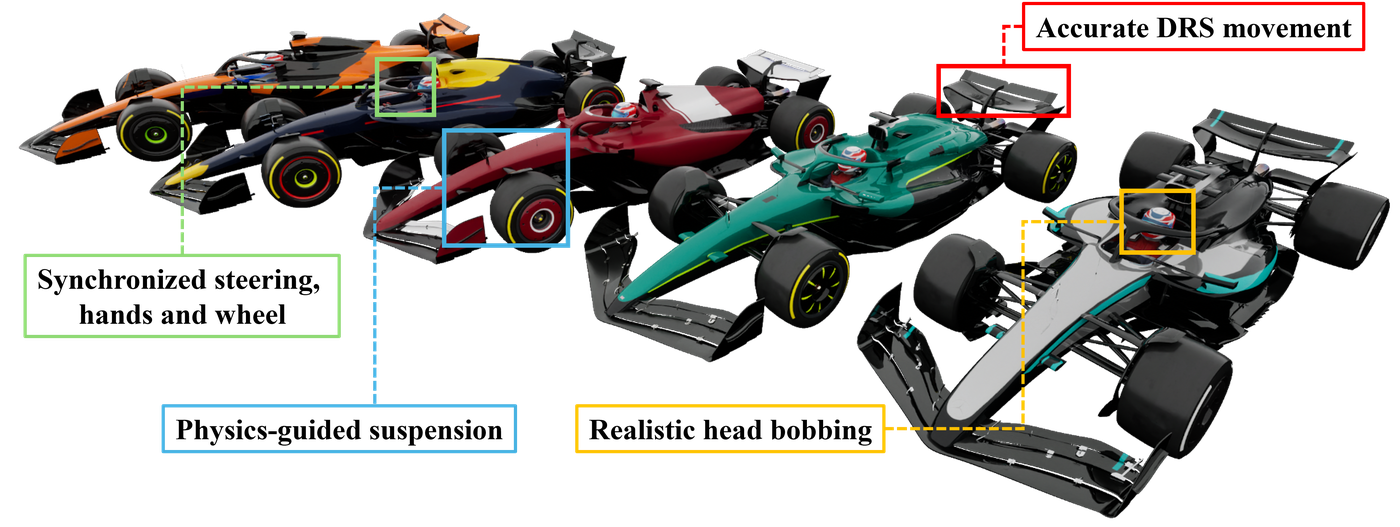}
    \caption{\textbf{Vehicle assets and dynamic rigging.} Each vehicle integrates physics-based suspension, DRS flap actuation triggered at circuit-accurate positions, a steering linkage that constrains the driver's hands to the wheel, and head bobbing coupled to the suspension state. These mechanisms enforce that vehicle appearance changes frame-to-frame in response to track forces rather than following canned animation.}
    \label{fig:car_assets}
\end{figure}

\begin{figure}
    \centering
    \includegraphics[width=1\linewidth]{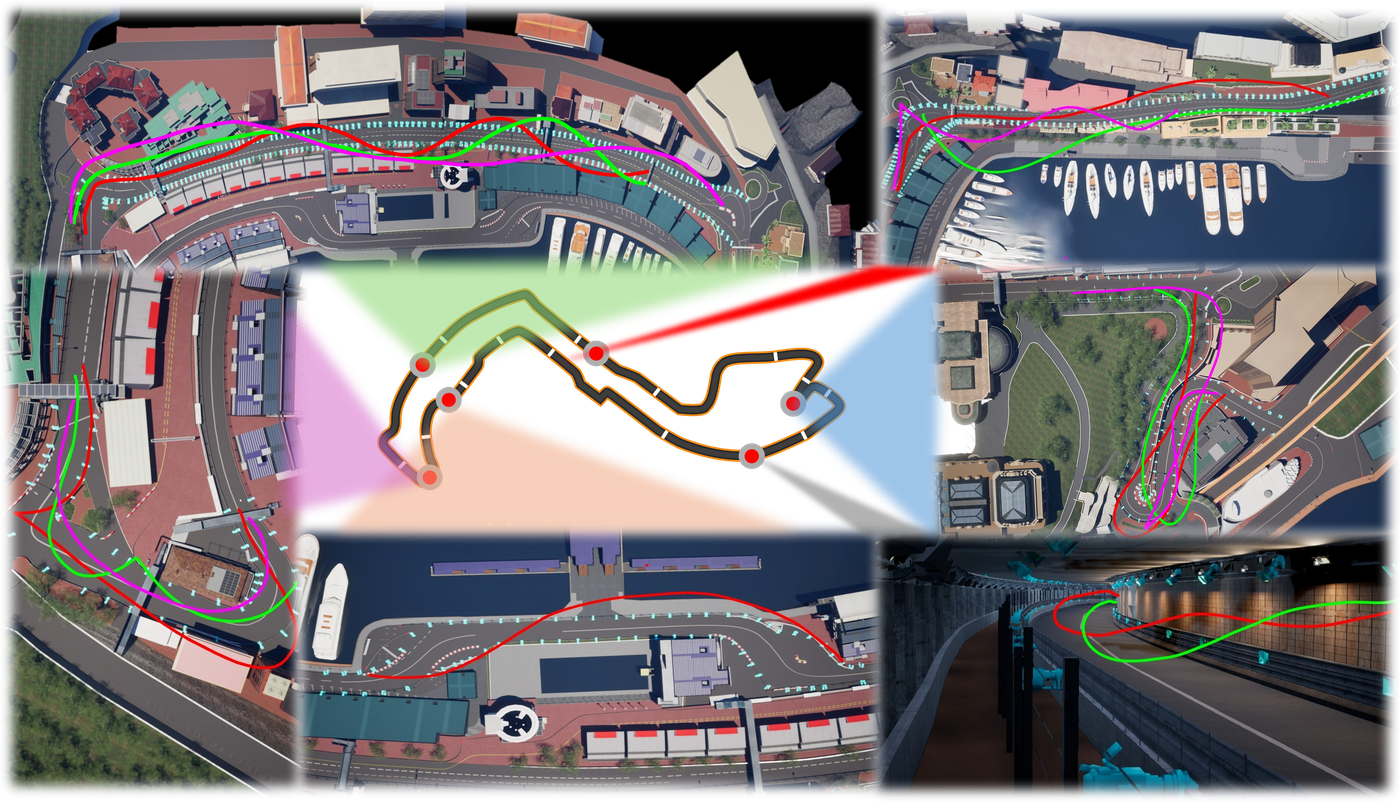}
    \caption{\textbf{Track layout, camera distribution, and drone trajectories.} Center: the full Monaco circuit with the seven sequence segments highlighted. Insets: per-section camera placements viewed from above, showing the density and angular spread of trackside coverage. Colored paths overlaid on each segment trace drone trajectories.}
    \label{fig:track_layout}
\end{figure}

\subsection{Vehicle Models and Animation}
\label{sec:supp_vehicles}

The Formula~1 car and driver models are based on commercially available assets, integrated into Unreal Engine with custom rigging (Fig.~\ref{fig:car_assets}).
The vehicle control rig incorporates physics-based suspension that produces dynamic body roll and pitch in response to track forces, realistic DRS flap actuation triggered at the correct circuit positions, and a steering linkage that constrains the driver's hands to the wheel so that driver pose follows steering input automatically.
Head bobbing is driven by the suspension state, coupling driver motion to vehicle dynamics rather than relying on canned animation.

\paragraph{Speed calibration.}
Vehicles follow trajectories along the track at speeds calibrated against real Formula~1 segment times: simulated lap-segment durations fall within 90\% of their real-world counterparts.
The effective speed was reduced by roughly 10\% so that each camera captures the vehicle across a larger number of frames per pass, increasing temporal coverage per observation without altering the spatial layout.
Even at this reduced speed, peak inter-frame displacements reach 200--400\,pixels for laterally oriented trackside cameras, placing the dataset well beyond the operating range of standard optical flow networks.

\subsection{Rendering Configuration}
\label{sec:supp_rendering}
 
Rendering uses the Unreal Engine Path Tracer with 64 spatial samples per pixel and 1 temporal sample.
Standard anti-aliasing is disabled and temporal AA samples are set to 8.
The path tracing denoiser is enabled with a 2-frame noise reduction window.
Frames are exported as 8-bit PNG sequences at an effective render resolution of 150\% screen percentage.
Geometry culling for instanced static meshes is disabled such that all geometry remains visible to the path tracer.
 
A 128-frame warm-up stage (both render and engine) precedes each sequence, to allow lighting and temporal state to converge before capture begins.

\subsection{Lighting and Atmosphere}
\label{sec:supp_lighting}
 
Illumination combines four physically motivated components:
 
\paragraph{Directional light.}
A directional light represents the sun, with intensity $10.0$ and indirect lighting intensity $6.0$.
Volumetric scattering is enabled for correct light-atmosphere interaction.
 
\paragraph{Sky light.}
A \texttt{SkyLight} at intensity $1.0$ provides diffuse environmental illumination captured from the atmospheric model.
The lower hemisphere uses a solid color to prevent unrealistic below-ground contributions.

\paragraph{Atmospheric scattering.}
The \texttt{SkyAtmosphere} component models Rayleigh scattering (scale $0.0331$) and Mie scattering (anisotropy $0.8$) with additional absorption for subtle atmospheric color filtering.

\paragraph{Volumetric clouds.}
A cloud layer spans $5-10$\,km altitude, with a tracing start distance of $350$\,km and maximum tracing distance of $50$\,km, introducing natural skylight variation and soft shadowing.

\subsection{Post-Processing and Motion Blur}
\label{sec:supp_postprocess}
 
A global Post Process Volume enables the ray tracing features required by the path tracer. No additional tone mapping, stylization, or image-space effects are applied.
Motion blur is explicitly disabled (motion blur amount set to 0) so that each frame represents a sharp instantaneous capture.

\subsection{Frame Rate}
\label{sec:supp_framerate}
 
All sequences are rendered at a base rate of $30$\,FPS.
Selected sequences are additionally rendered at $240$\,FPS to provide denser temporal sampling of vehicle motion, useful for tracking, reconstruction, and temporal interpolation tasks.

\end{document}